\documentclass[runningheads]{llncs}
\usepackage{graphicx}
\usepackage{comment}
\usepackage{amsmath,amssymb} 
\usepackage{cases}
\usepackage{color}
\usepackage{subfigure}
\usepackage{csquotes}
\newcommand*\rot{\rotatebox[origin=c]{90}}
\usepackage{multirow}
\usepackage{booktabs}
\newcommand{\mcrot}[4]{\multicolumn{#1}{#2}{\rlap{\rotatebox{#3}{#4}~}}}

\newcommand*{\twoelementtable}[3][l]%
{%
    \renewcommand{\arraystretch}{0.8}%
    \begin{tabular}[t]{@{}#1@{}}%
        #2\tabularnewline
        #3%
    \end{tabular}%
}
\usepackage{array}
\usepackage{tabularx}
\usepackage{romannum}

\usepackage{mathtools}

\DeclarePairedDelimiter\abs{\lvert}{\rvert}%
\DeclarePairedDelimiter\norm{\lVert}{\rVert}%

\makeatletter
\let\oldabs\abs
\def\abs{\@ifstar{\oldabs}{\oldabs*}}
\let\oldnorm\norm
\def\norm{\@ifstar{\oldnorm}{\oldnorm*}}
\makeatother

\begin{document}
\pagestyle{headings}
\mainmatter
\def\ECCVSubNumber{2442}  

\title{Contextual-Relation Consistent Domain Adaptation for Semantic Segmentation}


\titlerunning{Contextual-Relation Consistent Semantic Segmentation}
%
\author{Jiaxing Huang\inst{1}\orcidID{0000-0002-8681-0471} \and
Shijian Lu\inst{1}\orcidID{0000-0002-6766-2506} \and
Dayan Guan\inst{1}\orcidID{0000-0001-9752-1520} \and Xiaobing Zhang\inst{2}\orcidID{0000-0002-8149-1424}}
\authorrunning{J. Huang et al.}
%
\institute{Nanyang Technological University, 50 Nanyang Avenue, Singapore 639798
\email{\{jiaxing.huang, shijian.lu, dayan.guan\}@ntu.edu.sg}
\and
University of Electronic Science and Technology of China, China\\
\email{zhangxiaobing@std.uestc.edu.cn}}
\maketitle

\begin{abstract}
Recent advances in unsupervised domain adaptation for semantic segmentation have shown great potentials to relieve the demand of expensive per-pixel annotations.
However, most existing works address the domain discrepancy by aligning the data distributions of two domains at a global image level whereas the local consistencies are largely neglected. 
This paper presents an innovative local contextual-relation consistent domain adaptation (CrCDA) technique that aims to achieve local-level consistencies during the global-level alignment.
The idea is to take a closer look at region-wise feature representations and align them for local-level consistencies.
Specifically, CrCDA learns and enforces the prototypical local contextual-relations explicitly in the feature space of a labelled source domain while transferring them to an unlabelled target domain via backpropagation-based adversarial learning.
An adaptive entropy max-min adversarial learning scheme is designed to optimally align these hundreds of local contextual-relations across domain without requiring discriminator or extra computation overhead.
The proposed CrCDA has been evaluated extensively over two challenging domain adaptive segmentation tasks ($e.g.$, GTA5 $\rightarrow$ Cityscapes and SYNTHIA $\rightarrow$ Cityscapes), and experiments demonstrate its superior segmentation performance as compared with state-of-the-art methods.

\end{abstract}

\begin{figure}[!ht]
\centering
\subfigure {\includegraphics[width=0.98\linewidth]{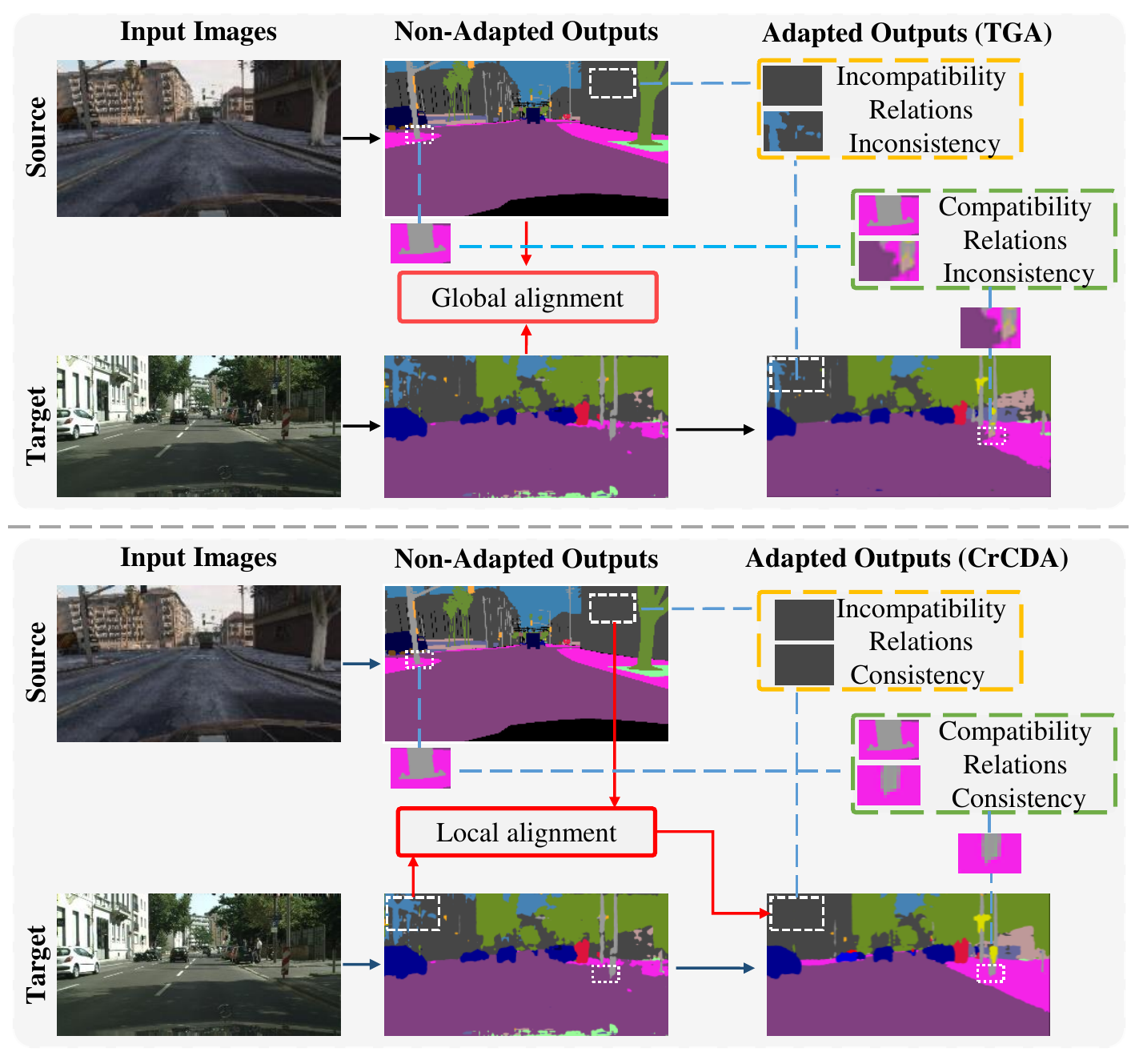}}
\caption{Our contextual-relation consistent domain adaptation (CrCDA) improves domain adaptive semantic segmentation significantly: The traditional domain adaptive segmentation shown in the upper part employs discriminators for global alignment in the output space \cite{tsai2018learning,vu2019advent,tsai2019domain} ($e.g.$, probability-/entropy-/patch-represented output), which tends to introduce segmentation errors due to the neglect of local contextual consistency.
Our CrCDA shown in the lower part adapts features at local level for contextual-relation consistency between the source and target domains which generates more accurate segmentation consistently.
In the graph, ``compatibility relations�refer to visual patterns with high co-occurrence frequency ($e.g.$, ``pole�should be beside the ``sidewalk"), and ``incompatibility relations�refer to visual patterns with low co-occurrence frequency ($i.e.$, "sky" should not in the "building").
}
\label{intro}
\end{figure}

\section{Introduction}

Semantic segmentation has been a longstanding challenge in computer vision, which aims to assign class labels to every pixel of an image \cite{zhu2016beyond}. Deep learning based approaches have achieved great successes at the price of large-scale densely-annotated datasets \cite{chen2017deeplab,long2015fully,cordts2016cityscapes} which are prohibitively expensive to collect \cite{cordts2016cityscapes}. One way of circumventing this constraint is to use synthesized images with automatically generated labels ($e.g.$, synthesized \cite{ros2016synthia} or game-engine produced \cite{richter2016playing} data) in network training. Unfortunately, such models usually undergo a drastic performance drop when applied to real-world images \cite{zhang2016understanding} due to the domain bias and shift \cite{saito2018maximum,luo2018macro,tsai2018learning,vu2018memory,Saito_2018_ECCV,luo2019significance}.

Unsupervised domain adaptation (UDA) has been introduced to address the domain bias/shift issue. To reduce the cross-domain discrepancy, most state-of-the-art UDA methods \cite{tzeng2017adversarial,tsai2018learning,luo2019taking,vu2019advent,tsai2019domain,guan2019unsupervised} exploit adversarial learning for distribution alignment in the intermediate feature \cite{tzeng2017adversarial}, output \cite{tsai2018learning,luo2019taking} or latent \cite{tsai2019domain,vu2019advent} space. Among this cohort of adversarial-based methods, a common and pivotal step is the employment of a discriminator \cite{goodfellow2014generative} that predicts a domain label for data being either source or target domain. However, the discriminator works only on image-level and merely achieves global consistency ($i.e.$, locational/spatial distributions consistency), where local contextual consistency ($i.e.$, region-wise contextual-relationships) is largely neglected.
Local contextual-relationships are ubiquitous and provide important cues for scene segmentation. They can be formulated in terms of semantic compatibility/incompatibility relations between one thing/stuff and its neighbouring things/stuff. Under this formulation, a compatibility relation is an indication of visual patterns with high co-occurrence frequency, e.g. a pole beside a sidewalk, and an incompatibility relation is an indication of visual patterns with low co-occurrence frequency, e.g. a person above a driving car. The contextual information has been extensively explored in supervised semantic segmentation, whereas the local contextual-relationships is largely neglected in unsupervised domain adaptive semantic segmentation though they're beneficial in addressing local contextual consistency and inconsistency in the target domain, as illustrated in Fig. \ref{intro}.

To this end, we propose an unsupervised domain adaptation method for semantic segmentation that explicitly models the local contextual-relations in the feature space of source domain (with label) and then transfers this contextual information into the target domain (without label), ultimately improving target domain segmentation quality, as shown in Fig. 1.
We first establish local contextual-relationships pseudo annotations in the source domain. This can be achieved by sampling regions from pixel-level ground-truth maps of source images and clustering the sampled regions to indexed $N$/$M$ groups via Dbscan \cite{ester1996density}, as illustrated in Fig. 4. With the local contextual-relationships pseudo annotations in source domain, we can train a classifier $C$ to explicitly models/learns the local contextual-relations in the feature space of source domain, and then transfers/enforces these local contextual-relations into target domain.

Following current discriminator-based global alignment methods \cite{tsai2018learning,tsai2019domain,vu2019advent,luo2019taking}, a intuitive idea is to employ hundreds of discriminators to align hundreds of contextual-relations across domain where a single discriminator focuses on a single contextual-relation, or employ just one discriminator to align all contextual-relations across domains. Obviously, the former is cumbersome which requires much redundant computation, while the latter is not aware of a variety of contextual-relations in the data distribution and may end up biasing to low-level/simple difference.
Therefore, different from current discriminator-based global alignment methods \cite{tsai2018learning,tsai2019domain,vu2019advent,luo2019taking}, we enforce these local contextual-relations on target domain via adaptive entropy max-minimizing (AEMM) between classifier $C$ and feature extractor $E$ that estimates prototypical feature representations of these local contextual-relations and congregates neighboring target incorrect samples/contextual-relations to the approximated correct source prototypes alternatively, ultimately leading to consistent local contextual-relations across domains. 
In this way, our method requires no discriminator which is normally used in UDA-based semantic segmentation and introduces training instability and extra components. In addition, this AEMM learning scheme can also be applied into pixel-/global-scale training.
The contributions of this work can be summarized in three aspects. First, we propose an unsupervised domain adaptation method for semantic segmentation that explicitly  models the local contextual-relations in the feature space of source domain (with label) and then transfers this contextual information into target domain (without label). 
To the best of our knowledge, this is the first effort to explore contextual information for UDA-based semantic segmentation.
Second, it introduces a novel adaptive entropy max-minimizing adversarial learning scheme to effectively align hundreds of local contextual-relations across domain, which requires no discriminator and adds no overhead.
Third, it shows the proposed method can be seamlessly integrated into existing domain adaptation techniques without extra overhead except two classifiers and achieves consistent improvements on semantic segmentation. 
Fourth, extensive evaluations over two challenging UDA tasks GTA5 $\rightarrow$ Cityscapes and SYNTHIA $\rightarrow$ Cityscapes show that our method achieves superior semantic segmentation performance consistently. 



\section{Related works}

Current UDA-based semantic segmentation methods are threefold: adversarial learning based approach \cite{ganin2014unsupervised,kang2019contrastive,kang2018deep,hoffman2016fcns,long2015learning,long2016unsupervised,tzeng2017adversarial,luo2019taking,tsai2018learning,chen2018road,du2019ssf,Chen_2018_CVPR,chen2018domain,yan2017mind}, image translation based approach \cite{hoffman2017cycada,sankaranarayanan2018learning,wu2018dcan,murez2018image,li2019bidirectional,zhan2019adaptive,Zhang2019lipreading,hong2018conditional,bousmalis2017unsupervised,choi2019self}, and pseudo-labels based approach \cite{zou2018unsupervised,zhu2005semi,zhong2019invariance,gong2012geodesic,huang2019learning}.

\textbf{Adversarial learning based approach:} Adversarial learning based UDA has been extensively explored for semantic segmentation, where a discriminator is employed to minimize the divergences between source and target domains in feature or output spaces. \cite{hoffman2016fcns} first applies adversarial learning for UDA based semantic segmentation by aligning feature space at global scale. 
Curriculum domain adaptation \cite{zhang2017curriculum} utilizes certain inferred properties ($e.g.$, superpixel and global label distributions) as the guidance to train the segmentation network. 
In \cite{tsai2018learning} and \cite{chen2018road}, the adversarial learning is used to align the global structure to benefit from the scene layout consistency across domains, where \cite{chen2018road} integrates a target guided distillation module to achieve style adaptation. In addition, \cite{saito2017adversarial,saito2018maximum} combines adversarial learning  and co-training to achieve domain adaptation via maximizing the discrepancy between two classifiers' outputs.

\textbf{Image translation based approach:} Inspired by the recent advances in image synthesis ($e.g.$, CycleGAN \cite{zhu2017unpaired}), a number of GAN-based methods are proposed to generate target images conditioned on the source, which can help reduce the domain discrepancy before training segmentation models. CyCADA \cite{hoffman2017cycada} uses CycleGAN to generate target images conditioned on the source images and achieves input space adaptation with a joint adversarial learning for feature alignment. A similar method, DCAN \cite{wu2018dcan}, implements channel-wise feature alignment to preserve spatial structures and semantic concepts in the generator and segmentation network. \cite{sankaranarayanan2018learning} transfers the information of the target domain to the learned embedding via the joint adversarial learning between generator and discriminator. Besides using GANs \cite{goodfellow2014generative} to align the embedding across domains, \cite{zhu2018penalizing} proposes a novel conservative loss to penalize the extremely easy and difficult cases while enhancing moderate examples.

\textbf{Re-training based approach:} Another approach of UDA based semantic segmentation is pseudo label re-training \cite{zou2018unsupervised,Zou_2019_ICCV,Lian_2019_ICCV} that uses high-confident predictions as pseudo ground truth for the target unlabelled data to finetune the model trained on the source data. In \cite{zou2018unsupervised}, class balancing and spatial prior are included to guide the iterative re-training in target domain. \cite{vu2019advent} proposes a soft-assigned version of re-training, where it enforces the \enquote{most-confused} pixels ($e.g.$, with equal probabilities for all classes) to become more confident ($i.e.$, with either low or high probability for each class) by entropy minimization. \cite{zou2018unsupervised} instead implements iterative learning on high-confident pixels.


Our method does not follow either global/class-wise feature space alignment using discriminators \cite{hoffman2016fcns,long2015learning,long2016unsupervised,tzeng2017adversarial,luo2019taking,chen2018road} or re-training on target data \cite{zou2018unsupervised,saleh2018effective}. Instead, we enforce multi-scale feature space alignment via multi-scale entropy max-minimizing. To the best of our knowledge, this is the first end-to-end multi-scale UDA network that achieves competitive performance on two challenging UDA tasks.




\section{Methods}

\begin{figure*}[t]
\centering
\subfigure {\includegraphics[width=0.98\linewidth]{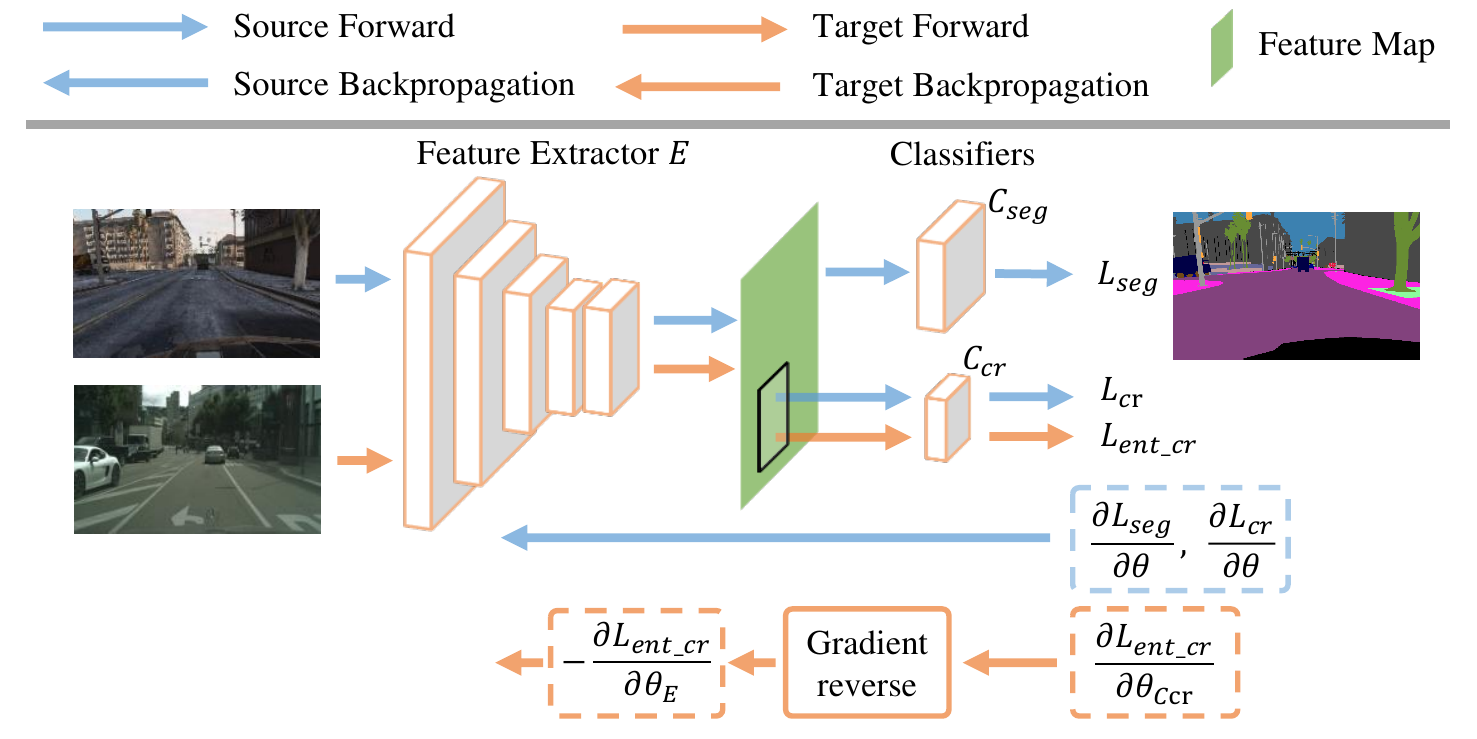}} 
\caption{Overview of our proposed contextual-relation consistent domain adaptation (CrCDA): Given images from source and target domains, the feature extractor $E$ extracts features and feeds them to classifier $C_{seg}$ and $C_{cr}$ for classification at pixel and region scales. In the source flow (highlighted by arrows in blue), $\mathcal{L}_{seg}$ is computed based on the segmentation probability map from $C_{seg}$,  $\mathcal{L}_{cr}$ is computed based on the classification probability maps from $C_{cr}$. 
In the target flow (highlighted by arrows in orange), $\mathcal{L}_{ent\_cr}$ is computed based on the classification probability maps from $C_{cr}$. 
The local-scale alignment is implemented in back-propagation by training the parts before and after the gradient reverse layer in adversarial scheme $w.r.t$ $\mathcal{L}_{ent\_cr}$.} 

\label{stru}
\end{figure*}

In this section, we present our framework for contextual-relationships consistent domain adaptation (CrCDA): a discriminator-free adversarial training scheme between a feature extractor module and a classifier via adaptive entropy max-minimizing (AEMM) to align local contextual-relationships across domains. Fig. \ref{stru} illustrates our network architecture.

\subsection{Problem Definition}
We focus on the problem of unsupervised domain adaptation (UDA) in semantic segmentation. Given the source data $X_{s} \subset \mathbb{R}^{H \times W \times 3}$ with C-class pixel-scale segmentation labels $Y_{s} \subset (1,C)^{H \times W}$ ($e.g.$, stimulated images from game engines) and the target data $X_{t} \subset \mathbb{R}^{H \times W \times 3}$ without labels ($i.e.$, real images), our goal is to learn a semantic segmentation model $G$ that performs well on the target dataset $X_{t}$. Current adversarial learning methods rely heavily on discriminators to align the distributions of source and target domains via two loss functions: segmentation loss on source data and adversarial loss for alignment. 

%

However, there exists a crucial limitation for these approaches: even if perfect adaptation is achieved through a discriminator, the alignment is implemented on global level ($i.e.$, image-level), where local contextual information may be lost/deconstructed. The reason lies in that the discriminator can only implement alignment at global level, which inputs the whole map but outputs a digit to represent domain labels ($e.g.$, 0 or 1). In some cases, parts of local regions ($i.e.$, local contextual-relations) have been well aligned across domains. However, the discriminator might deconstruct this existing local alignment during implementing the global marginal distribution alignment. In this paper, we define this phenomenon as ``lack of local consistency ($i.e.$, local contextual inconsistency)", which is important to semantic segmentation in dense pixel-scale prediction.

\subsection{Overview of Network Architecture}
\begin{figure*}[t]
\centering
\subfigure {\includegraphics[width=1\linewidth]{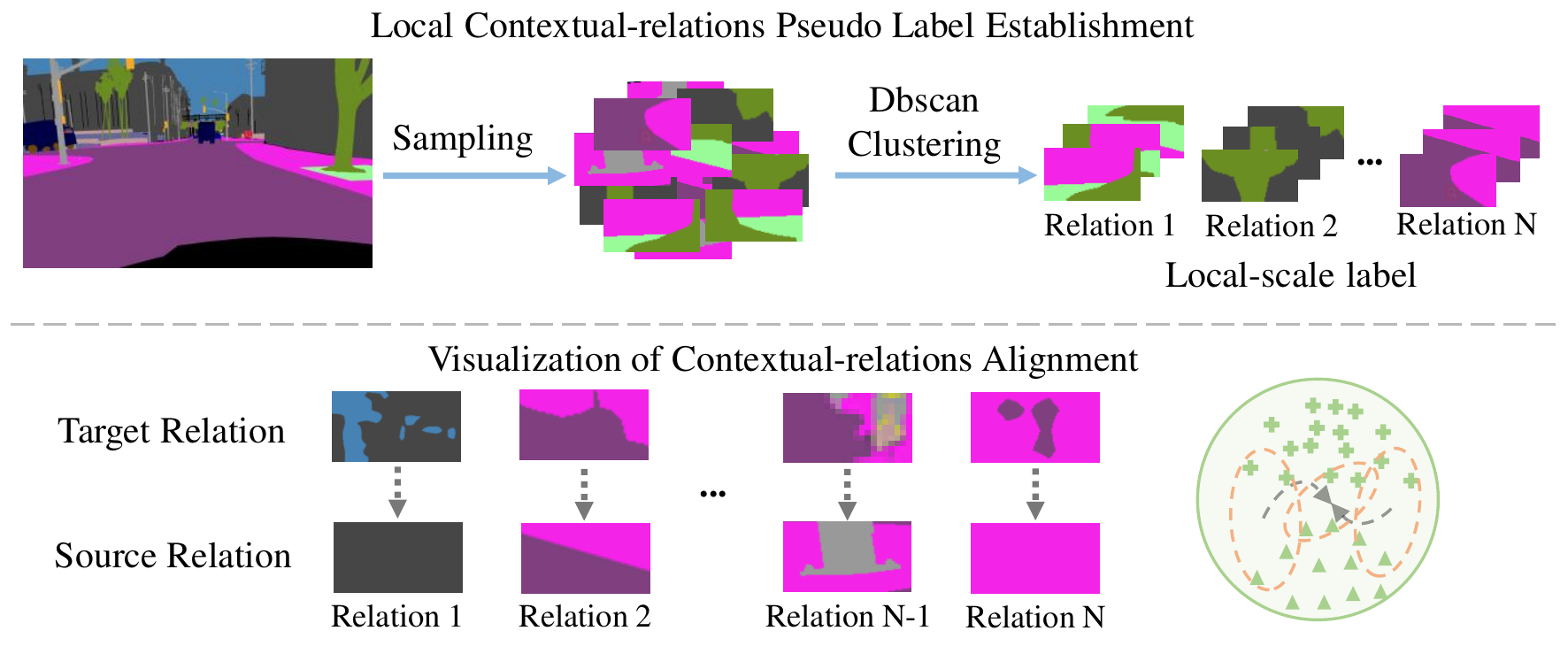}} 
\caption{Overview of local contextual-relation pseudo label establishment: \enquote{Dbscan clustering} means implementing Dbscan clustering based on the histogram of gradient. The effect of local contextual-relations alignment is shown at right-bottom part, with more visualization details provided in Fig. \ref{adapt}.}
\label{patch}
\end{figure*}

As shown in Fig. \ref{stru}, our semantic segmentation model $G$ consists of a feature extractor $E$ and two classifiers ($i.e.$, $C_{seg}$ and $C_{cr}$) where $C_{seg}$ is for pixel-scale segmentation and $C_{cr}$ is for local-scale contextual-relations learning/classification.
$E$ extracts features from input images. $C_{seg}$ and $C_{cr}$ classify features generated by $E$ into pre-defined semantic classes.
Specifically, $C_{seg}$ processes features at pixel-scale, which aims to predict pixel-scale labels. The pre-defined semantic class domain for $C_{seg}$ is the pixel-scale ground-truth, so there is no difference between $C_{seg}$ and traditional segmentation classifier.
$C_{cr}$ processes features at local scales, which aims to predict region-scale/contextual-relations labels. The pre-defined semantic class domain for $C_{cr}$ is the clustered contextual-relations ground-truth. 
The establishment procedure of clustered contextual-relations labels is described in Section 3.3 and shown in Fig. \ref{patch}. 
We train $E$ and the classifiers ($i.e.$, $C_{seg}$ and $C_{cr}$) in an adversarial scheme to reduce domain shifts at local scales to achieve local contextual-relation consistency.

\subsection{Contextual-relation consistent domain adaptation}
This subsection introduces our contextual-relation consistent domain adaptation at local scales, denoted as CrCDA$^{\ast}$, via adaptive entropy max-minimizing, as shown in Fig. \ref{stru}.

\textbf{Contextual-relation pseudo label establishment.} In order to implement local-scale task, 
we sample regions on the feature space and implement domain alignment at local scales to achieve local contextual-relation consistent domain adaptation, as shown in Fig. \ref{patch}. Different from \cite{isola2017image} that implements mode-agnostic patches alignment or \cite{tsai2019domain} that aligns patch-indexed representation of the whole image only at global scales by a discriminator ($i.e.$, the probability distributions of patch index prediction of the whole images.), we aim to aligns inter-class relations within each single patch, $i.e.$, the probability distributions of pixel class prediction within each patch, $w.r.t$ its mode via a classifier. Thus, the preliminary is to establish the region-scale label, where we first crop the pixel-scale ground-truth to many larger regions and then use Dbscan \cite{ester1996density} to cluster them to assign each region a certain index label ($i.e.$, contextual-relation pseudo label). Specifically, we assign the index label to regions according to the clustering results based on the histogram of gradient. 
For region-scale label ($i.e.$, contextual-relation pseudo label), we cluster regions into different groups based on the histogram of gradient and assign the index label.
These region-scale/contextual-relation pseudo labels can assist our network to implement alignment at local scales. Detailed information about the region-scale/contextual-relation pseudo labels is in the supplementary materials.

\begin{figure*}[!t]
\centering
\subfigure {\includegraphics[width=1\linewidth]{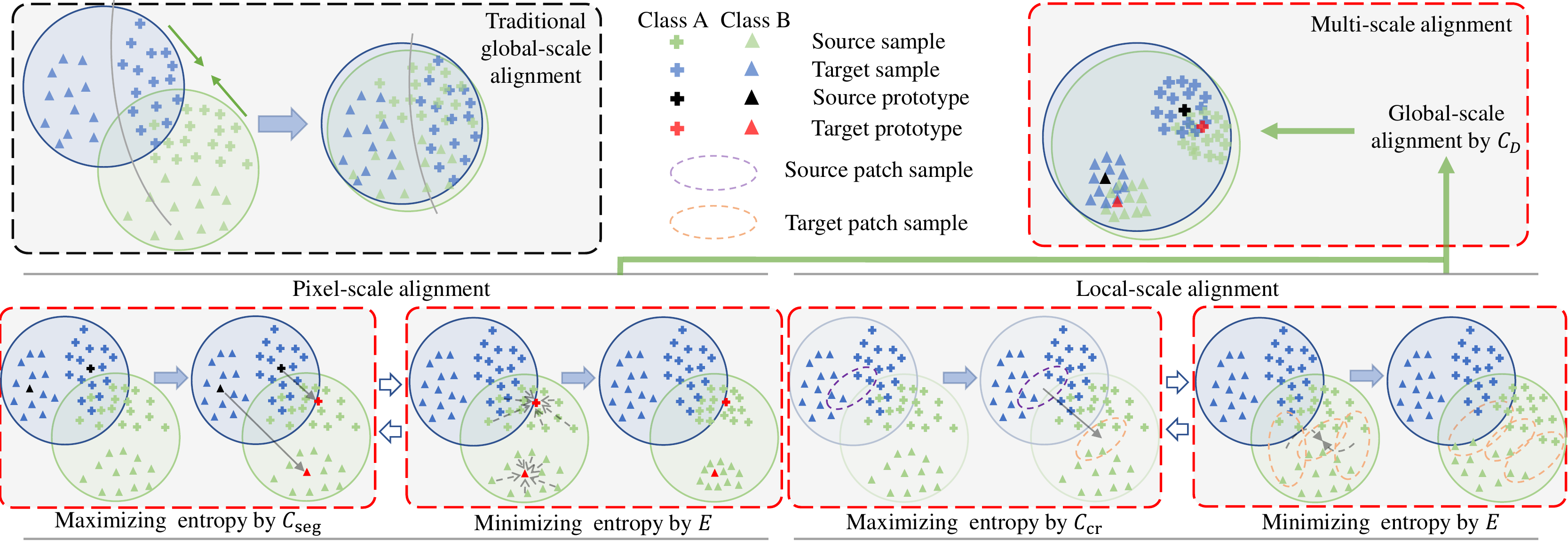}} 
\caption{Overview and comparison of the proposed AEMM at different scales: The mechanism of traditional global-scale domain adaptation is shown in the black box, where some samples are adapted into the wrong area due to the lack of local consistency ($i.e.$, local contextual-relation consistency). Our method is shown in the red boxes illustrating the alignment in pixel-scale, local-scale and global-scale. 
In pixel-scale alignment, $C_{seg}$ firstly approximates the target prototypical features by maximizing entropy on target data and then $E$ aims to congregate the features to the approximated prototypical features by minimizing entropy. Local-scale alignment works in the same scheme of pixel-scale adaptation while the only difference is the processing unit size (the former adapts a larger group of features; the latter adapts single pixel-scale features). As shown above, the global alignment is implemented by a domain classifier. Finally, the proposed AEMM can achieve feature alignment in different scales simultaneously.}
\label{adapt}
\end{figure*}

\textbf{Adaptive entropy max-minimizing adversarial learning scheme.} In local-scale adaptation, $C_{cr}$ aims to approximate the prototypical feature representations for each contextual-relation ($e.g.$, road-sidewalk, sky-building, pole-sidewalk, etc.) by implementing entropy maximization in target domain according to the source prototypical feature representations found via supervised learning in source domain. $E$ focuses on extracting discriminative feature representations (near the approximated prototypical feature representations) by implementing entropy minimization. Specifically, the prototypical feature representations of source domain found with supervision are first utilized to estimate the prototypical feature representations for target data by entropy maximizing $w.r.t$ $C_{cr}$. $E$ then adapts the extracted feature representations to the corresponding prototypical feature representations by minimizing the entropy. The overall unsupervised domain adaptation at local scales is achieved by the adversarial training between $C_{cr}$ and $E$ as illustrated in Fig. \ref{adapt}. Different from that applied in semi-supervised learning \cite{saito2019semi}, our unsupervised domain adaptation training method, referred as adaptive entropy max-minimizing (AEMM) implements entropy max-min with a regularizer $\mathcal{R}(P)=ave\{ P\log P\} \times \lambda _{R}$ ($\lambda _{R}$ decreases with training iteration, details are shown in appendix) for better estimating the prototypes in the target domain where no labels are available.

\textbf{Source Flow.} In our local-scale adaptation setting, the source data contributes to $L_{seg}$ and $L_{cr}$. Given a source image $x_{s} \subset X_{s}$, its corresponding segmentation label $y_{s} \subset Y_{s}$ and contextual-relation pseudo-label $y_{s\_cr} \subset Y_{s\_cr}$, $P_{s}^{(h, w, c)} = C_{seg}(E(x_{s}))$ is the predicted probability map $w.r.t$ each pixel over C classes; $P_{s\_cr}^{(i, j, n)} = C_{cr}(E(x_{s}))$ is the predicted probability map $w.r.t$ each region over N pre-defined contextual-relations classes. Therefore, it is a simple supervised learning objective to 
minimize $L_{seg}$ and $L_{cr}$, which are expressed as:

\begin{equation}
\mathcal{L}_{seg}(E, C_{seg}) = \sum_{h, w} \sum_{c} -y_{s}^{(h, w, c)} \log P_{s}^{(h, w, c)}
\end{equation}

\begin{equation}
\mathcal{L}_{cr}(E, C_{cr}) = \sum_{i, j} \sum_{n} -y_{s\_cr}^{(i, j, n)} \log P_{s\_cr}^{(i, j, n)}
\end{equation}

\textbf{Target Flow.} As the target label is not accessible, we introduce the adversarial training scheme between feature extractor $E$ and classifier $C_{cr}$ to extract discriminative features for target data via adaptively max-minimizing entropy in target domain. Given a target image $x_{t} \subset X_{t}$, $P_{t\_cr}^{(i, j, n)} = C_{cr}(E(x_{t}))$ is the predicted probability map $w.r.t$ each region over N pre-defined contextual-relations classes. The entropy loss $L_{ent_cr}$ is expressed as:

\begin{equation}
\mathcal{L}_{ent\_cr}(E, C_{cr}) = - \frac{1}{C}\sum_{i, j} \sum_{n} max\{P_{t\_cr}^{(i, j, n)} \log P_{t\_cr}^{(i, j, n)} - \mathcal{R}(P_{t\_cr}^{(i, j, n)}), 0\}
\end{equation}

For local-scale adaptation, we use the same back-propagation optimizing scheme with the gradient reverse layer mentioned in [57]. The training objective can be express as:

\begin{equation}
\begin{split}
& \min_{\theta_{E}} \mathcal{L}_{seg} + \lambda_{cr} \mathcal{L}_{cr} + \lambda_{ent} \mathcal{L}_{ent\_cr} \\
& \min_{\theta_{C_{seg}}} \mathcal{L}_{seg} \\
& \min_{\theta_{C_{cr}}} \mathcal{L}_{cr} - \lambda_{ent} \mathcal{L}_{ent\_cr} \\
\end{split}
\end{equation}
where $\lambda_{ent}$ is a weight factor to control the balance of unsupervised adaptation on target data and supervised learning on source data.


\subsection{CrCDA with pixel-/global-scale}
This subsection introduces our CrCDA with pixel-/global-scale, denoted as CrCDA, via adaptive entropy max-minimizing, as shown in Fig. \ref{stru}.
Our discriminator-free AEMM adversarial training scheme can also be extended into pixel-scale and global/image-scale to form multi-scale domain adaptation.

In multi-scale adaptation, for $\mathcal{L}_{seg}$, $\mathcal{L}_{cr}$ and $\mathcal{L}_{ent\_cr}$, the objectives are the same as that in local-scale adaptation. We extend the AEMM adversarial training scheme mentioned before into pixel-scale and global-scale adaptation. For pixel-scale adaptation, we implement pixel-scale entropy loss $\mathcal{L}_{ent}$ on target data to $E$ and $C_{seg}$. For global-scale adaptation, we implement global-scale entropy loss $\mathcal{L}_{ent\_D}$ on target data to $E$ and $C_{D}$, where $C_{D}$ is a domain classifier.
$C_{D}$ takes the layout probability map concatenated by the two probability maps generated from $C_{seg}$ and $C_{cr}$ as input, and predicts domain label for it (\textit{e.g.}, 0 for source domain, 1 for target domain). The global-alignment is achieved by the adversarial training between $C_{D}$ and $(E, C_{seg}, C_{cr})$. Finally, our multi-scale consistent domain adaptation network is able to align domain shift at global scales, local-scale and pixel-scale simultaneously.

Similar to local-scale adaptation, we formulate the pixel-scale entropy loss as:
\begin{equation}
\mathcal{L}_{ent\_pix}(E, C_{seg}) = - \frac{1}{C}\sum_{h, w} \sum_{c} max\{P_{t\_pix}^{(h, w, c)} \log P_{t\_pix}^{(h, w, c)}-\mathcal{R}(P_{t\_pix}^{(h, w, c)}), 0\}
\end{equation}

For multi-scale adaptation, we also use the same back-propagation optimizing scheme with the gradient reverse layer mentioned in \cite{ganin2014unsupervised,ganin2016domain}. The training objective can be express as:

\begin{equation}
\begin{split}
& \min_{\theta_{E}} \mathcal{L}_{seg} + \lambda_{cr}\mathcal{L}_{C_{cr}} + \lambda_{ent} (\mathcal{L}_{ent\_pix}+\mathcal{L}_{ent\_cr}) + \lambda_{D} \mathcal{L}_{D} \\
& \min_{\theta_{C_{seg}}} \mathcal{L}_{seg} - \lambda_{ent} \mathcal{L}_{ent\_pix} + \lambda_{D} \mathcal{L}_{D}\\
& \min_{\theta_{C_{cr}}} \mathcal{L}_{C_{cr}} - \lambda_{ent} \mathcal{L}_{ent\_cr} + \lambda_{D} \mathcal{L}_{D}\\
& \max_{\theta_{C_{D}}} \lambda_{D} \mathcal{L}_{D}\\
\end{split}
\end{equation}
where $\mathcal{L}_{D}$ is provided in supplementary materials; $\lambda_{cr}$, $\lambda_{ent}$ and $\lambda_{D}$ are the weight factor to balance the unsupervised adaptation on target data and the task-specific objectives on source data.




\section{Experiments}

\subsection{Datasets}
We evaluate our unsupervised domain adaptation networks for semantic segmentation on two challenging synthesized-to-real tasks: GTA5 \cite{richter2016playing} $\rightarrow$ Cityscapes \cite{cordts2016cityscapes} and SYNTHIA \cite{ros2016synthia} $\rightarrow$ Cityscapes. GTA5 contains $24,966$ synthesized images with high-resolution and we use the 19 common categories between GTA5 and Cityscapes in the same setting as in \cite{tsai2018learning}. SYNTHIA contains $9,400$ synthetic images with 16 common categories in Cityscapes. We use either GTA5 or SYNTHIA as source domain. We use the unlabelled training set of Cityscapes as target domain, which includes $2975$ real-world images. 

\subsection{Implementation Details}
For a fair comparison, similar to \cite{vu2019advent} \cite{luo2019taking} \cite{tsai2018learning}, we utilize Deeplab-V2 architecture \cite{chen2017deeplab} with ResNet-101 pretrained on ImageNet \cite{deng2009imagenet} as our single-scale semantic segmentation network $(E + C_{seg})$. 
To extend our model to multi-scale network, we simply copy and modify $C_{seg}$ to create $C_{cr}$ and $C_{D}$ with different output channels ($e.g.$,$N$ and $1$) and different output sizes due to various scales ($i.e.$, region-size and global-size). We also apply our methods on VGG-16 \cite{simonyan2014very} in the same way as employing ResNet-101. Following \cite{ganin2014unsupervised} \cite{tzeng2014deep}, a gradient reverse layer is employed to reverse the entropy loss between $E$ and ($C_{seg}, C_{cr}$) during pixel-/region-scale adaptation to achieve adversarial training. The domain classifier $C_{D}$ works similar to a discriminator for global-scale alignment. During training, we utilize SGD \cite{bottou2010large} to optimize our networks with a momentum of $0.9$ and a weight decay of $1e-4$. The initial learning rate is set as $2.5e-4$ and decayed by a polynomial policy with a power of $0.9$, as illustrated in \cite{chen2017deeplab}. For all experiments, the hyper-parameters $\lambda_{ent}$, $\lambda_{D}$, $\lambda_{cr}$ and $N$ are set as $2.5e-5$, $2.5e-5$, $5e-3$ and $100$, respectively. 

\renewcommand\arraystretch{1.2}
\begin{table*}[!t]
\caption{Results of domain adaptation task GTA5 $\rightarrow$ Cityscapes. \enquote{V} means the VGG16-based model and \enquote{R} means the ResNet101-based model.}
\centering
\begin{tiny}
\begin{tabular}{p{1.5cm}|p{0.2cm}|*{20}{p{0.4cm}}}
 \hline
 \hline
 \multicolumn{22}{c}{\textbf{GTA5 $\rightarrow$  Cityscapes}} \\[0.05cm]
 \hline
 Networks &\rot{Oracle}     &\mcrot{1}{l}{60}{road}     &\mcrot{1}{l}{60}{side.}     &\mcrot{1}{l}{60}{build.}    &\mcrot{1}{l}{60}{wall}     &\mcrot{1}{l}{60}{fence}     &\mcrot{1}{l}{60}{pole}     &\mcrot{1}{l}{60}{light}     &\mcrot{1}{l}{60}{sign}     &\mcrot{1}{l}{60}{vege.}     &\mcrot{1}{l}{60}{terr.}     &\mcrot{1}{l}{60}{sky}     &\mcrot{1}{l}{60}{pers.}     &\mcrot{1}{l}{60}{rider}     &\mcrot{1}{l}{60}{car}     &\mcrot{1}{l}{60}{truck}     &\mcrot{1}{l}{60}{bus}     &\mcrot{1}{l}{60}{train}     &\mcrot{1}{l}{60}{motor}     &\mcrot{1}{l}{60}{bike}     &mIoU\\
 \hline
 FCN Wild \cite{hoffman2016fcns} &V &70.4 &32.4 &62.1 &14.9 &5.4 &10.9 &14.2 &2.7 &79.2 &21.3 &64.6 &44.1 &4.2 &70.4 &8.0 &7.3 &0.0 &3.5 &0.0 &27.1\\
 CDA \cite{zhang2017curriculum} &V &74.9  &22.0  &71.7   &6.0   &11.9   &8.4   &16.3  &11.1  &75.7  &13.3  &66.5  &38.0   &9.3   &55.2  &18.8  &18.9  &0.0  &16.8  &14.6    &28.9\\
 CyCADA \cite{hoffman2017cycada} &V &83.5 &\textbf{38.3} &76.4 &20.6 &16.5 &22.2 &26.2 &\textbf{21.9} &80.4 &28.7 &65.7 &49.4 &4.2 &74.6 &16.0 &26.6 &2.0 &8.0 &0.0 &34.8\\
 AdaptSeg \cite{tsai2018learning} &V &87.3 &29.8 &78.6 &21.1 &18.2 &22.5 &21.5 &11.0 &79.7 &29.6 &71.3 &46.8 &6.5 &80.1 &23.0 &26.9 &0.0 &10.6 &0.3 &35.0\\
 CBST \cite{zou2018unsupervised} &V &66.7 &26.8 &73.7 &14.8 &9.5 &\textbf{28.3} &\textbf{25.9} &10.1 &75.5 &15.7 &51.6 &47.2 &6.2 &71.9 &3.7 &2.2 &5.4 &\textbf{18.9} &\textbf{32.4} &30.9\\
 CLAN \cite{luo2019taking} &V &\textbf{88.0} &30.6 &79.2 &23.4 &20.5 &26.1 &23.0 &14.8 &\textbf{81.6} &\textbf{34.5} &72.0 &45.8 &7.9 &80.5 &\textbf{26.6} &\textbf{29.9} &0.0 &10.7 &0.0 &36.6\\
 AdvEnt \cite{vu2019advent} &V &86.9 &28.7 &78.7 &28.5 &\textbf{25.2} &17.1 &20.3 &10.9 &80.0 &26.4 &70.2 &47.1 &8.4 &81.5 &26.0 &17.2 &\textbf{18.9} &11.7 &1.6 &36.1\\
PatAlign \cite{tsai2019domain} &V &87.3 &35.7 &79.5 &\textbf{32.0} &14.5 &21.5 &24.8 &13.7 &80.4 &32.0 &70.5 &50.5 &16.9 &81.0 &20.8 &28.1 &4.1 &15.5 &4.1 &37.5\\
CrCDA (ours) &V &86.8	&37.5	&\textbf{80.4}	&30.7	&18.1	&26.8	&25.3	&15.1	&81.5	&30.9	&\textbf{72.1}	&\textbf{52.8}	&\textbf{19.0}	&\textbf{82.1}	&25.4	&29.2	&10.1	&15.8	&3.7	&\textbf{39.1}
\\
\hline
AdaptSeg \cite{tsai2018learning} &R &86.5 &36.0 &79.9 &23.4 &23.3 &23.9 &35.2 &14.8 &83.4 &33.3 &75.6 &58.5 &27.6 &73.7 &32.5 &35.4 &3.9 &30.1 &28.1 &42.4\\
CBST \cite{zou2018unsupervised} &R
&91.8 &53.5 &80.5 &32.7 &21.0 &34.0 &28.9 &20.4 &83.9 &34.2 &80.9 &53.1 &24.0 &82.7 &30.3 &35.9 &16.0 &25.9 &\textbf{42.8} &45.9\\
 CLAN \cite{luo2019taking} &R &87.0 &27.1 &79.6 &27.3 &23.3 &28.3 &35.5 &24.2 &83.6 &27.4 &74.2 &58.6 &28.0 &76.2 &33.1 &36.7 &\textbf{6.7} &\textbf{31.9} &31.4 &43.2\\
 AdvEnt \cite{vu2019advent} &R &89.4 &33.1 &81.0 &26.6 &26.8 &27.2 &33.5 &24.7 &83.9 &36.7 &78.8 &58.7 &30.5 &\textbf{84.8} &38.5 &44.5 &1.7 &31.6 &32.4 &45.5\\
MaxSquare\cite{chen2019domain} &R
&89.4 &43.0 &82.1 &30.5 &21.3 &30.3 &\textbf{34.7} &24.0 &\textbf{85.3} &\textbf{39.4} &78.2 &\textbf{63.0} &22.9 &84.6 &36.4 &43.0 &5.5 &34.7 &33.5 &46.4\\
PatAlign \cite{tsai2019domain} &R
&92.3 &51.9 &82.1 &29.2 &25.1 &24.5 &33.8 &33.0 &82.4 &32.8 &82.2 &58.6 &27.2 &84.3 &33.4 &\textbf{46.3} &2.2 &29.5 &32.3 &46.5\\
CRST \cite{Zou_2019_ICCV} &R
&91.0 &\textbf{55.4} &80.0 &\textbf{33.7} &21.4 &\textbf{37.3} &32.9 &24.5 &85.0 &34.1 &80.8 &57.7 &24.6 &84.1 &27.8 &30.1 &\textbf{26.9} &26.0 &42.3 &47.1\\
CrCDA (ours) &R &\textbf{92.4}	&55.3	&\textbf{82.3}	&31.2	&\textbf{29.1}	&32.5	&33.2	&\textbf{35.6}	&83.5	&34.8	&\textbf{84.2}	&58.9	&\textbf{32.2}	&84.7	&\textbf{40.6}	&46.1	&2.1	&31.1	&32.7	&\textbf{48.6}
 \\
 \hline
\end{tabular}
\end{tiny}
\label{bench1}
\end{table*}

\subsection{Comparison with state-of-art}
We compare the experimental results of our method and state-of-the-art algorithms in two \enquote{Synthetic-to-real} UDA tasks with two different architectures: VGG-16 and ResNet-101.
For \enquote{GTA5 $\rightarrow$ Cityscapes}, we present the results in Table \ref{bench1} with comparisons to the state-of-the-art domain adaptation methods \cite{vu2019advent,luo2019taking,tsai2018learning,hoffman2016fcns,zhang2017curriculum,hoffman2017cycada,zhang2018fully,zhao2017pyramid}. 
Our contextual-relation consistent domain adaptation, expressed as CrCDA, achieves comparable performance to other state-of-the-art approaches on both architectures. 
Compared to Adapt-SegMap (output space global alignment) \cite{tsai2018learning}, category-level adversarial network (output space class-wise alignment) \cite{luo2019taking} and patch-represented global alignment \cite{tsai2019domain} (patch-indexed latent space alignment), CrCDA consistently brings over $+2.1\%$ mIoU improvements on ResNet-101. We reckon this gain is from our end-to-end/concurrent multi-scale alignment, which indicates that local consistency ($i.e.$, local contextual-relation consistency) is very important as well as global consistency and they are complementary to each other. 
In Table \ref{bench2}, we present the adaptation result for the task \enquote{SYNTHIA $\rightarrow$ Cityscapes} and consistent improvements are observed $w.r.t$ state-of-the-arts. Detailed analysis is included in next subsection.

\renewcommand\arraystretch{1.2}
\begin{table*}[t]
\caption{Results of domain adaptation task SYNTHIA $\rightarrow$ Cityscapes. \enquote{V} means the VGG16-based model and \enquote{R} means the ResNet101-based model. \enquote{mIoU} and \enquote{mIoU*} are calculated over 16 and 13 classes, respectively.}
\centering
\begin{tiny}
\begin{tabular}{p{1.6cm}|p{0.2cm}|*{16}{p{0.44cm}}p{0.6cm}p{0.6cm}}
 \hline
 \hline
 \multicolumn{20}{c}{\textbf{SYNTHIA $\rightarrow$  Cityscapes}} \\[0.05cm]
 \hline
 Networks &\rot{Oracle}     &\mcrot{1}{l}{60}{road}     &\mcrot{1}{l}{60}{side.}     &\mcrot{1}{l}{60}{build.}    &\mcrot{1}{l}{60}{wall}     &\mcrot{1}{l}{60}{fence}     &\mcrot{1}{l}{60}{pole}     &\mcrot{1}{l}{60}{light}     &\mcrot{1}{l}{60}{sign}     &\mcrot{1}{l}{60}{vege.}    &\mcrot{1}{l}{60}{sky}     &\mcrot{1}{l}{60}{pers.}     &\mcrot{1}{l}{60}{rider}     &\mcrot{1}{l}{60}{car}     &\mcrot{1}{l}{60}{bus}         &\mcrot{1}{l}{60}{motor}     &\mcrot{1}{l}{60}{bike}     &mIoU  &mIoU*\\
 \hline
 FCNs Wild \cite{hoffman2016fcns} &V &11.5 &19.6 &30.8 &4.4 &0.0 &20.3 &0.1 &11.7 &42.3 &68.7 &51.2 &3.8 &54.0 &3.2 &0.2 &0.6 &20.2 &22.1\\
 CDA \cite{zhang2017curriculum} &V &65.2   &26.1   &74.9    &0.1 &0.5 &10.7 &3.7 &3.0    &76.1   &70.6   &47.1    &8.2    &43.2 &20.7 &0.7    &13.1 &29.0 &34.8\\
 AdaptSeg \cite{tsai2018learning} &V &78.9 &29.2 &75.5 &- &- &- &0.1 &4.8 &72.6 &76.7 &43.4 &8.8 &71.1 &16.0 &3.6 &8.4 &- &37.6\\
 CBST \cite{zou2018unsupervised} &V &69.6 &28.7 &69.5 &\textbf{12.1} &0.1 &\textbf{25.4} &\textbf{11.9} &\textbf{13.6} &\textbf{82.0} &\textbf{81.9} &\textbf{49.1} &14.5 &66.0 &6.6 &3.7 &\textbf{32.4} &\textbf{35.4} &36.1\\
 CLAN \cite{luo2019taking} &V &\textbf{80.4} &\textbf{30.7} &74.7 &- &- &- &1.4 &8.0 &77.1 &79.0 &46.5 &8.9 &\textbf{73.8} &18.2 &2.2 &9.9 &- &39.3\\
 AdvEnt \cite{vu2019advent} &V &67.9 &29.4 &71.9 &6.3 &0.3 &19.9 &0.6 &2.6 &74.9 &74.9 &35.4 &9.6 &67.8 &21.4 &4.1 &15.5 &31.4 &36.6\\

PatAlign \cite{tsai2019domain} &V
&72.6 &29.5 &77.2 &3.5 &0.4 &21.0 &1.4 &7.9 &73.3 &79.0 &45.7 &14.5 &69.4 &19.6 &7.4 &16.5 &33.7 &39.6\\
CrCDA (ours) &V &74.5	&30.5	&\textbf{78.6}	&6.6	&\textbf{0.7}	&21.2	&2.3	&8.4	&77.4	&79.1	&45.9	&\textbf{16.5}	&73.1	&\textbf{24.1}	&\textbf{9.6}	&14.2	&35.2	&\textbf{41.1}
\\
\hline
 AdaptSeg \cite{tsai2018learning} &R &84.3 &42.7 &77.5 &- &- &- &4.7 &7.0 &77.9 &82.5 &54.3 &21.0 &72.3 &32.2 &18.9 &32.3 &- &46.7\\
 CLAN \cite{luo2019taking} &R &81.3 &37.0 &80.1 &- &- &- &\textbf{16.1} &{13.7} &78.2 &81.5 &53.4 &21.2 &73.0 &32.9 &\textbf{22.6} &30.7 &- &47.8\\
 AdvEnt \cite{vu2019advent} &R &85.6 &42.2 &79.7 &{8.7} &0.4 &25.9 &5.4 &8.1 &{80.4} &84.1 &\textbf{57.9} &23.8 &73.3 &36.4 &14.2 &{33.0} &41.2 &48.0\\
MaxSquare\cite{chen2019domain} &R
&82.9 &40.7 &\textbf{80.3} &\textbf{10.2} &\textbf{0.8} &25.8 &12.8 &\textbf{18.2} &\textbf{82.5} &82.2 &53.1 &18.0 &\textbf{79.0} &31.4 &10.4 &\textbf{35.6} &41.4 &48.2\\
PatAlign \cite{tsai2019domain} &R &82.4 &38.0 &78.6 &8.7 &0.6 &26.0 &3.9 &11.1 &75.5 &84.6 &53.5 &21.6 &71.4 &32.6 &19.3 &31.7 &40.0 &46.5\\
CrCDA (ours) &R &\textbf{86.2}	&\textbf{44.9}	&79.5	&8.3	&{0.7}	&\textbf{27.8}	&9.4	&11.8	&78.6	&\textbf{86.5}	&57.2	&\textbf{26.1}	&{76.8}	&\textbf{39.9}	&21.5	&32.1	&\textbf{42.9}	&\textbf{50.0}
\\
\hline
\end{tabular}
\end{tiny}
\label{bench2}
\end{table*}

\subsection{Ablation Studies and Analysis}
We analyze our proposed CrCDA with several state-of-the-art baselines. In general, both single-scale form (CrCDA$^{\ast}$) and multi-scale form (CrCDA) achieve comparable results to all the baselines in all the settings. 

As shown on the first three rows in Table \ref{abla}, our pixel-scale AEMM adversarial network brings $+1.4\%$ improvements in terms of mIoU over MinEnt \cite{vu2019advent}. The reason lies in that direct entropy minimization does not take the domain gap into account while our AEMM training scheme pushes the source distribution closer to target distribution during maximizing entropy on target data. 

For our CrCDA with single-scale form (CrCDA$^{\ast}$) via AEMM, it outperforms MinEnt-based contextual-relations alignment by $+1.6\%$ on ResNet-101, as shown on the second block (row4-5) in Table \ref{abla}. We reckon that these improvements are contributed by our adaptive entropy max-min training scheme which considers the domain mismatch/gap while MinEnt neglects.

\begin{figure*}[!t]
\begin{tabular}{p{2.3cm}<{\centering} p{2.35cm}<{\centering} p{2.35cm}<{\centering} p{2.35cm}<{\centering} p{2.35cm}<{\centering}}
\raisebox{-0.5\height}{\includegraphics[width=1.02\linewidth,height=0.51\linewidth]{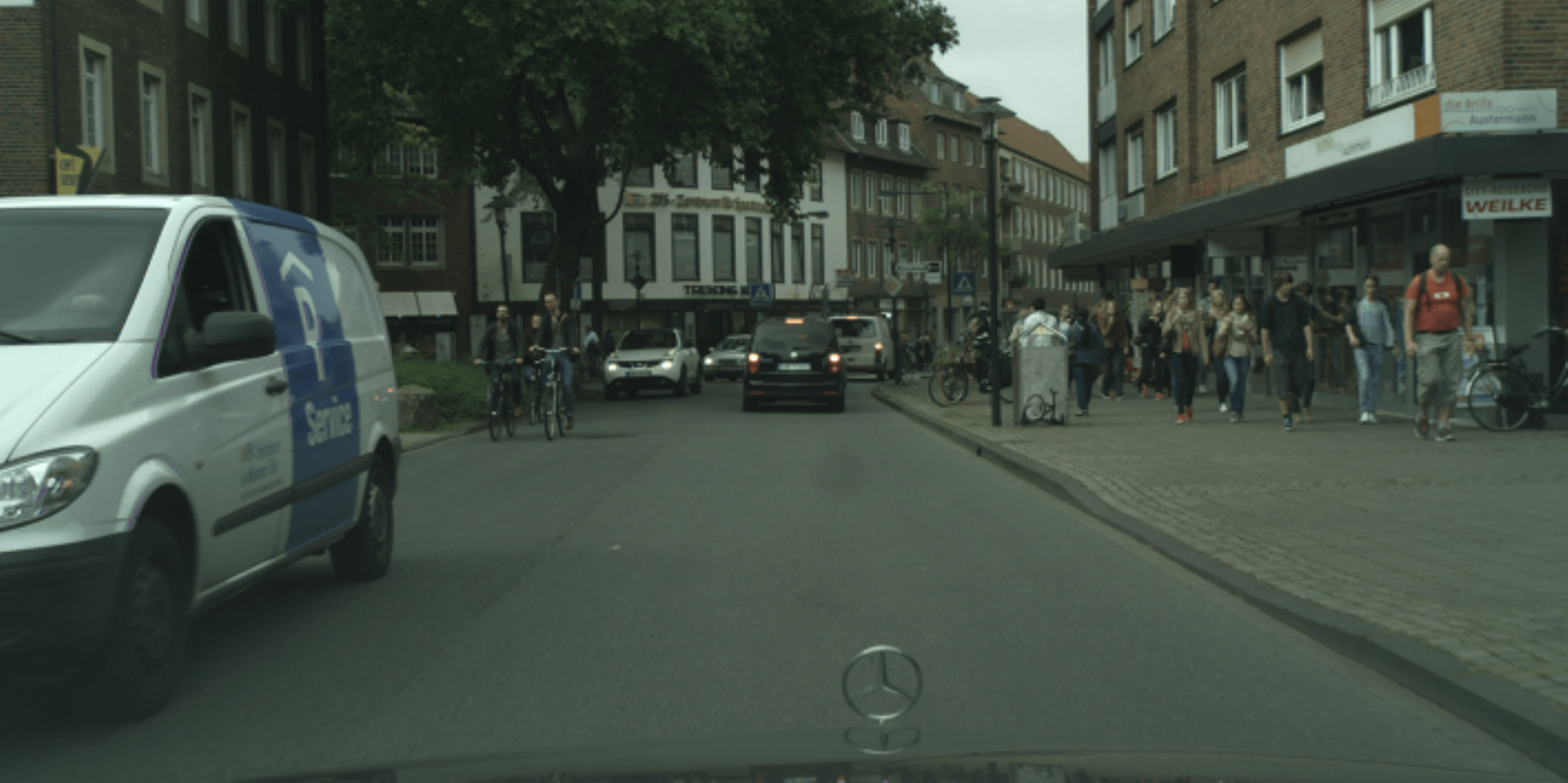}} 
 & \raisebox{-0.5\height}{\includegraphics[width=1.02\linewidth,height=0.51\linewidth]{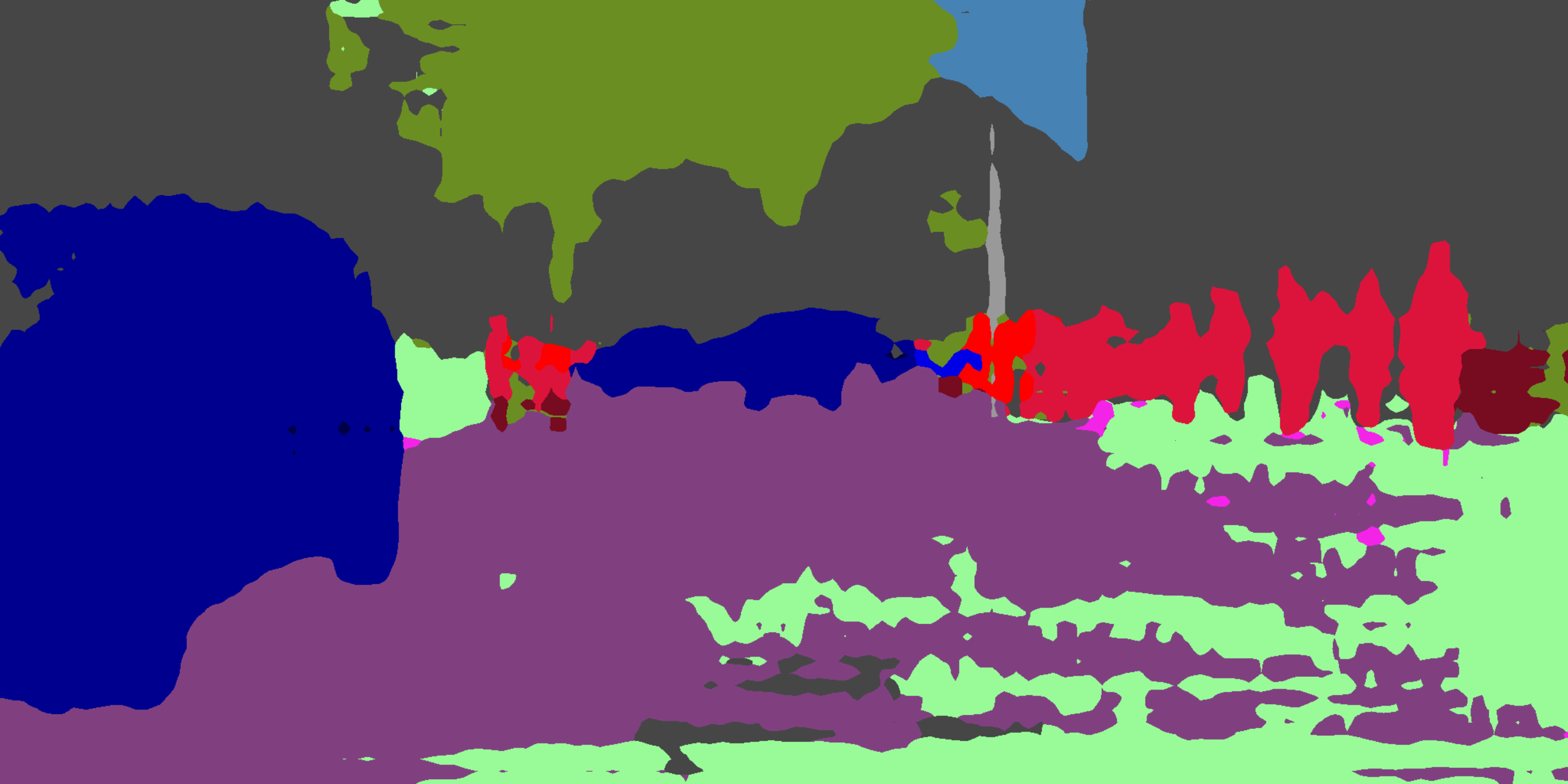}}
& \raisebox{-0.5\height}{\includegraphics[width=1.02\linewidth,height=0.51\linewidth]{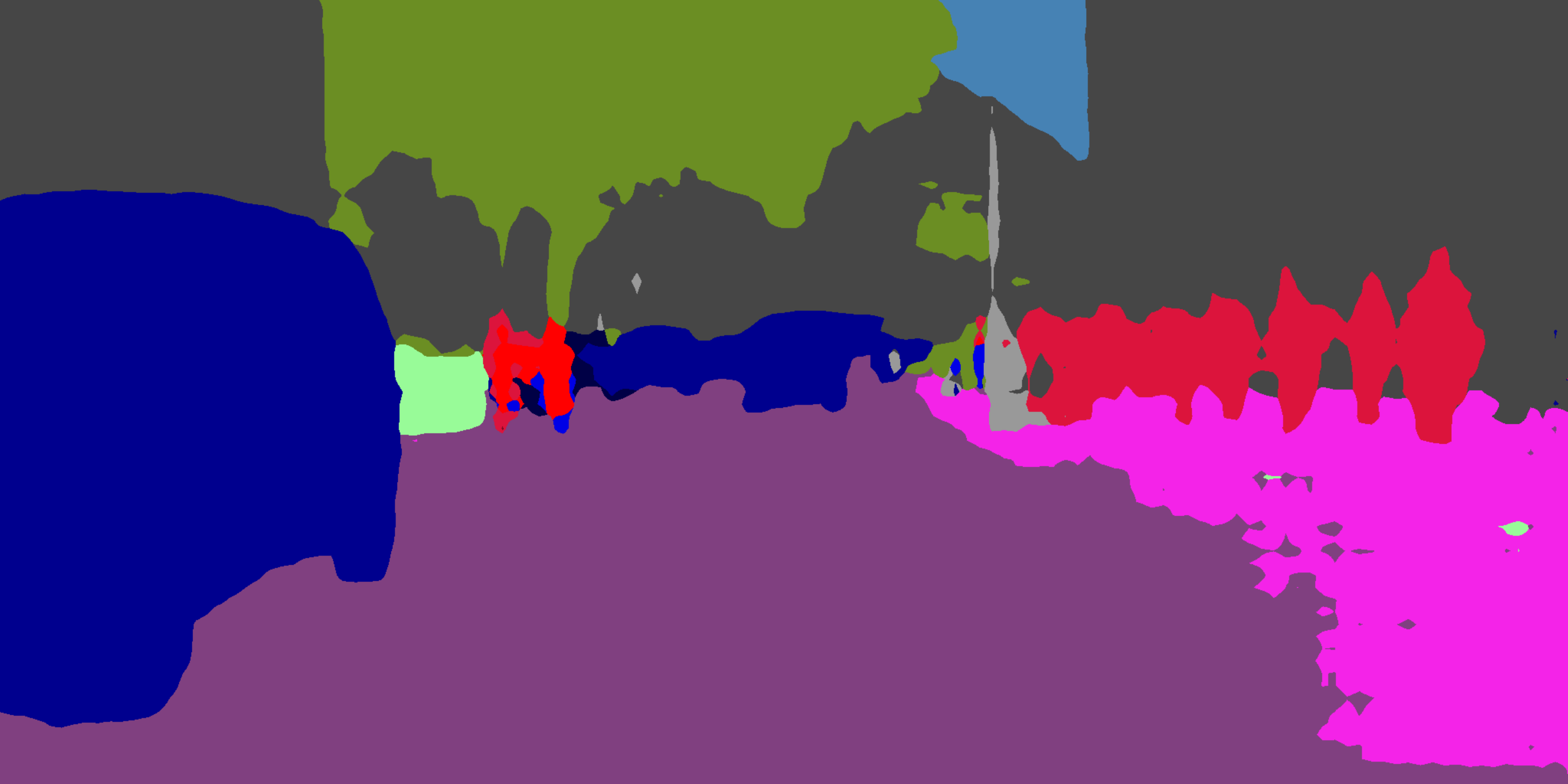}}
& \raisebox{-0.5\height}{\includegraphics[width=1.02\linewidth,height=0.51\linewidth]{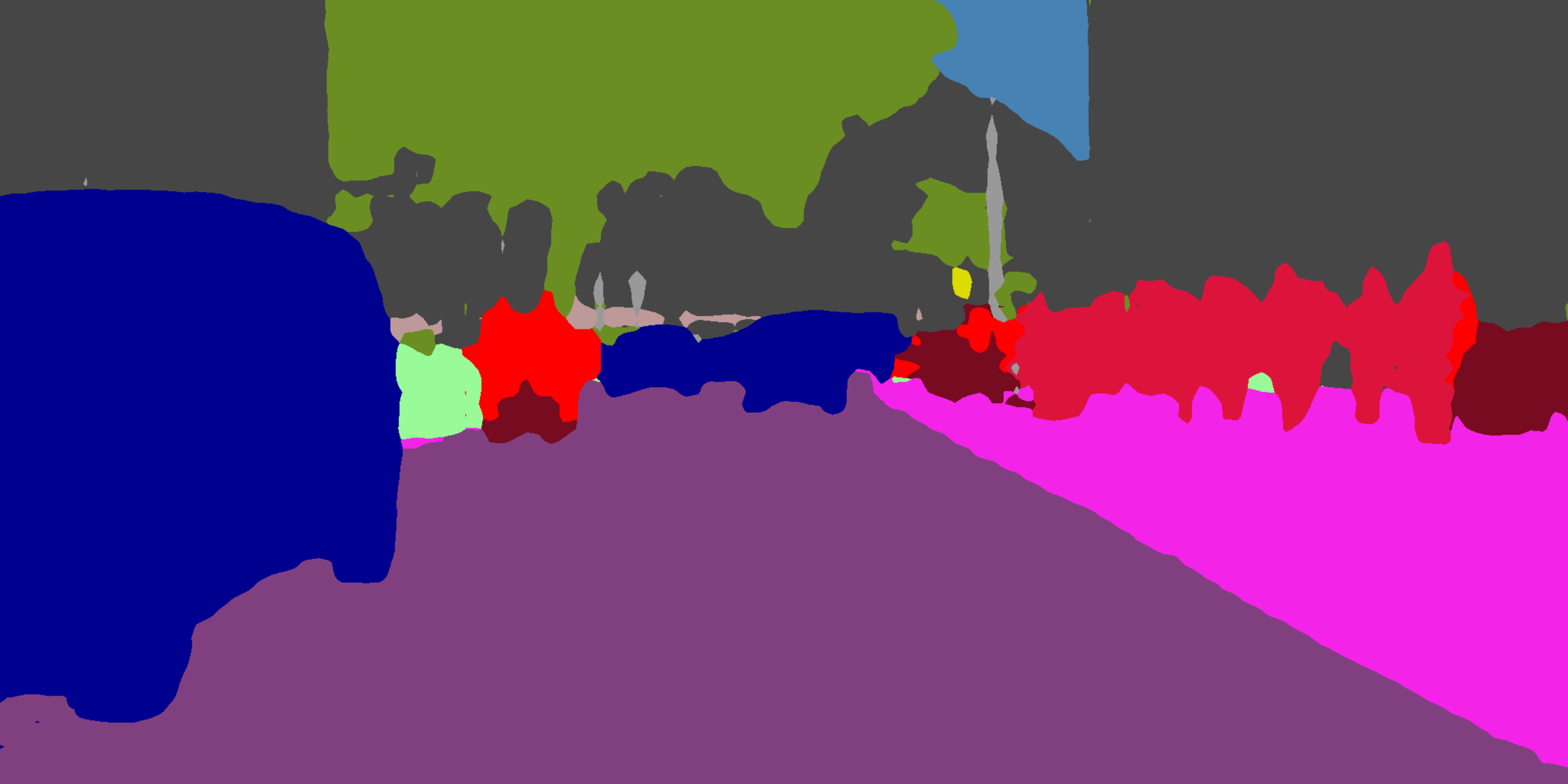}}
& \raisebox{-0.5\height}{\includegraphics[width=1.02\linewidth,height=0.51\linewidth]{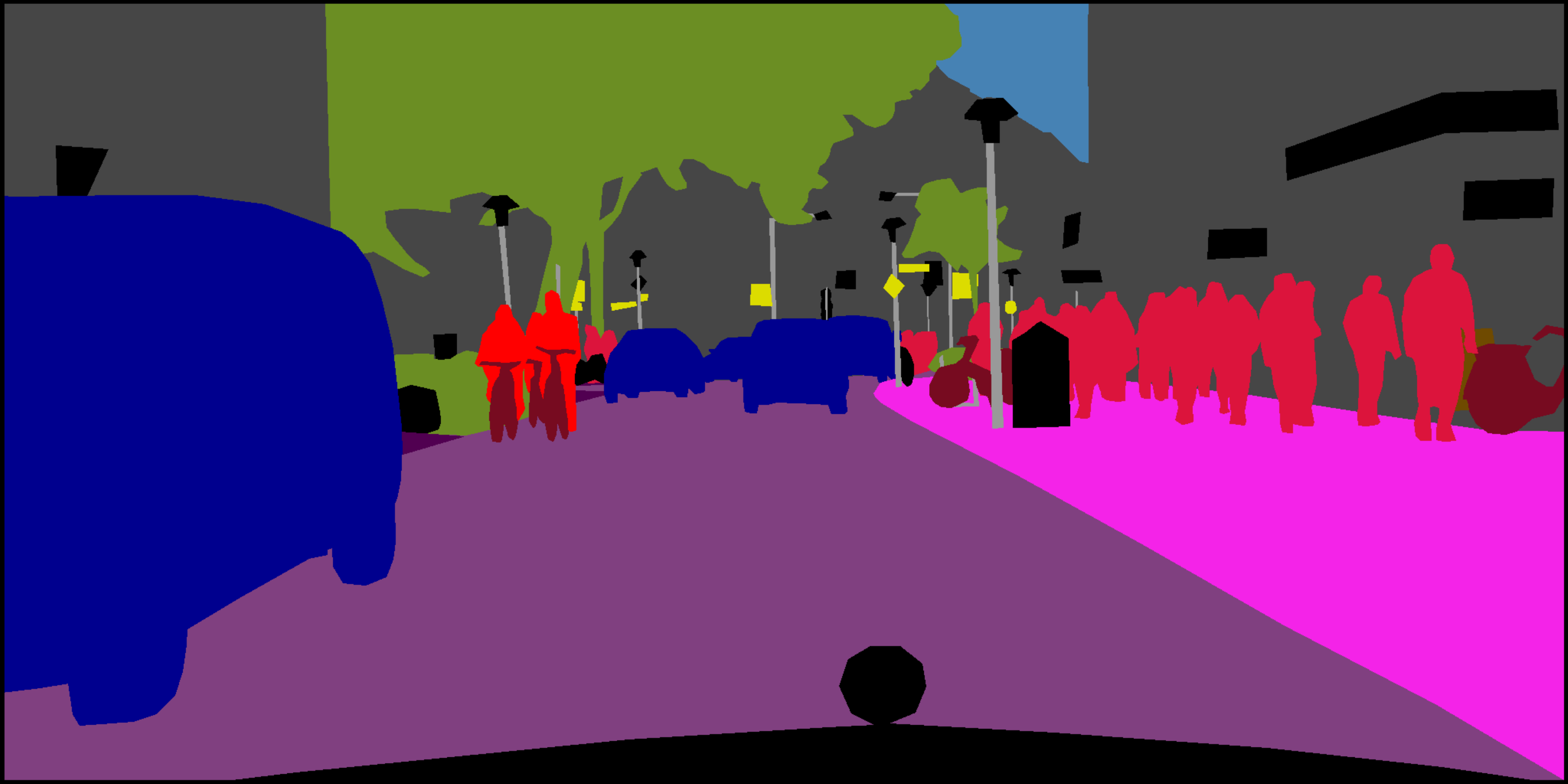}}
\\
\raisebox{-0.5\height}{\includegraphics[width=1.02\linewidth,height=0.51\linewidth]{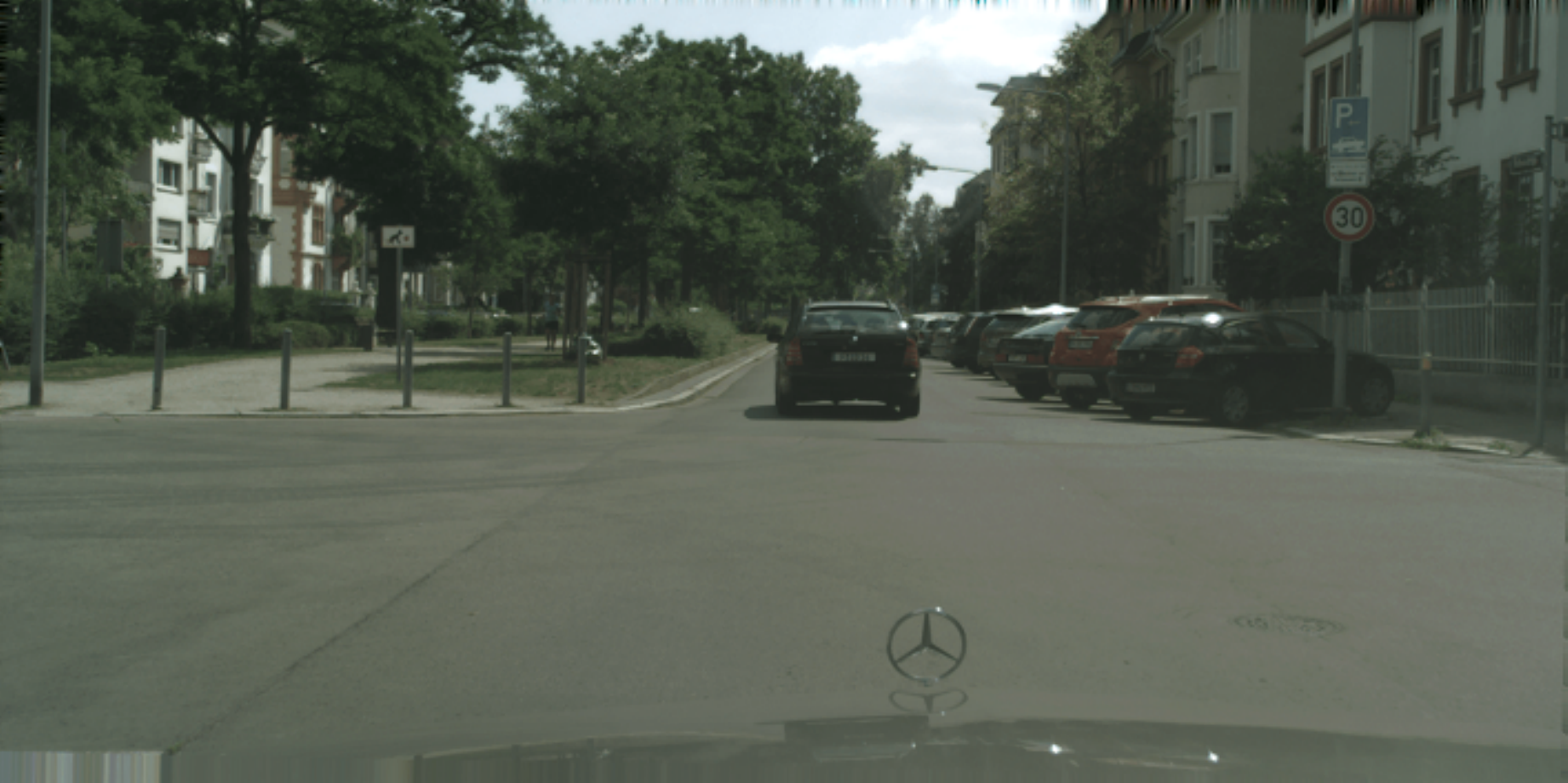}} 
 & \raisebox{-0.5\height}{\includegraphics[width=1.02\linewidth,height=0.51\linewidth]{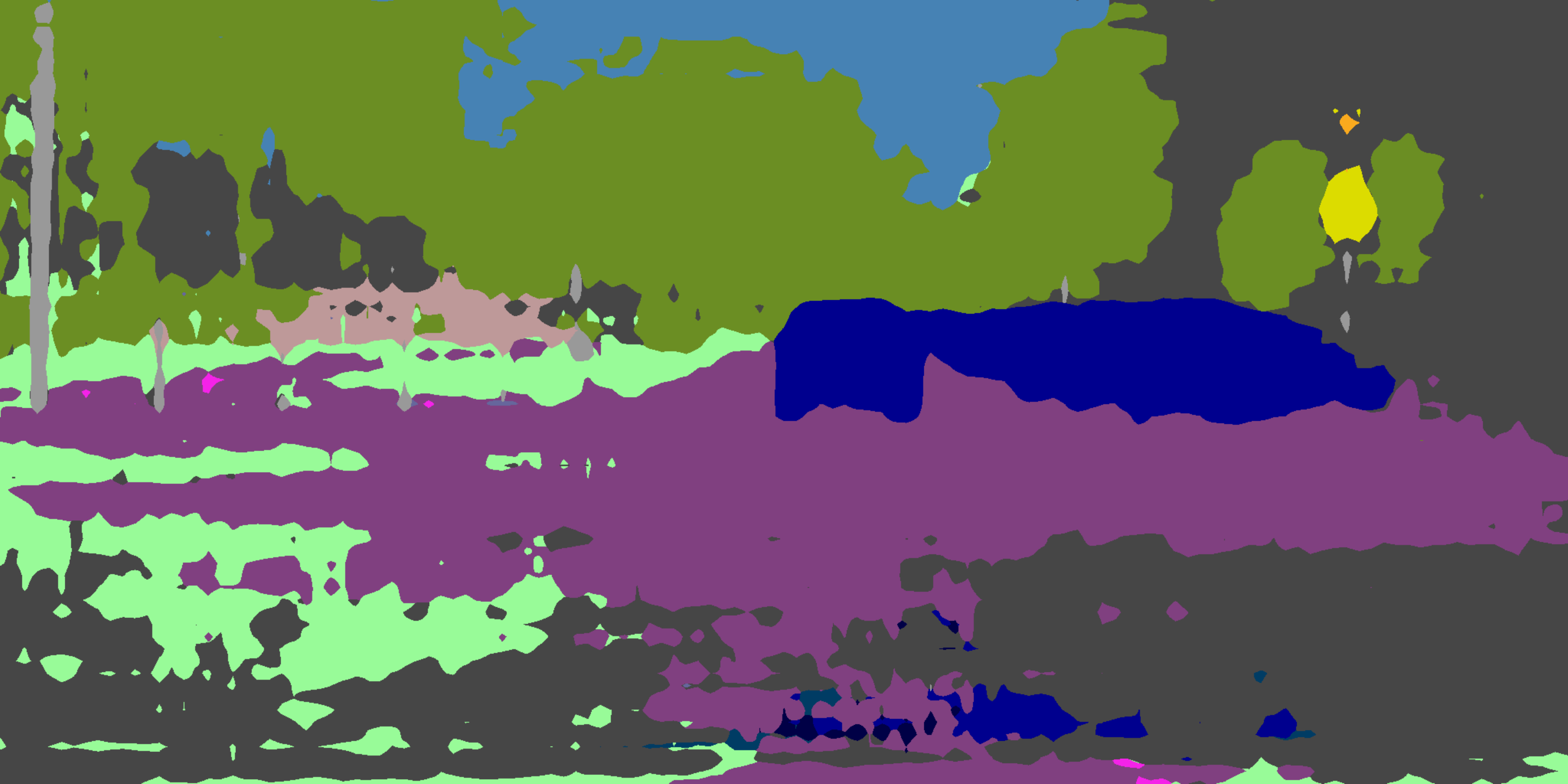}}
& \raisebox{-0.5\height}{\includegraphics[width=1.02\linewidth,height=0.51\linewidth]{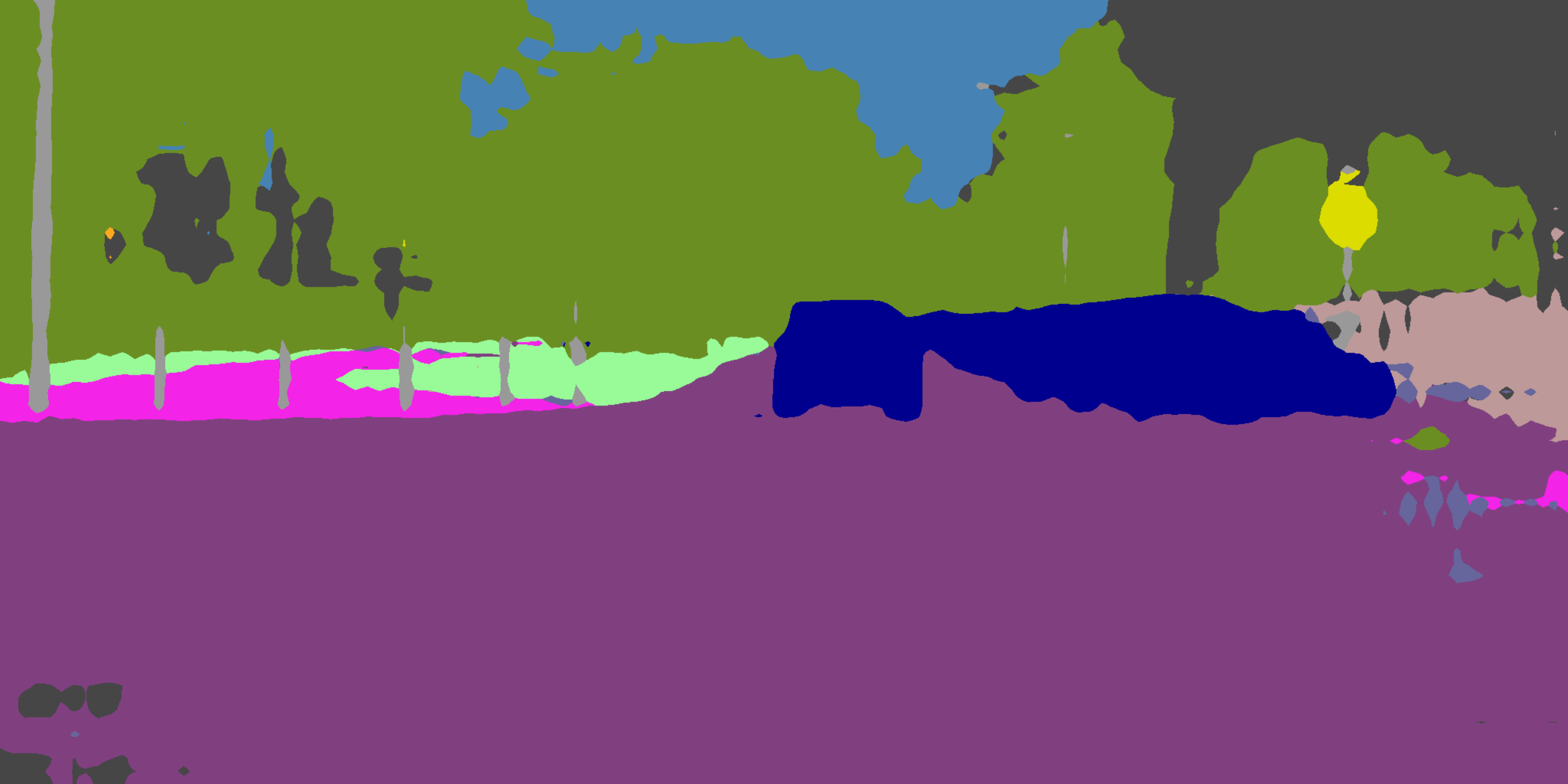}}
& \raisebox{-0.5\height}{\includegraphics[width=1.02\linewidth,height=0.51\linewidth]{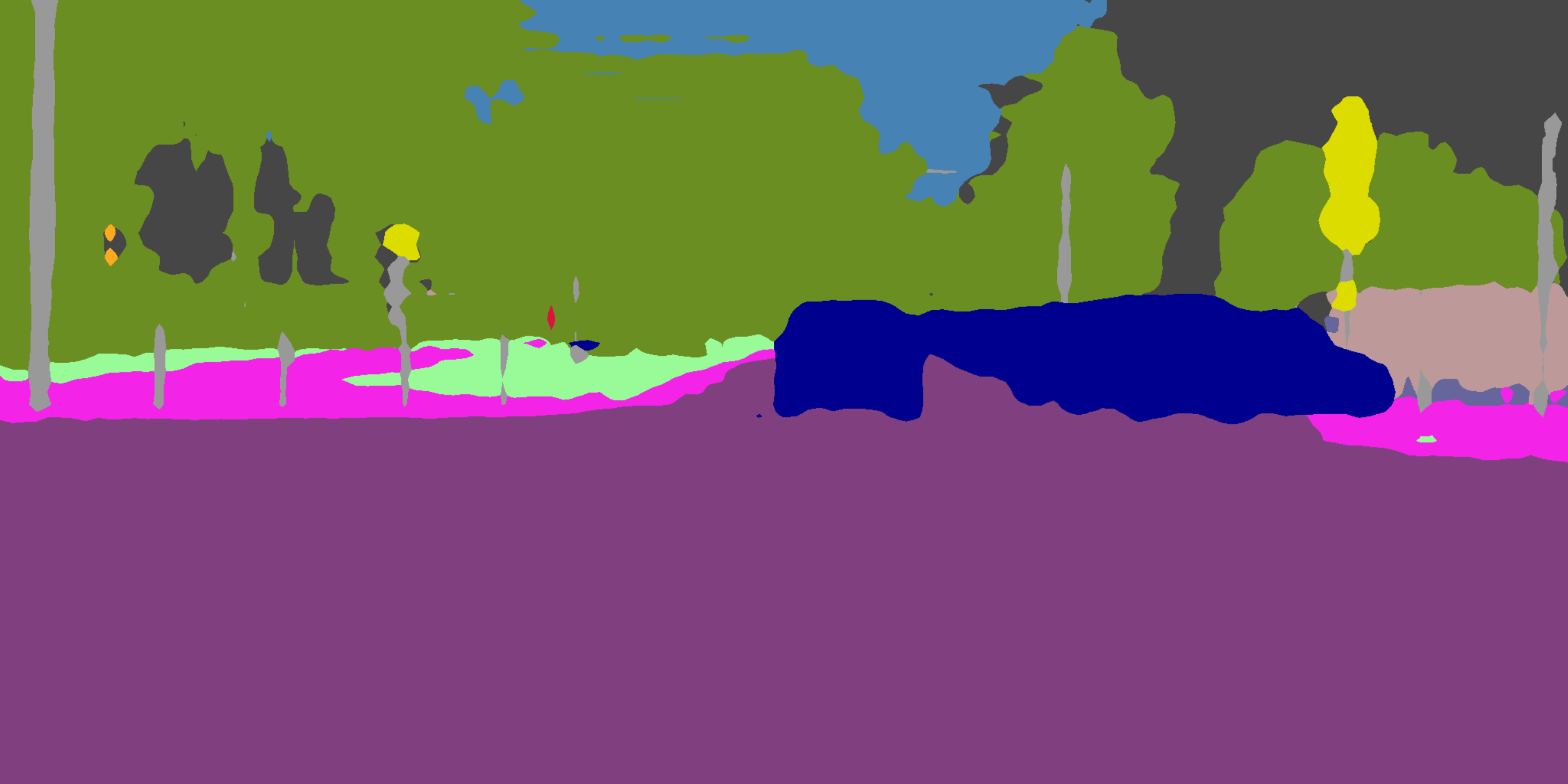}}
& \raisebox{-0.5\height}{\includegraphics[width=1.02\linewidth,height=0.51\linewidth]{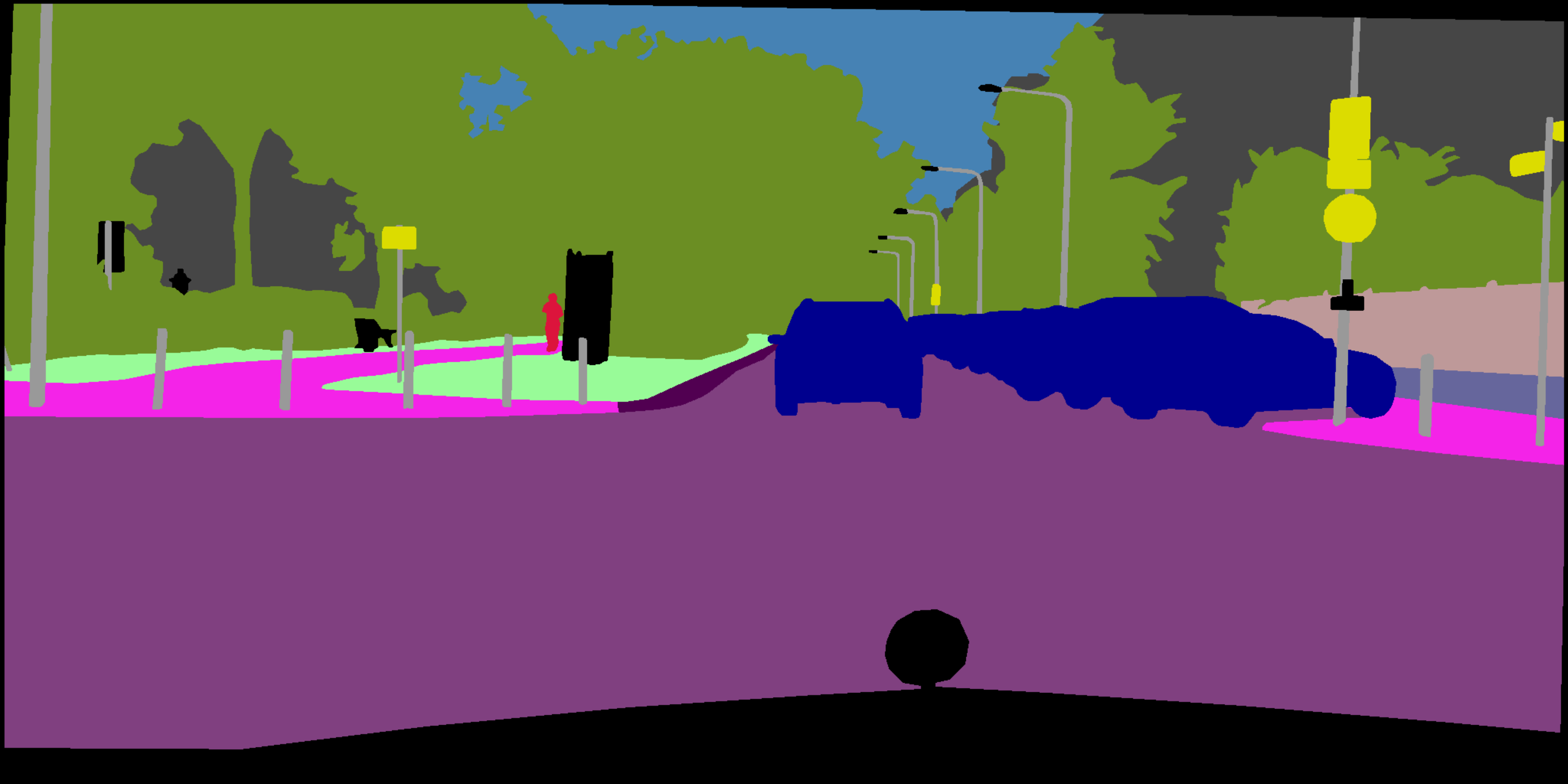}}
\\
\raisebox{-0.5\height}{\footnotesize{Target Image}}
 & \raisebox{-0.5\height}{\footnotesize{Without Ada.}}
& \raisebox{-0.5\height}{\footnotesize{Ada.(AdvEnt)}}
& \raisebox{-0.5\height}{\footnotesize{Ada.(CrCDA)}}
& \raisebox{-0.5\height}{\footnotesize{Ground Truth}}
\\

\end{tabular}
\caption{Qualitative results for GTA5 $\rightarrow$ Cityscapes. Our approach (CrCDA) aligns low-level features ($e.g.$, boundaries of sidewalk, car and person $etc.$) as well as high-level features by multi-scale adversarial learning. In contrast, AdvEnt ignores low-level information because global alignment focuses more on high-level information. Thus, as shown above, CrCDA achieves both local and global consistencies while AdvEnt only achieves global consistency.}
\label{results}
\end{figure*}

\begin{figure*}[t]
\begin{tabular}{p{4cm}<{\centering} p{4cm}<{\centering} p{4cm}<{\centering}}

\raisebox{-0.5\height}{\includegraphics[width=1\linewidth,height=0.58\linewidth]{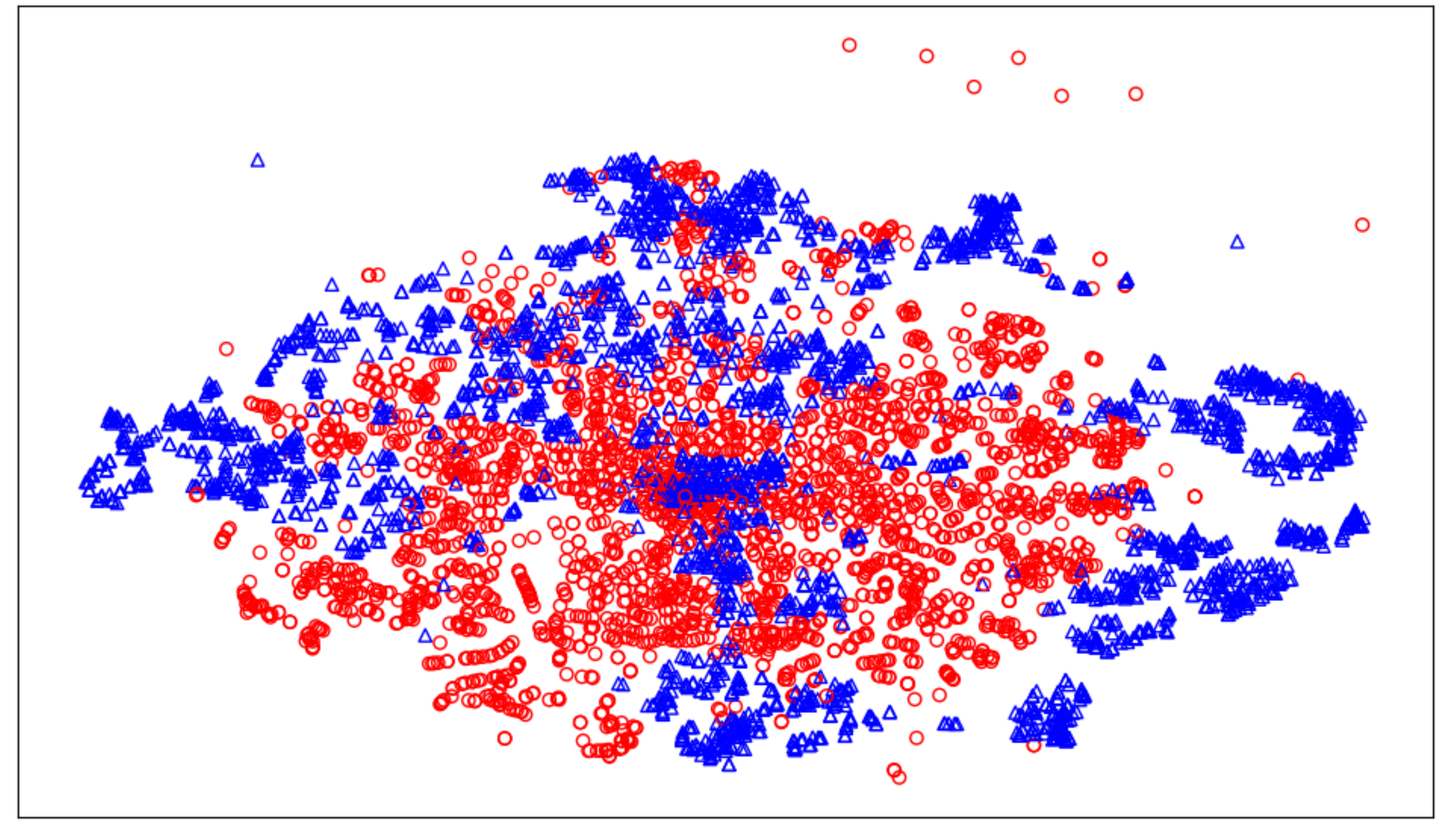}}
& \raisebox{-0.5\height}{\includegraphics[width=1\linewidth,height=0.58\linewidth]{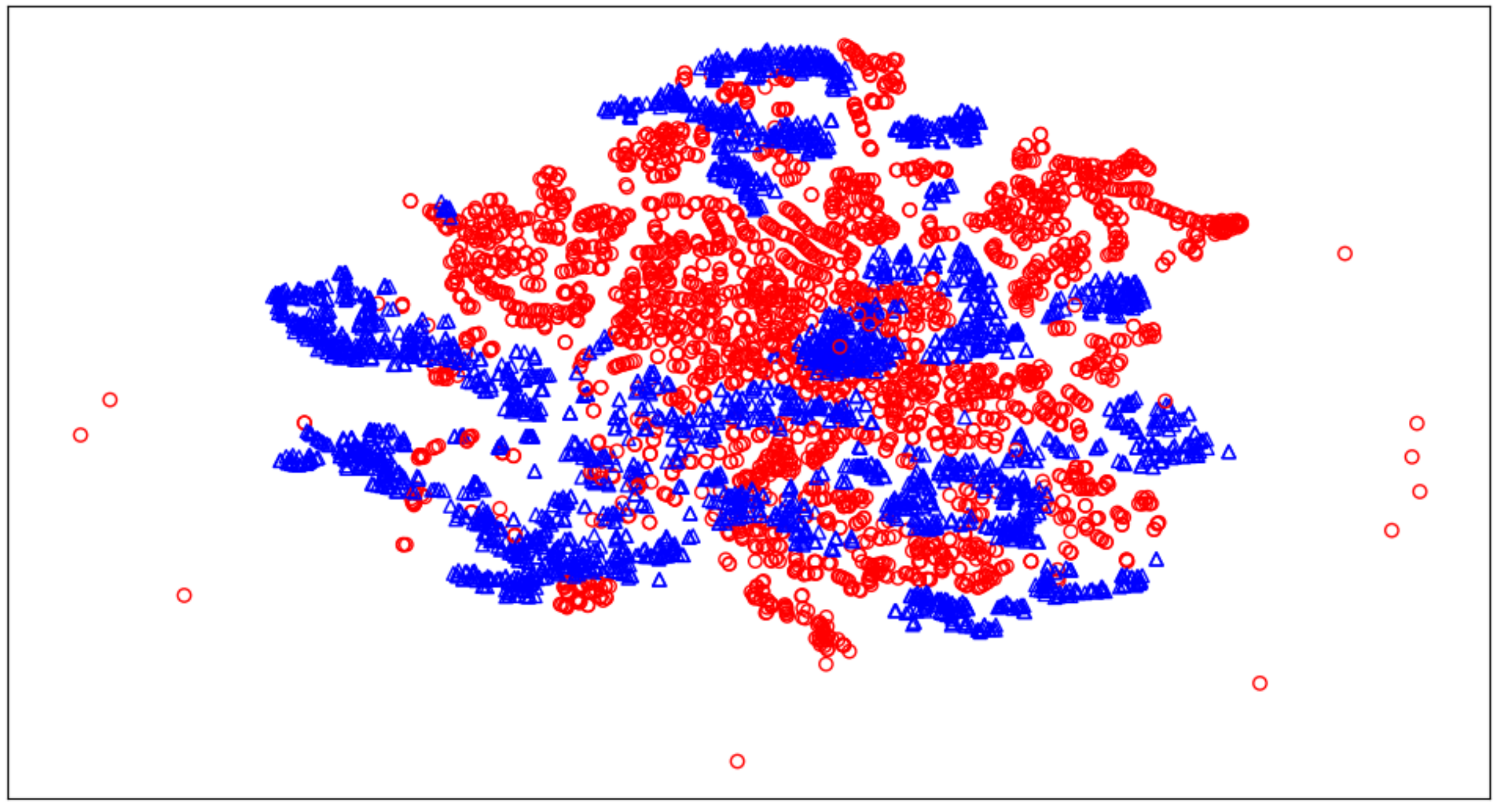}}
& \raisebox{-0.5\height}{\includegraphics[width=1\linewidth,height=0.58\linewidth]{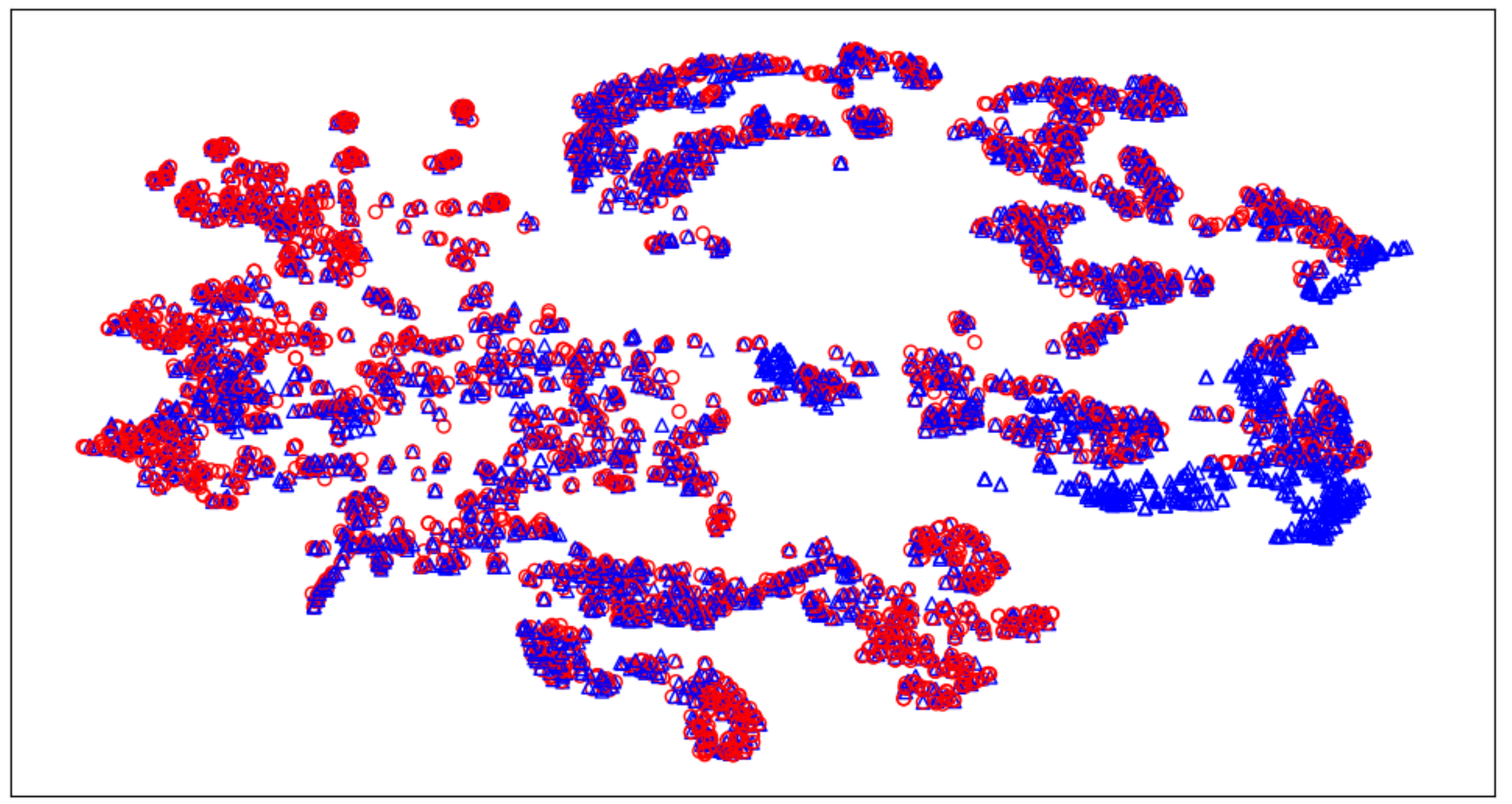}}
\\
\raisebox{-0.5\height}{Without Adaptation}
& \raisebox{-0.5\height}{Adapted(TGA)}
& \raisebox{-0.5\height}{Adapted(CrCDA)}
\\
\end{tabular}
\caption{Visualization of feature distributions via t-SNE \cite{maaten2008visualizing}. \enquote{\textcolor{blue}{Blue}}: Source. \enquote{\textcolor{red}{Red}}: Target. As shown in the first column, the feature distribution of source data is naturally more discriminative (discrete) than that of target data (uniformly distributed) due to only source supervision is available. 
Traditional global alignment (TGA) aligns them in global scale, where global consistency is achieved while local consistency is ignored. Thus the adapted target feature distribution is not discriminative. CrCDA aligns them with local-scale consistency ($i.e.$, local contextual-relation consistency), where both local and global consistencies are achieved. Thus the adapted target feature distribution is more discriminative and consistent with that of the source.}
\label{feature}
\end{figure*}

Our CrCDA with multi-scale form integrating three scales' adaptation (pixel-, local- and global-scale), termed as CrCDA shown on the bottom block in Table \ref{abla}, achieves state-of-the-art performances $48.6\%$ mIoU on ResNet-101.
Besides, CrCDA also outperforms all current methods by over $+1.5\%$. Compared to \enquote{Pixel+Global}, CrCDA brings $+2.6\%$ improvement in mIoU, which demonstrates that local-scale alignment is essential as well as other scales ($e.g.$, pixel-scale and global-scale). In fact, the local contextual-relation consistent adaptation loss ($i.e.$, $L_{ent\_cr}$) penalize groups of pixels predictions to achieve local-scale alignment, where global-scale adaptation loss operates more on image-scale ($e.g.$, scene layout) while that of pixel-scale works on the feature representation alignment of each independent pixels. The consistent results with different settings further confirm that complementary information has been learned in different scales' adaptation.
The qualitative results and visualization of feature distributions are provided in Fig. \ref{results} and \ref{feature}, which further demonstrate our conjectures mentioned above. We also provide the complementary studies to demonstrate that our local contextual-relations alignment method is complementary to most existing global-scale alignment approaches, as shown in Table \ref{comple}. 


\renewcommand\arraystretch{1.1}
\begin{table}[!t]
\caption{Ablation study of the proposed contextual-relation consistent domain adaptation on GTA5-to-Cityscapes using the ResNet-101 network.
All settings/methods are with "$L_{seg}$" (bold texts represent our methods). CrCDA$^{\ast}$ represents the contextual-relation consistent domain adaptation with only single-scale (local scale).
}
\centering
\begin{scriptsize}
\begin{tabular}{p{2.5cm}|p{1cm}p{1cm}|p{1cm}p{1cm}|p{1cm}p{1cm}p{1cm}p{1cm}|p{1.6cm}}
\hline
\hline
& \multicolumn{2}{c|}{pixel-scale Ada.}
& \multicolumn{2}{c|}{local-scale Ada.} & \multicolumn{4}{c|}{global-scale Ada.} & \multicolumn{1}{c}{mIoU}
\\\hline
Method 
& \multicolumn{1}{c}{$L_{minent}$} &
\multicolumn{1}{c|}{$L_{ent}^{ours}$} 
& \multicolumn{1}{c}{$L_{minent}$} &
\multicolumn{1}{c|}{$L_{ent}^{ours}$}
&\multicolumn{1}{c}{$L_{adv}$} & \multicolumn{1}{c}{$L_{patadv}$}
& \multicolumn{1}{c}{$L_{advent}$} & \multicolumn{1}{c|}{$L_{D}^{ours}$}
\\\hline
Without Ada. &   & &  &   &   &    &   & &\multicolumn{1}{c}{36.6}\\
MinEnt \cite{zhu2005semi,vu2019advent} &\multicolumn{1}{c}{\checkmark}   &   &  &  &   & &  &   &\multicolumn{1}{c}{42.4}\\
\textbf{Pixel-AEMM}  &   &\multicolumn{1}{c|}{\checkmark}   &    &  & &   &  & &\multicolumn{1}{c}{43.8}\\
\hline

CrCDA$^{\ast}$-MinEnt  &   &   &\multicolumn{1}{c}{\checkmark}  &     &   & &   & &\multicolumn{1}{c}{42.1}\\
\textbf{CrCDA$^{\ast}$-AEMM}  &   &   &  &\multicolumn{1}{c|}{\checkmark}   &     & &   & &\multicolumn{1}{c}{43.7}\\
\hline

AdaptSeg \cite{tsai2018learning} &   &   & &   &\multicolumn{1}{c}{\checkmark}  &   &     & &\multicolumn{1}{c}{41.4}\\
PatAlign \cite{tsai2019domain} &   &   & &   &  &\multicolumn{1}{c}{\checkmark}   &     & &\multicolumn{1}{c}{41.3}\\
AdvEnt \cite{vu2019advent} &   &   & &   &  &   &\multicolumn{1}{c}{\checkmark}     & &\multicolumn{1}{c}{43.8}\\
\textbf{Global-AEMM}  &   &   & &   &  &   &     &\multicolumn{1}{c|}{\checkmark} &\multicolumn{1}{c}{44.3}\\
\hline
\textbf{Pixel+CrCDA$^{\ast}$}  &   &\multicolumn{1}{c|}{\checkmark} &   &\multicolumn{1}{c|}{\checkmark}  &  &   &     & &\multicolumn{1}{c}{45.6}\\
\textbf{Pixel+Global}  &   &\multicolumn{1}{c|}{\checkmark} &   &  &  &   &     &\multicolumn{1}{c|}{\checkmark} &\multicolumn{1}{c}{46.0}\\
\textbf{CrCDA$^{\ast}$+Global} &   & &   &\multicolumn{1}{c|}{\checkmark}  &  &   &     &\multicolumn{1}{c|}{\checkmark} &\multicolumn{1}{c}{46.1}\\
\textbf{CrCDA}  &   &\multicolumn{1}{c|}{\checkmark}   &  &\multicolumn{1}{c|}{\checkmark}   &   &   &  &\multicolumn{1}{c|}{\checkmark} &\multicolumn{1}{c}{\textbf{48.6}}\\
\hline

\end{tabular}
\end{scriptsize}
\label{abla}
\end{table}
\renewcommand\arraystretch{1.1}
\begin{table}[t]
\caption{Complementary study of the proposed contextual-relation consistent domain adaptation with local-scale to current global alignment UDA methods on GTA5-to-Cityscapes using the ResNet-101 network.
All methods are default with "$L_{seg}$".
}
\centering
\begin{scriptsize}
\begin{tabular}{p{3.6cm}|p{1cm}|p{1cm}p{1cm}p{1cm}p{1cm}|p{1.6cm}}
\hline
\hline
& \multicolumn{1}{c|}{local-level Ada.} & \multicolumn{4}{c|}{global Ada.} & \multicolumn{1}{c}{mIoU}
\\\hline
Method 
&\multicolumn{1}{c|}{$L_{ent}^{ours}$}
&\multicolumn{1}{c}{$L_{adv}$} & \multicolumn{1}{c}{$L_{patadv}$} & \multicolumn{1}{c}{$L_{advent}$} & \multicolumn{1}{c|}{$L_{D}^{ours}$}
\\\hline
CrCDA$^{\ast}$ (ours)  &\multicolumn{1}{c|}{\checkmark}   &  &  &   & &\multicolumn{1}{c}{43.7}\\
\hline

AdaptSeg \cite{tsai2018learning}  &   &\multicolumn{1}{c}{\checkmark}  &  &   & &\multicolumn{1}{c}{41.4}\\
PatAdv \cite{tsai2019domain} &   &  &\multicolumn{1}{c}{\checkmark}  &   & &\multicolumn{1}{c}{41.3}\\
AdvEnt \cite{vu2019advent}  &   &  &  &\multicolumn{1}{c}{\checkmark}   & &\multicolumn{1}{c}{43.8}\\
GlobalAlign (ours)   &   &  &  &   &\multicolumn{1}{c|}{\checkmark} &\multicolumn{1}{c}{44.3}\\
\hline
AdaptSeg\cite{tsai2018learning}+PatAlign\cite{tsai2019domain}   &   &\multicolumn{1}{c}{\checkmark}  &\multicolumn{1}{c}{\checkmark}  &   & &\multicolumn{1}{c}{43.2}\\
CrCDA$^{\ast}$ + AdaptSeg\cite{tsai2018learning}  &\multicolumn{1}{c|}{\checkmark}   &\multicolumn{1}{c}{\checkmark}  &  &   & &\multicolumn{1}{c}{44.8}\\
CrCDA$^{\ast}$ + PatAlign\cite{tsai2019domain}   &\multicolumn{1}{c|}{\checkmark}   &  &\multicolumn{1}{c}{\checkmark}  &   & &\multicolumn{1}{c}{44.7}\\
CrCDA$^{\ast}$ + AdvEnt\cite{vu2019advent}   &\multicolumn{1}{c|}{\checkmark}   &  &  &\multicolumn{1}{c}{\checkmark}   & &\multicolumn{1}{c}{45.2}\\
CrCDA$^{\ast}$+GlobalAlign (ours)   &\multicolumn{1}{c|}{\checkmark}   &  &  &   &\multicolumn{1}{c|}{\checkmark} &\multicolumn{1}{c}{\textbf{46.1}}\\
\hline

\end{tabular}
\end{scriptsize}
\label{comple}
\end{table}

\section{Conclusions}

In this paper, we present the local contextual-relation consistent domain adaptation (CrCDA) to address the task of unsupervised domain adaptation for semantic segmentation. By taking a closer look at the local inconsistency ($i.e.$, local contextual-relations inconsistency) while implementing global adaptation, CrCDA is able to align the domain shift in local and global scales at the same time, where local semantic consistency is normally ignored by current approaches. 
The experimental results on the two challenging segmentation UDA tasks validate the state-of-the-art of CrCDA.

\section{Acknowledgement}
This research was conducted in collaboration with Singapore Telecommunications Limited and partially supported by the Singapore Government through the Industry Alignment Fund - Industry Collaboration Projects Grant.

\clearpage
%
%
\bibliographystyle{splncs04}
\bibliography{ref}

\clearpage
\section*{A. Appendix}
\section*{A.1.  Method - Details of region-scale/contextual-relation pseudo labels and regularizer weight}
We would share more details about the region-scale/contextual-relation pseudo labels and the weight of regularizer used in this paper. For the \textbf{source domain}, the sizes of the input image for datasets GTA5 and SYNTHIA are $720 \times 1280$ and $760 \times 1280$ , respectively. In this paper, we use two types of regions with two different sizes. The first sizes of regions for datasets GTA5 and SYNTHIA are $18 \times 32$ and $19 \times 32$, respectively. The second sizes of regions for datasets GTA5 and SYNTHIA are $36 \times 64$ and $38 \times 64$, respectively. For the \textbf{target domain} (dataset Cityscapes), the size of input image is $512 \times 1024$. The sizes of regions are $16 \times 32$ and $32 \times 64$, respectively. We use two independent contextual-relations (CR) classifiers to deal with these two types of regions with two different sizes. The weight of the regularizer in adaptive entropy max-minimizing adversarial learning scheme decreases with training iteration, which is expressed as: $\lambda_{R} = (1 - \frac{iter}{max\_iter})^{power}$ with ${power} = 0.9$.

\section*{A.2. Method - Traditional Losses}
For the source domain, traditional approaches learn a supervised segmentation model $G$ that aims to minimize a segmentation loss. For the target domain, UDA networks using adversarial learning  train $G$ to extract domain-invariant features though the minimaxing game between $G$ and a domain discriminator $D$. The overall loss in the UDA networks can therefore be formulated by:
\begin{equation}
\mathcal{L}(X_{s}, X_{t}) = \mathcal{L}_{seg}(G) + \mathcal{L}_{adv}(G, D)
\end{equation}

\section*{A.3. Method - Loss in Multi-Scale Adaptation}
\textbf{Source Flow:} 
In our contextual-relation consistent domain adaptation (CrCDA) with multi-scale form, the source-domain data contribute to $\mathcal{L}_{seg}$, $\mathcal{L}_{cr}$ and $\mathcal{L}_{D}$. Given a source-domain image $x_{s} \subset X_{s}$ and the corresponding pixel-scale label $y_{s} \subset Y_{s}$, region-scale (contextual-relations) pseudo label $y_{s\_cr} \subset Y_{s\_cr}$, $P_{s}^{(h, w, c)} = C_{seg}(E(x_{s}))$ is the predicted probability map $w.r.t$ each pixel over C classes; $P_{s\_cr}^{(i, j, n)} = C_{cr}(E(x_{s}))$ is the predicted probability map $w.r.t$ each region over $N$ classes. The layout probability map $P_{s\_layout}^{(h, w, c+n)}$ is generated by concatenating $P_{s}^{(h, w, c)}$ and up-sampled $P_{s\_cr}^{(i, j, n)}$. $\mathcal{L}_{seg}$ and $\mathcal{L}_{cr}$ are provided in the submitted manuscript. $\mathcal{L}_{s_d}$ is formulated as follows:

\begin{equation}
\begin{split}
& \mathcal{L}_{s_d}(E, C_{seg}, C_{cr}, C_{D}) = \sum_{h, w} E[\log C_{D}(P_{s\_layout}^{(h, w, c+n)})]\\
\end{split}
\end{equation}

\textbf{Target Flow:} 
As the target label is not accessible, we design an adversarial training scheme between feature extractor $E$ and classifiers ($C_{seg}$, $C_{cr}$ and $C_{D}$) that extracts discriminative features via max-minimizing entropy in the target domain. Given a target image $x_{t} \subset X_{t}$, $P_{t}^{(h, w, c)} = C_{seg}(E(x_{t}))$ is the predicted probability map $w.r.t$ each target pixel over C classes; $P_{t\_cr}^{(i, j, n)} = C_{cr}(E(x_{t}))$ is the predicted probability map $w.r.t$ each target region over $N$ classes. The layout probability map $P_{t\_layout}^{(h, w, c+n)}$ of the target-domain image is generated by concatenating $P_{t}^{(h, w, c)}$ and up-sampled $P_{t\_cr}^{(i, j, n)}$. $\mathcal{L}\_{ent\_pix}$ and $\mathcal{L}_{ent\_cr}$ are provided in the submitted manuscript. $\mathcal{L}_{t_d}$ is expressed as:
\begin{equation}
\begin{split}
& \mathcal{L}_{t_d}(E, C_{seg}, C_{cr}, C_{D}) = \sum_{h, w} E[\log (1 - C_{D}(P_{t\_layout}^{(h, w, c+n)}))]\\
\end{split}
\end{equation}

Therefore, the overall global alignment loss is expressed as:
\begin{equation}
\begin{split}
& \mathcal{L}_{D}(E, C_{seg}, C_{cr}, C_{D}) = \mathcal{L}_{s_d} + \mathcal{L}_{t_d} + Ent_{s_d} + Ent_{t_d}\\
\end{split}
\end{equation}
where domain classifier entropy is $Ent_{s_d} = - C_{D}(P_{s\_layout}^{(h, w, c+n)}) \log C_{D}(P_{s\_layout}^{(h, w, c+n)})$ for source domain; similarly, $Ent_{t_d} = - C_{D}(P_{t\_layout}^{(h, w, c+n)}) \log C_{D}(P_{t\_layout}^{(h, w, c+n)})$ for target domain.

\section*{A.4. Experiment - More Qualitative Results}
\begin{figure*}[t]
\begin{tabular}{p{2.35cm}<{\centering} p{2.35cm}<{\centering} p{2.35cm}<{\centering} p{2.35cm}<{\centering} p{2.35cm}<{\centering}}
\raisebox{-0.5\height}{\includegraphics[width=1.02\linewidth,height=0.51\linewidth]{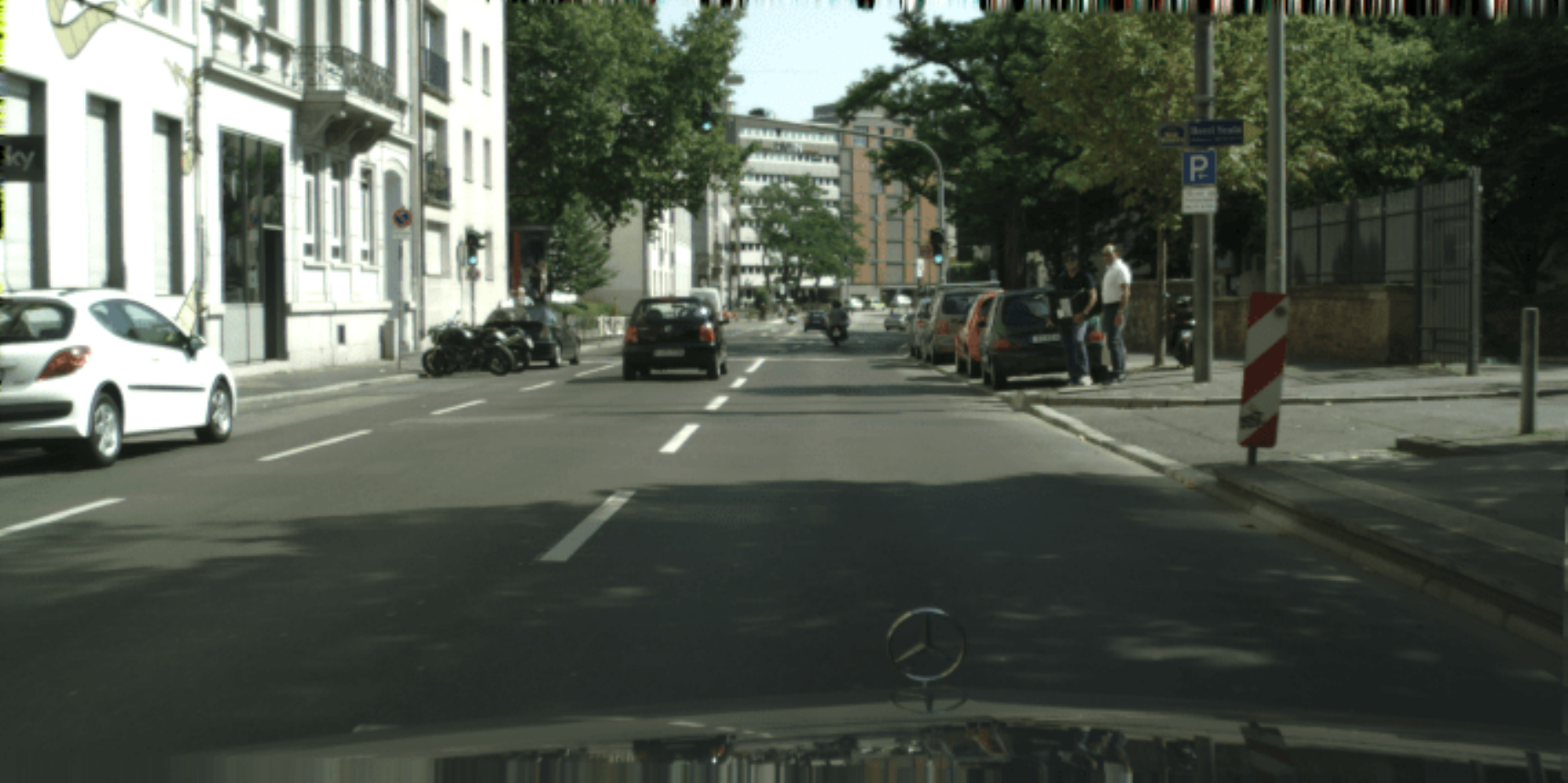}} 
 & \raisebox{-0.5\height}{\includegraphics[width=1.02\linewidth,height=0.51\linewidth]{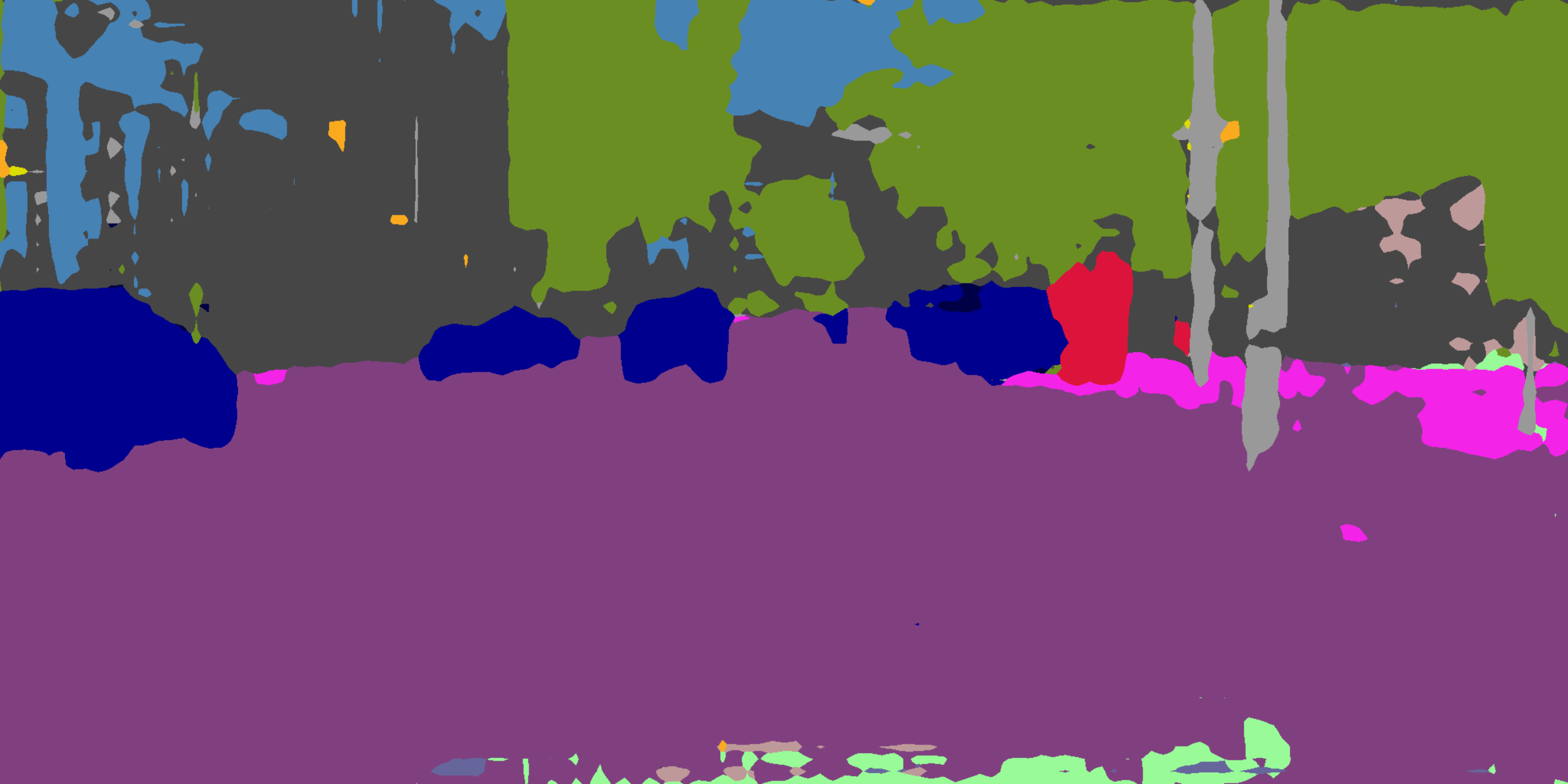}}
  & \raisebox{-0.5\height}{\includegraphics[width=1.02\linewidth,height=0.51\linewidth]{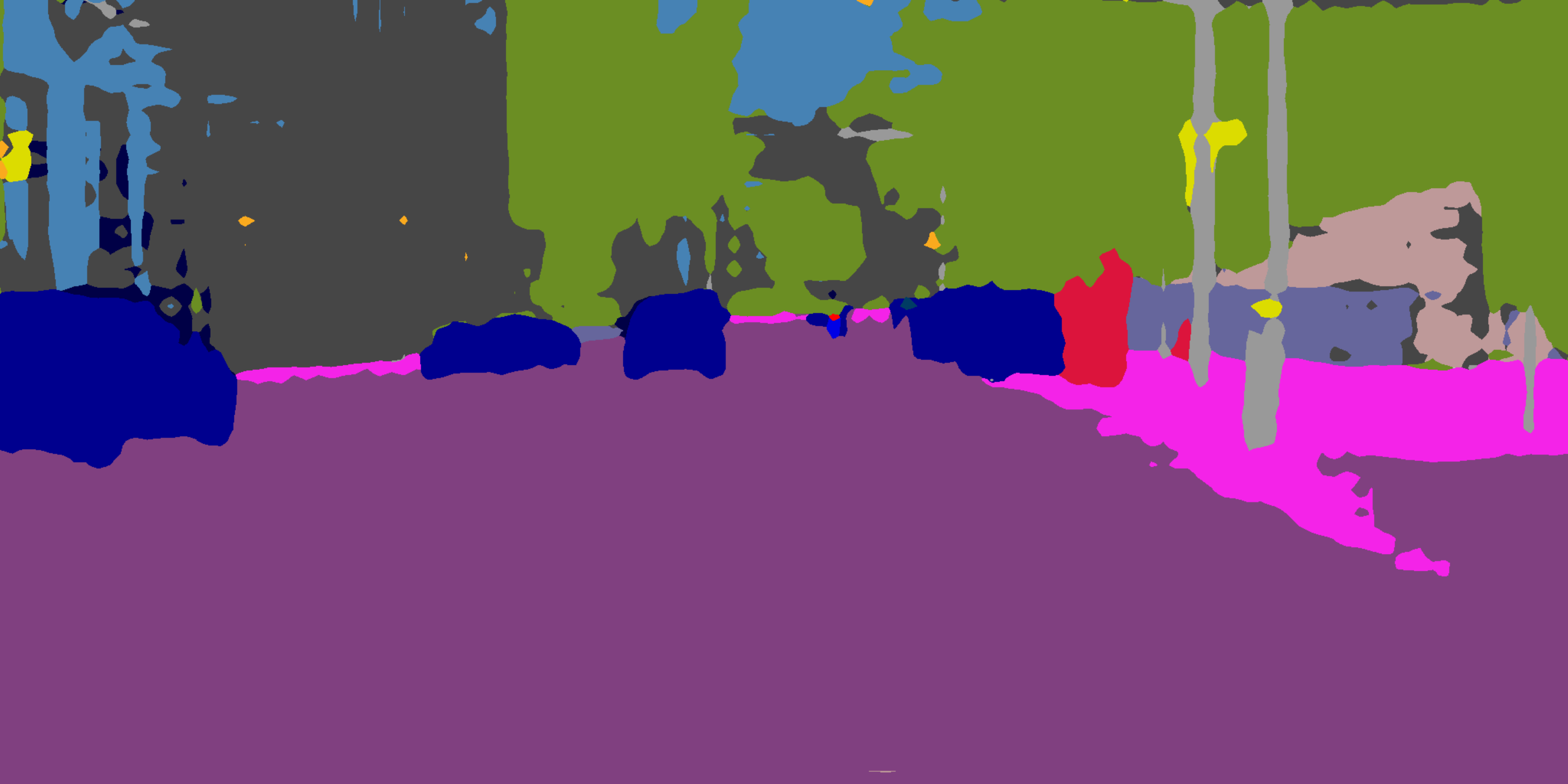}}
& \raisebox{-0.5\height}{\includegraphics[width=1.02\linewidth,height=0.51\linewidth]{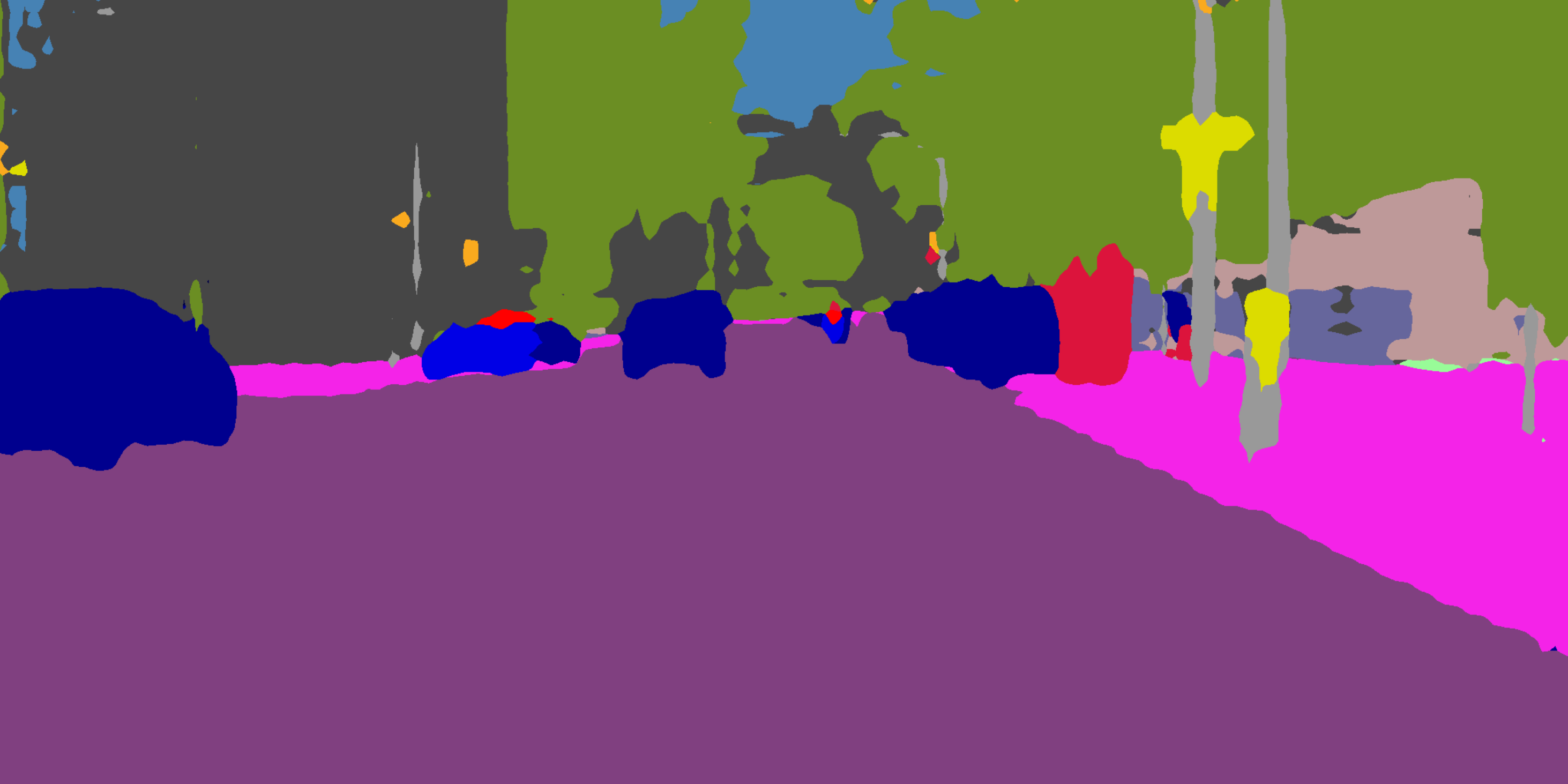}}
& \raisebox{-0.5\height}{\includegraphics[width=1.02\linewidth,height=0.51\linewidth]{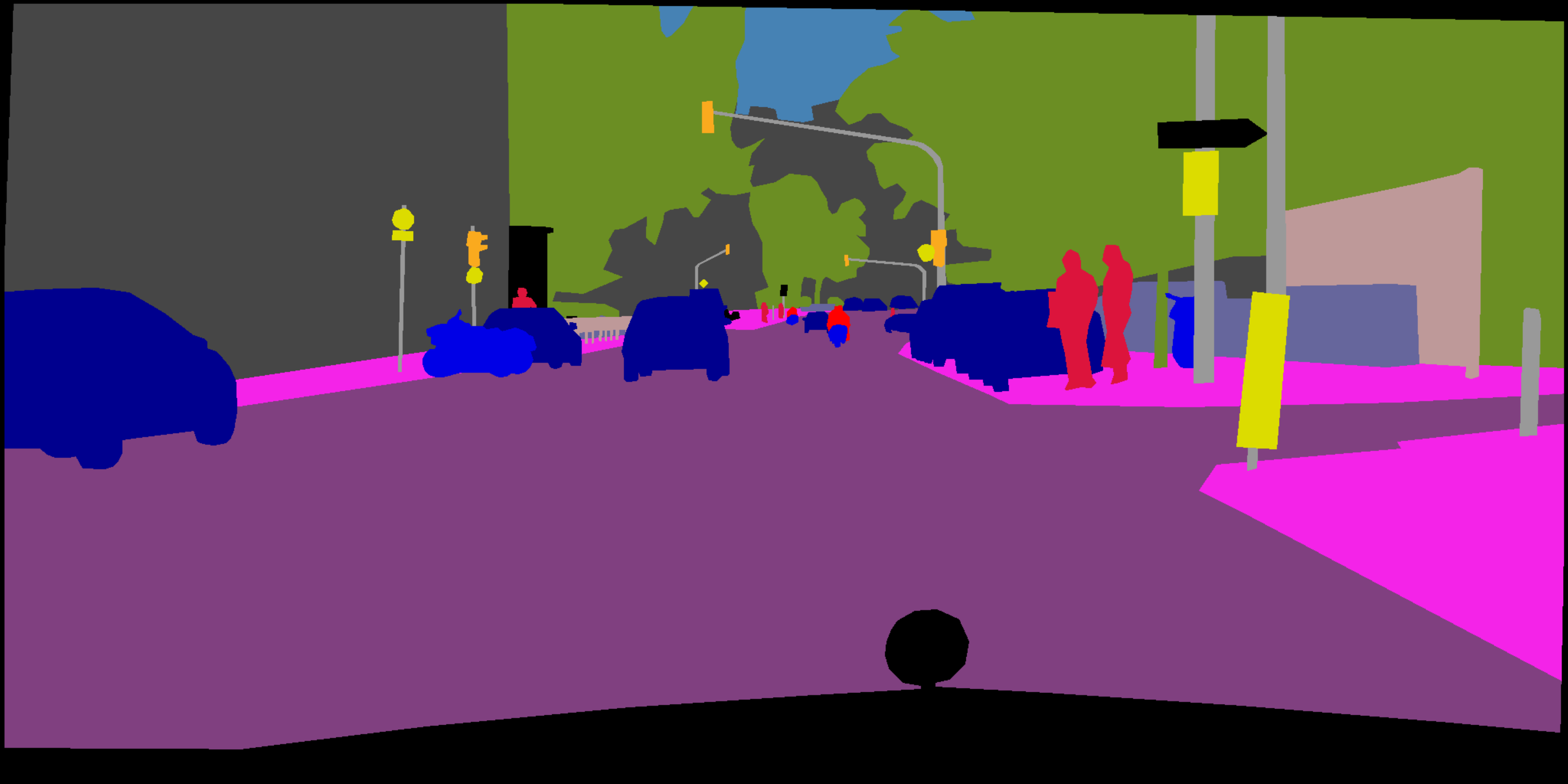}}
\\

\raisebox{-0.5\height}{\includegraphics[width=1.02\linewidth,height=0.51\linewidth]{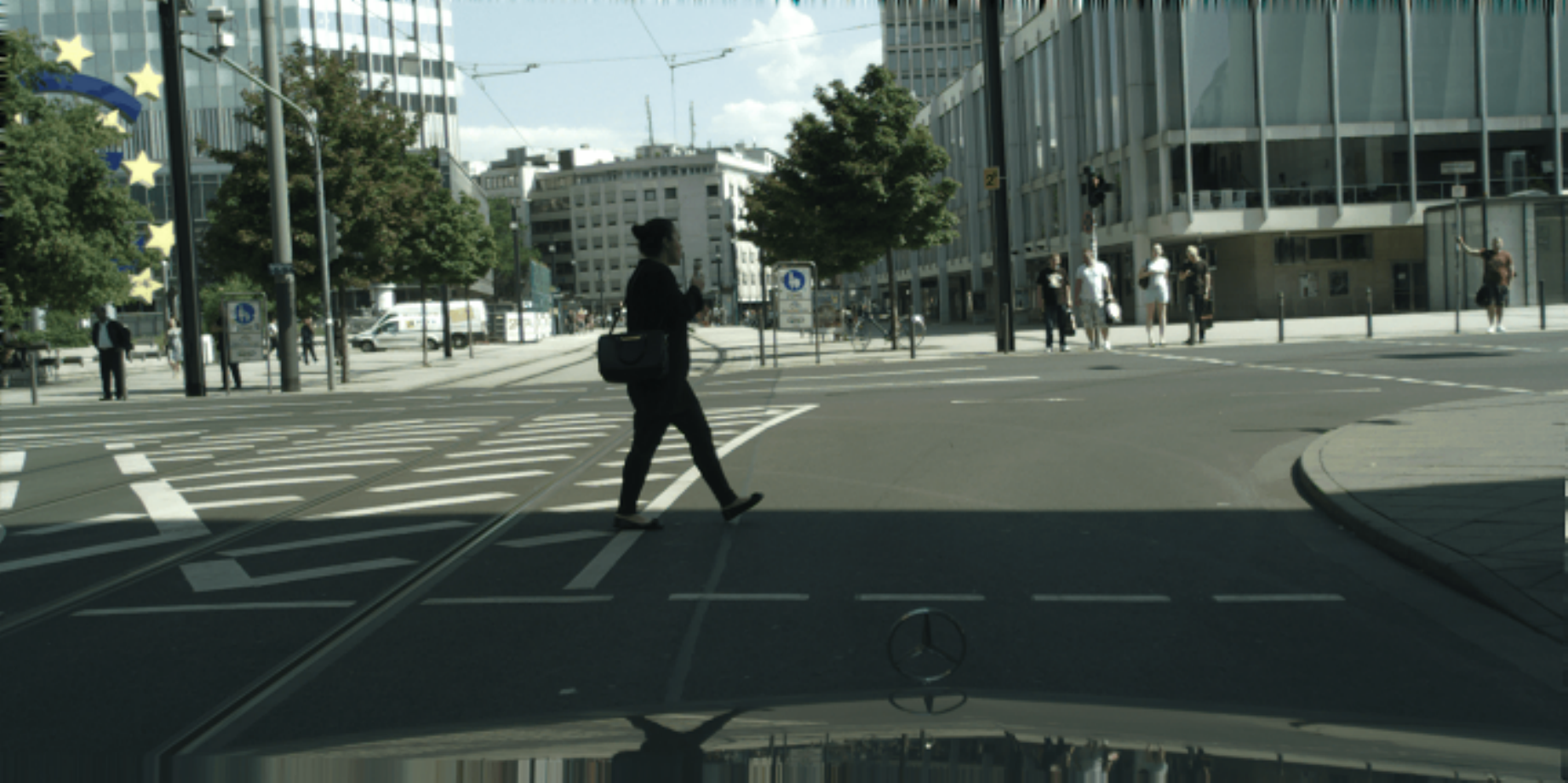}} 
 & \raisebox{-0.5\height}{\includegraphics[width=1.02\linewidth,height=0.51\linewidth]{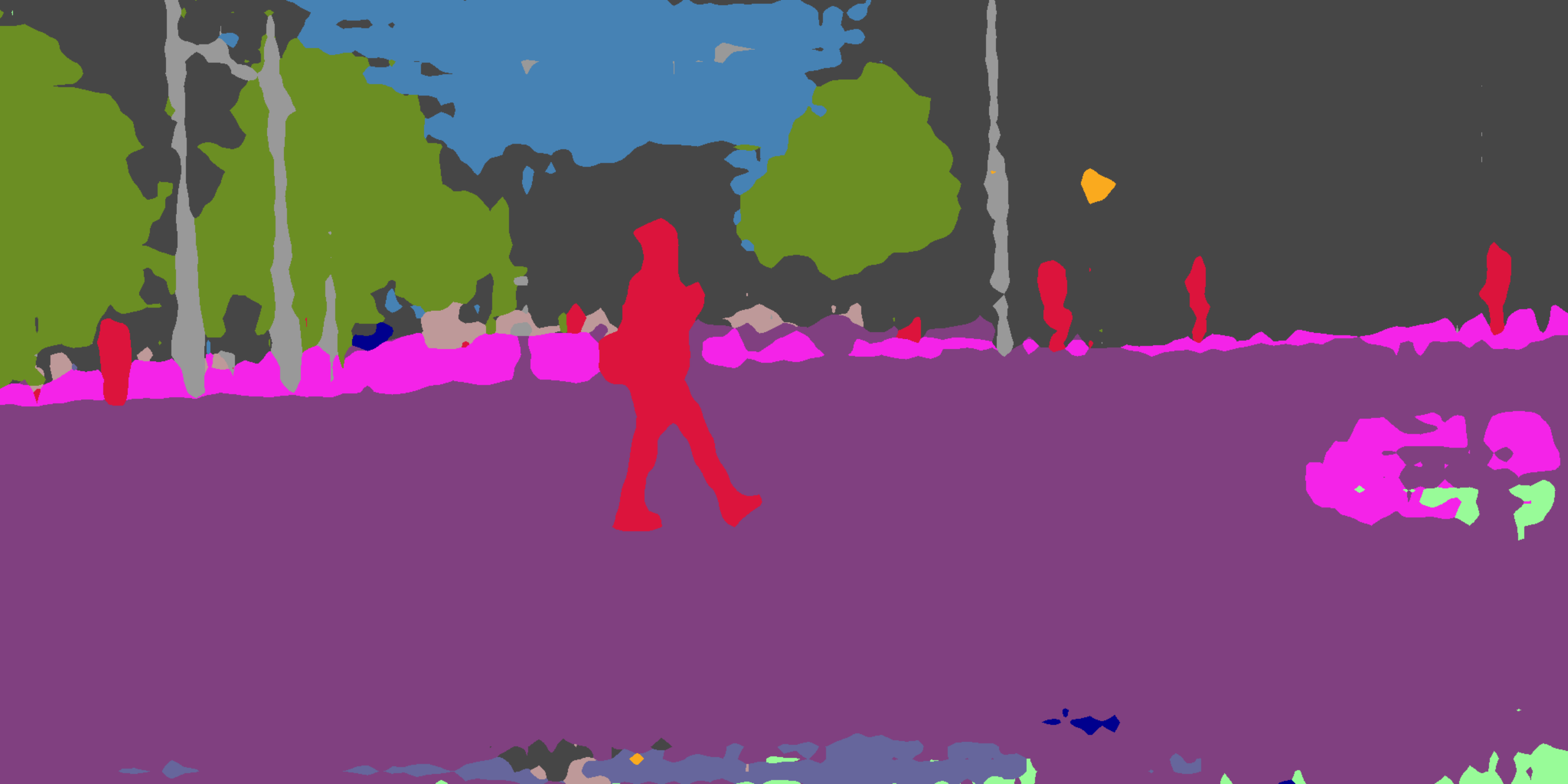}}
  & \raisebox{-0.5\height}{\includegraphics[width=1.02\linewidth,height=0.51\linewidth]{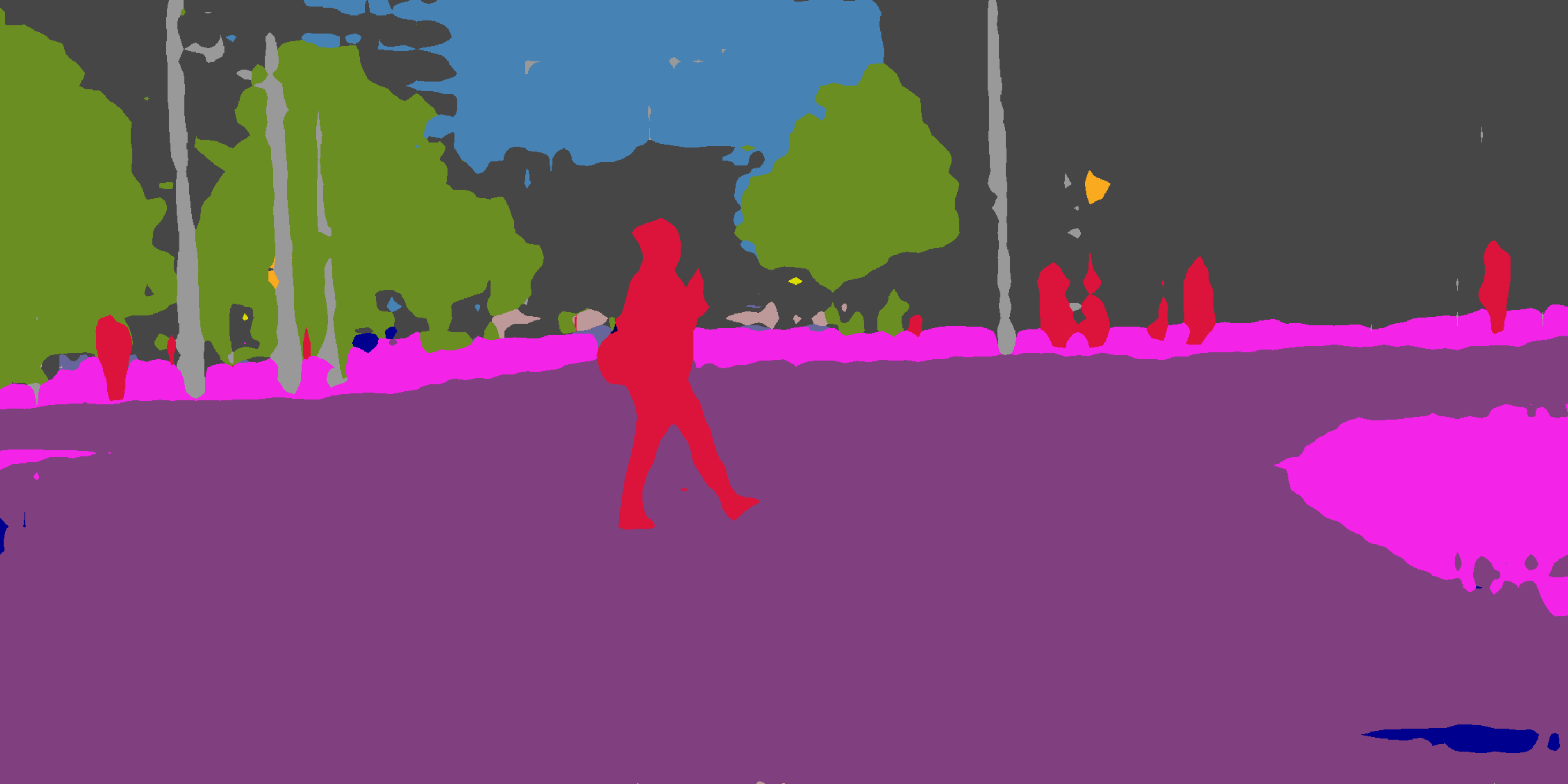}}
& \raisebox{-0.5\height}{\includegraphics[width=1.02\linewidth,height=0.51\linewidth]{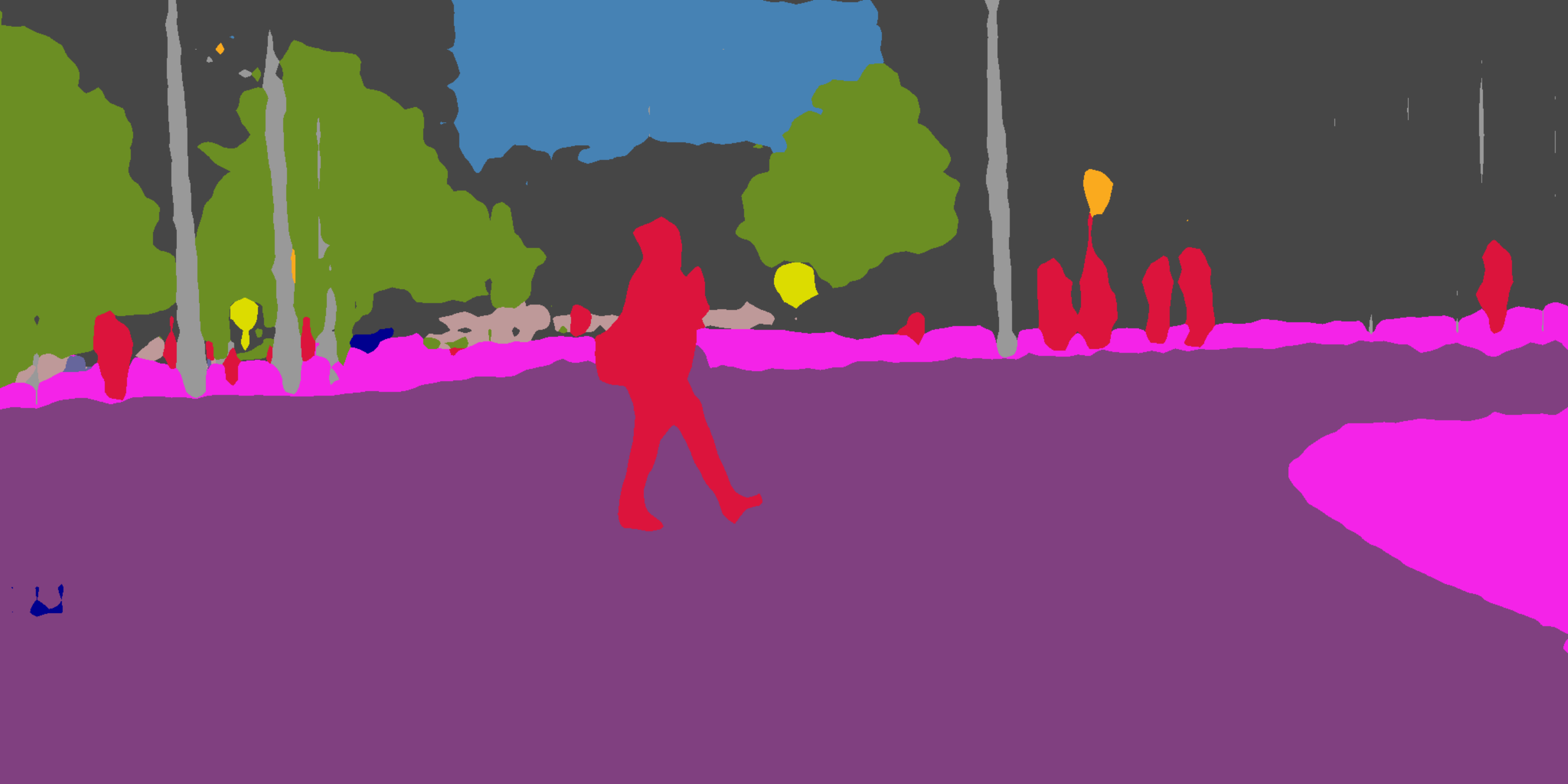}}
& \raisebox{-0.5\height}{\includegraphics[width=1.02\linewidth,height=0.51\linewidth]{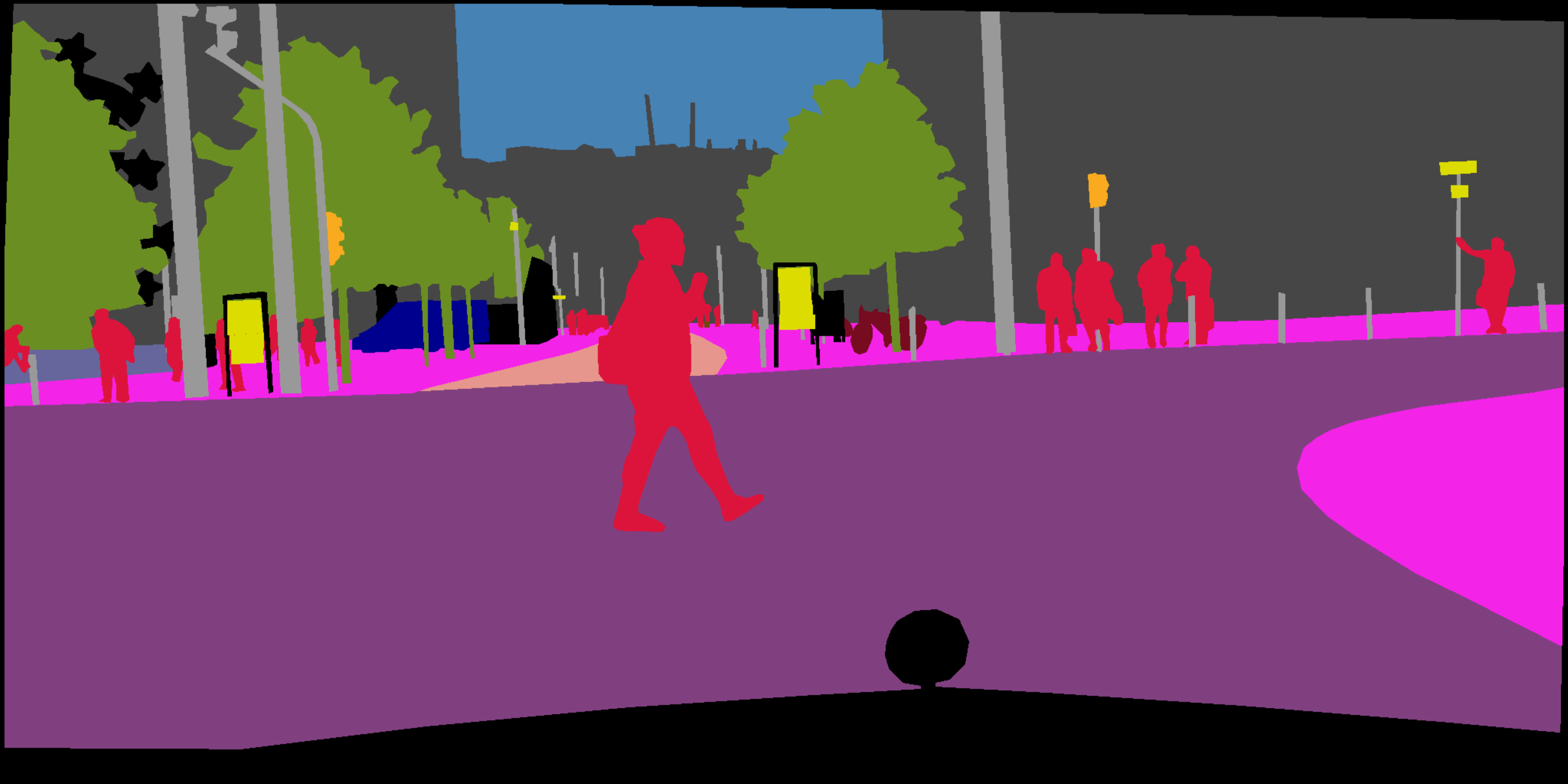}}
\\

\raisebox{-0.5\height}{\includegraphics[width=1.02\linewidth,height=0.51\linewidth]{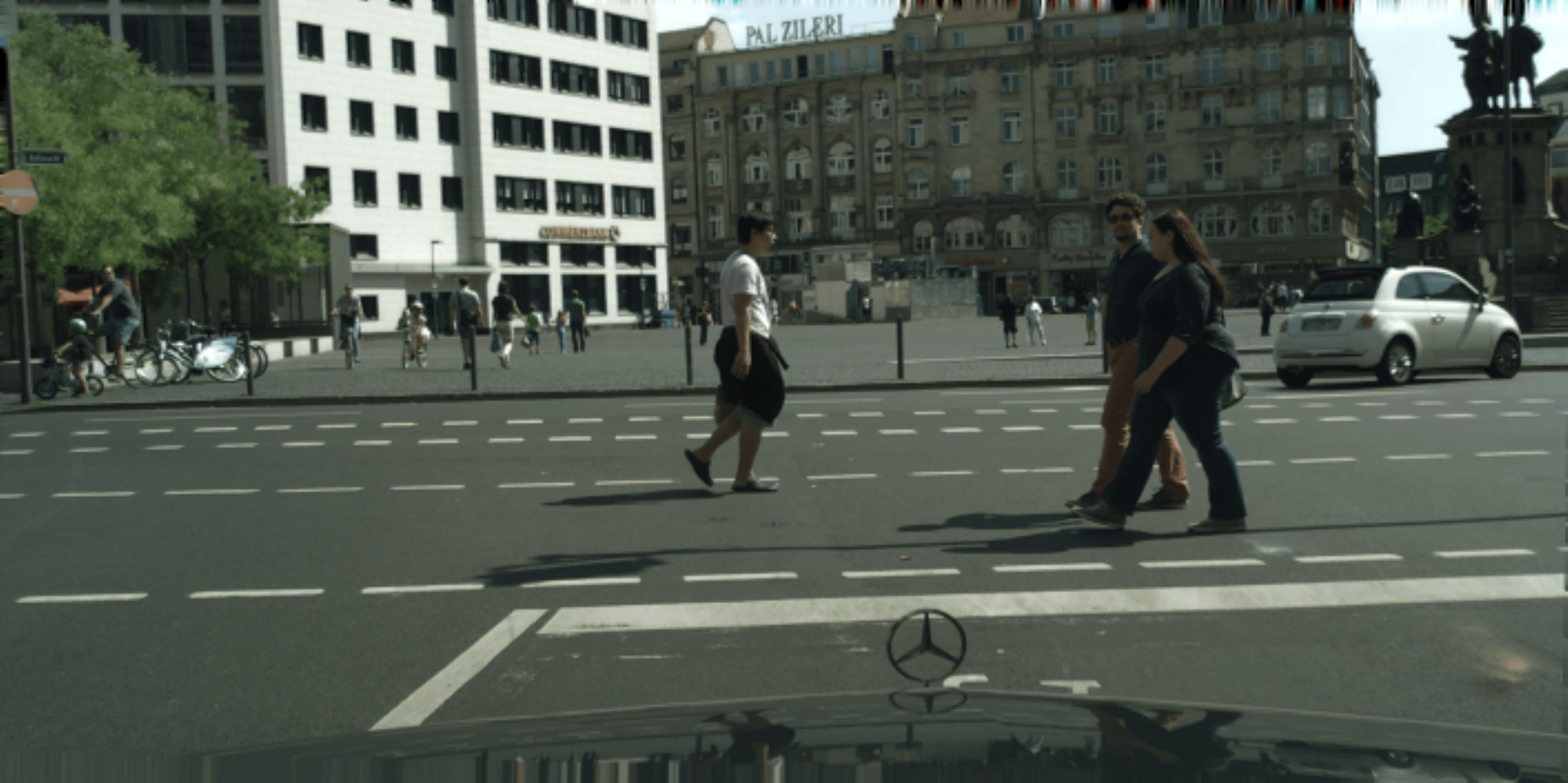}} 
 & \raisebox{-0.5\height}{\includegraphics[width=1.02\linewidth,height=0.51\linewidth]{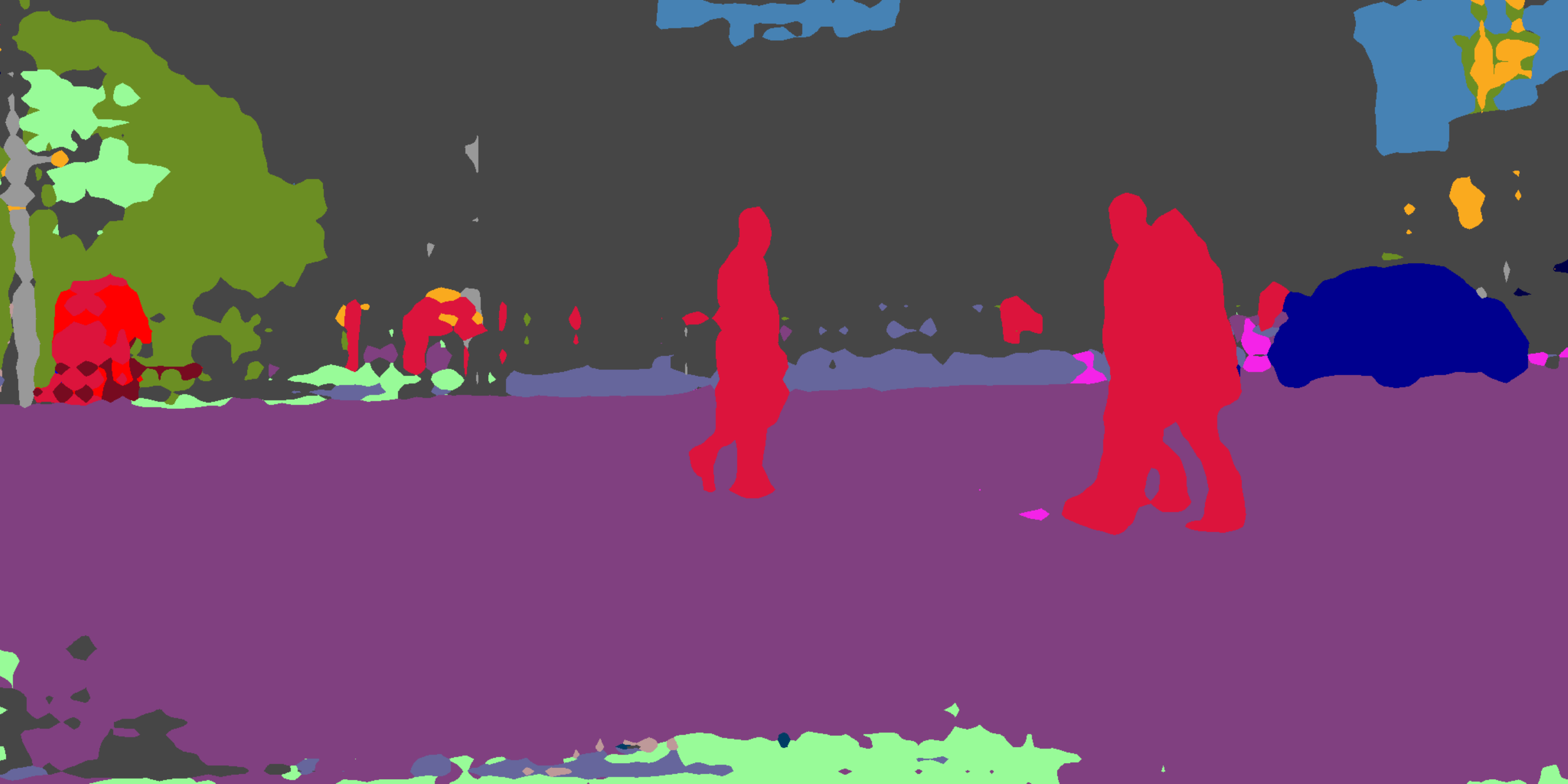}}
  & \raisebox{-0.5\height}{\includegraphics[width=1.02\linewidth,height=0.51\linewidth]{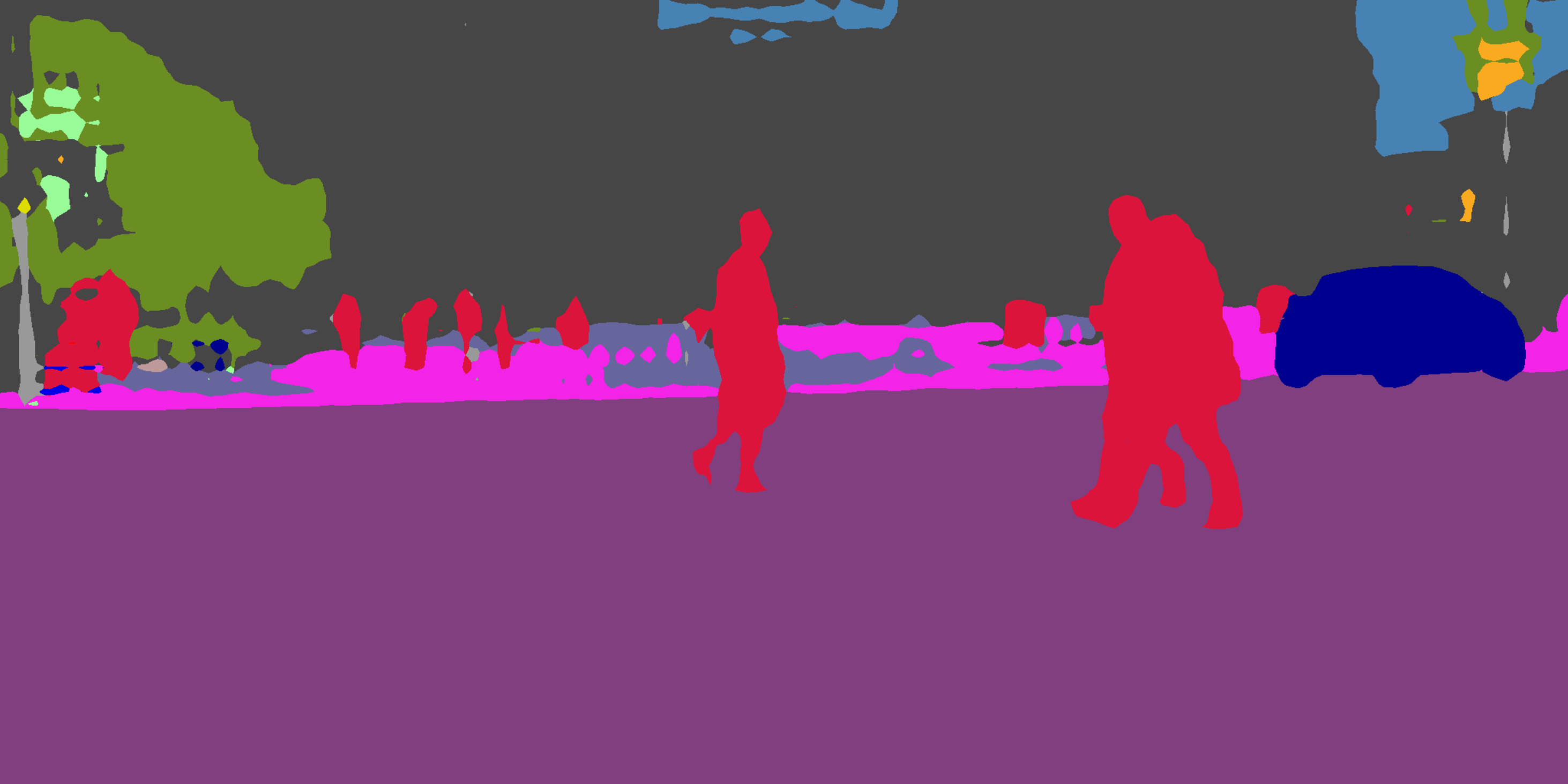}}
& \raisebox{-0.5\height}{\includegraphics[width=1.02\linewidth,height=0.51\linewidth]{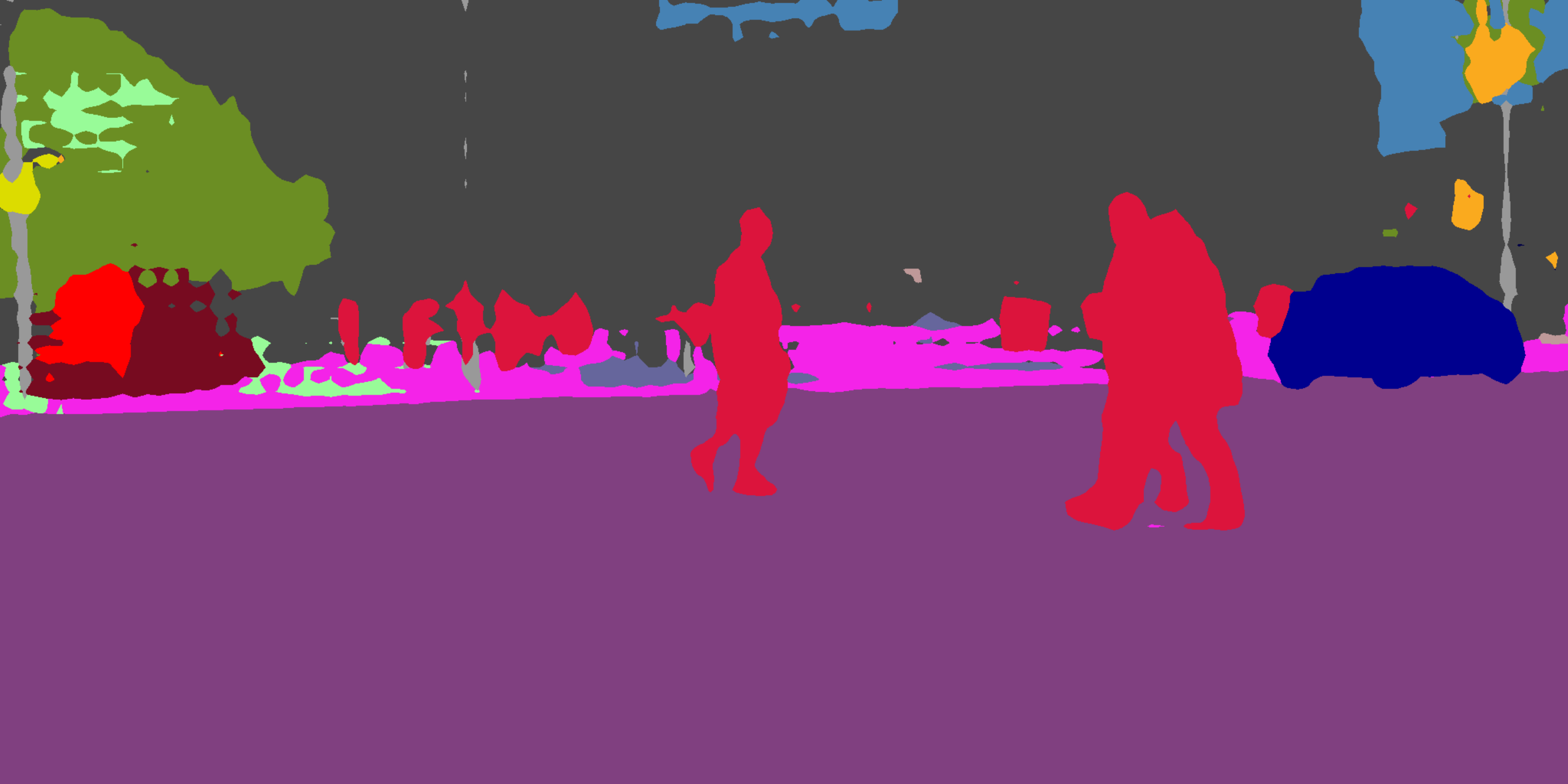}}
& \raisebox{-0.5\height}{\includegraphics[width=1.02\linewidth,height=0.51\linewidth]{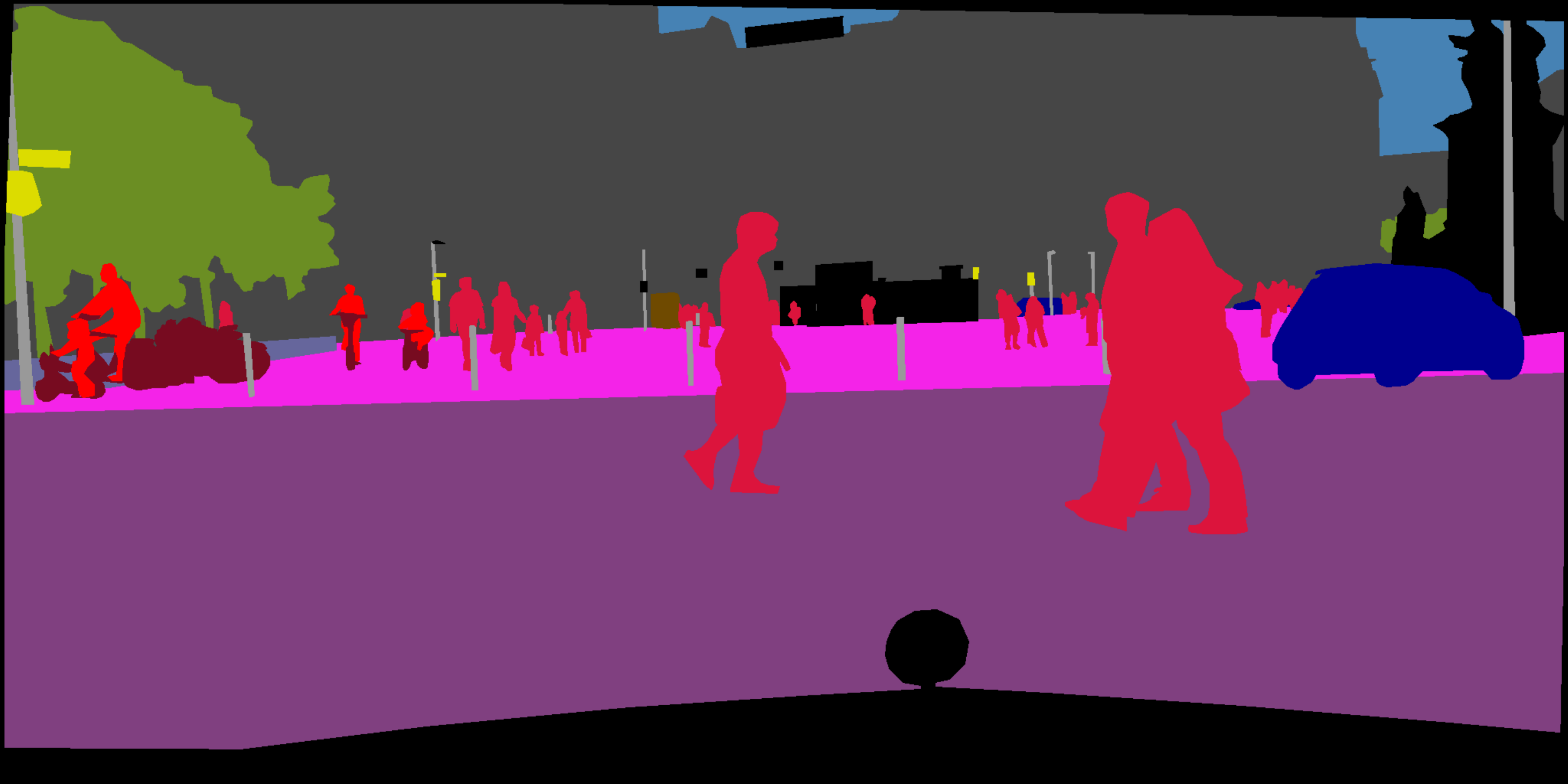}}
\\

\raisebox{-0.5\height}{\includegraphics[width=1.02\linewidth,height=0.51\linewidth]{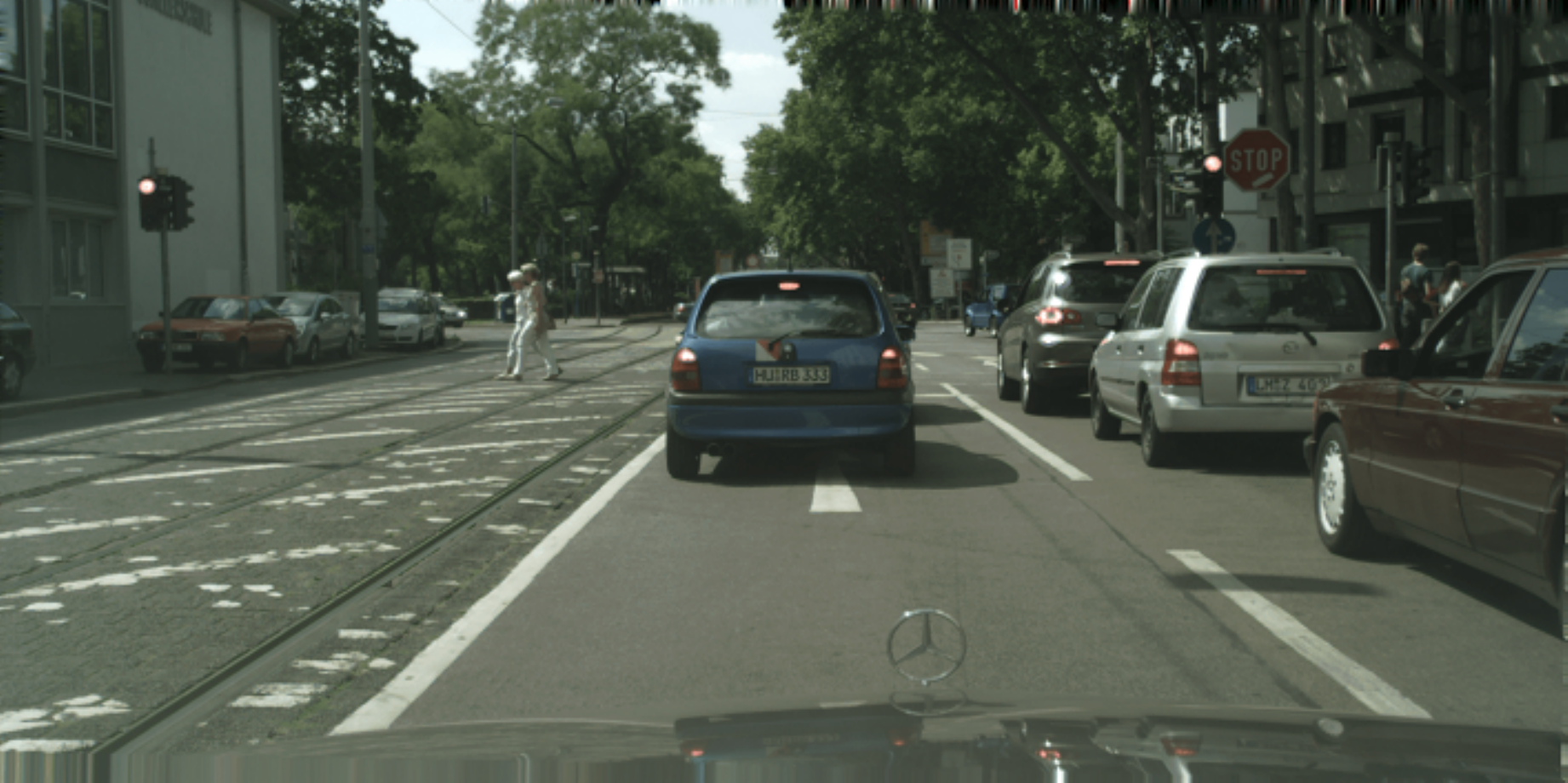}} 
 & \raisebox{-0.5\height}{\includegraphics[width=1.02\linewidth,height=0.51\linewidth]{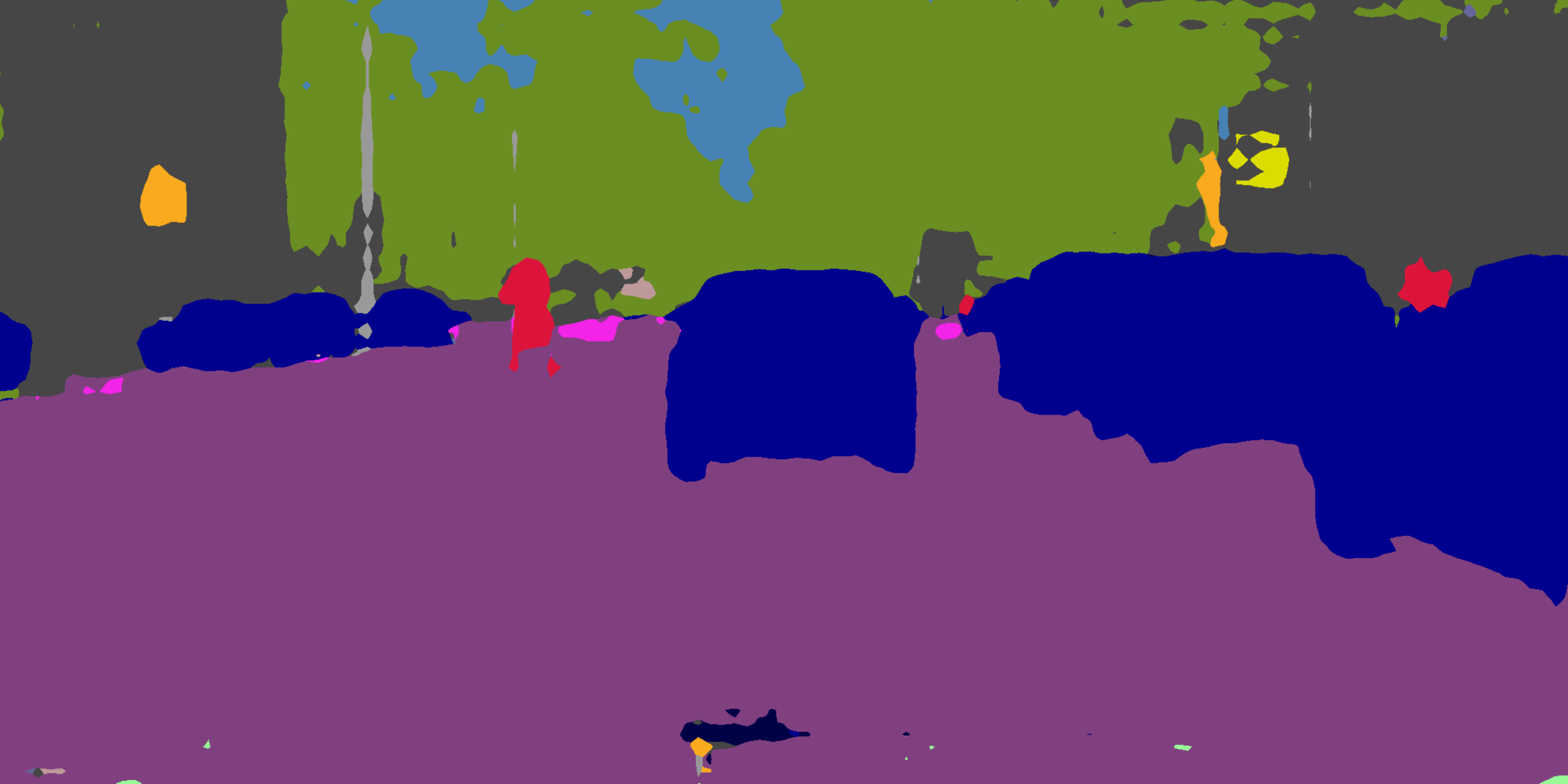}}
  & \raisebox{-0.5\height}{\includegraphics[width=1.02\linewidth,height=0.51\linewidth]{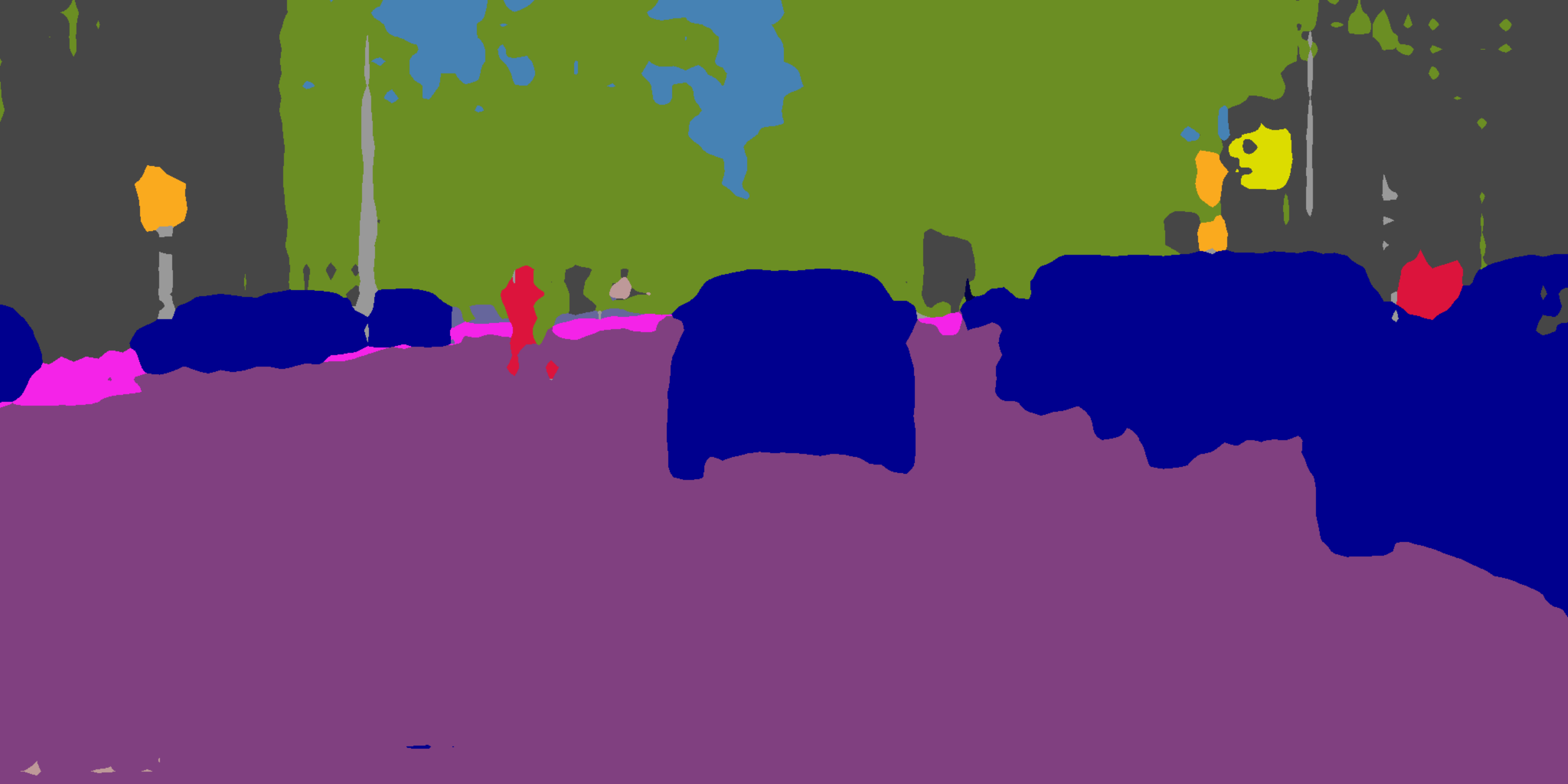}}
& \raisebox{-0.5\height}{\includegraphics[width=1.02\linewidth,height=0.51\linewidth]{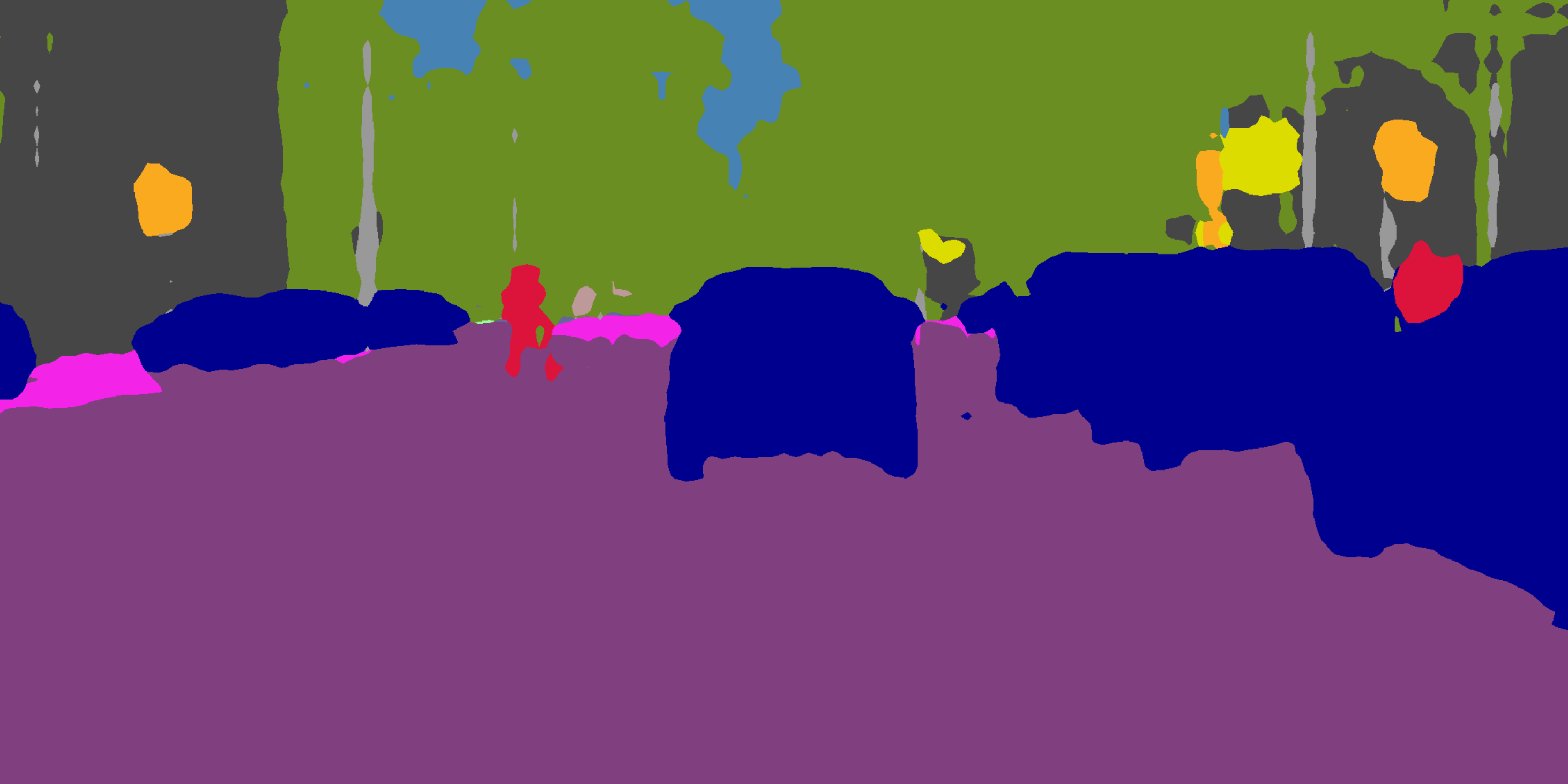}}
& \raisebox{-0.5\height}{\includegraphics[width=1.02\linewidth,height=0.51\linewidth]{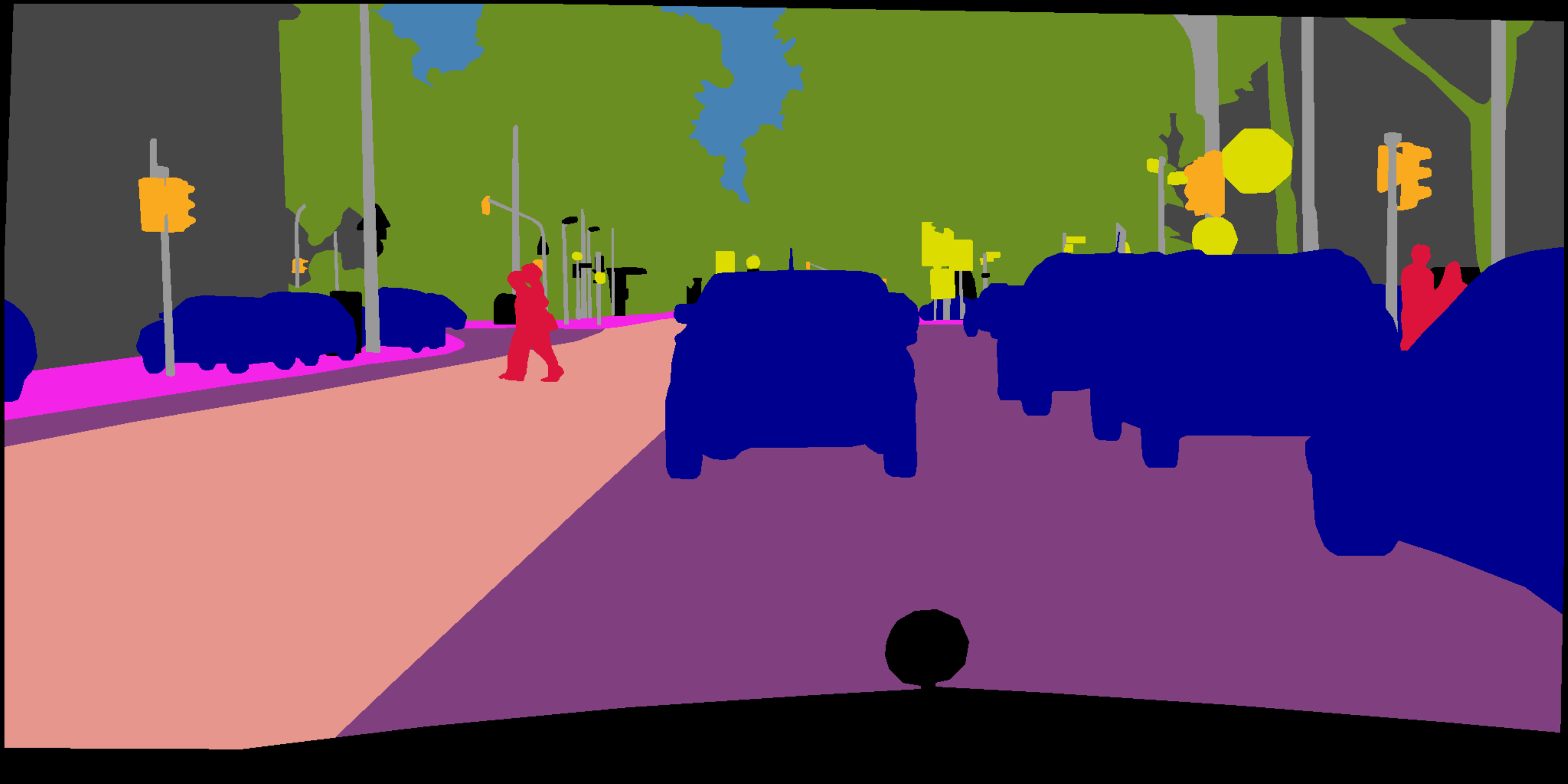}}
\\

\raisebox{-0.5\height}{\includegraphics[width=1.02\linewidth,height=0.51\linewidth]{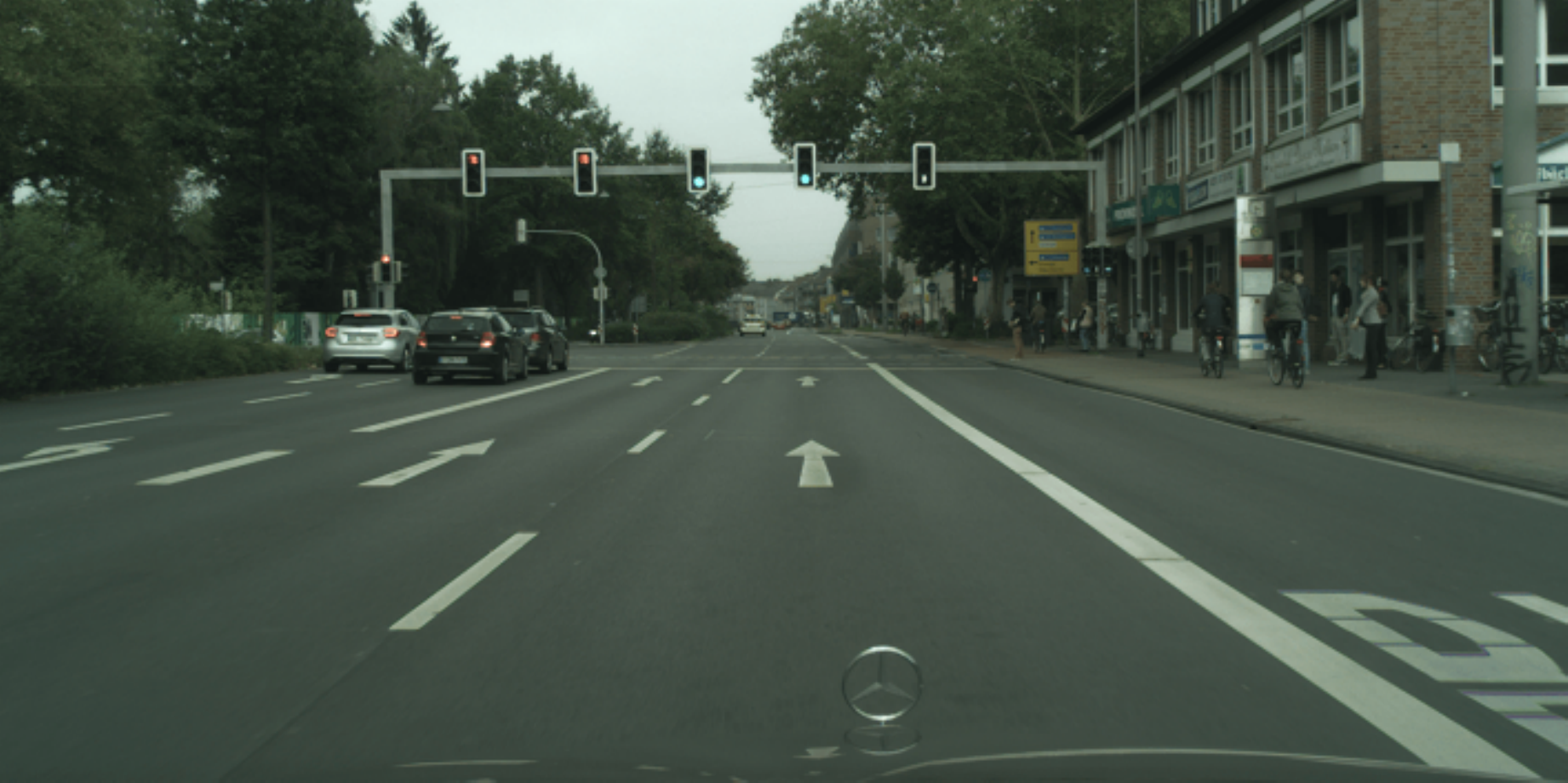}} 
 & \raisebox{-0.5\height}{\includegraphics[width=1.02\linewidth,height=0.51\linewidth]{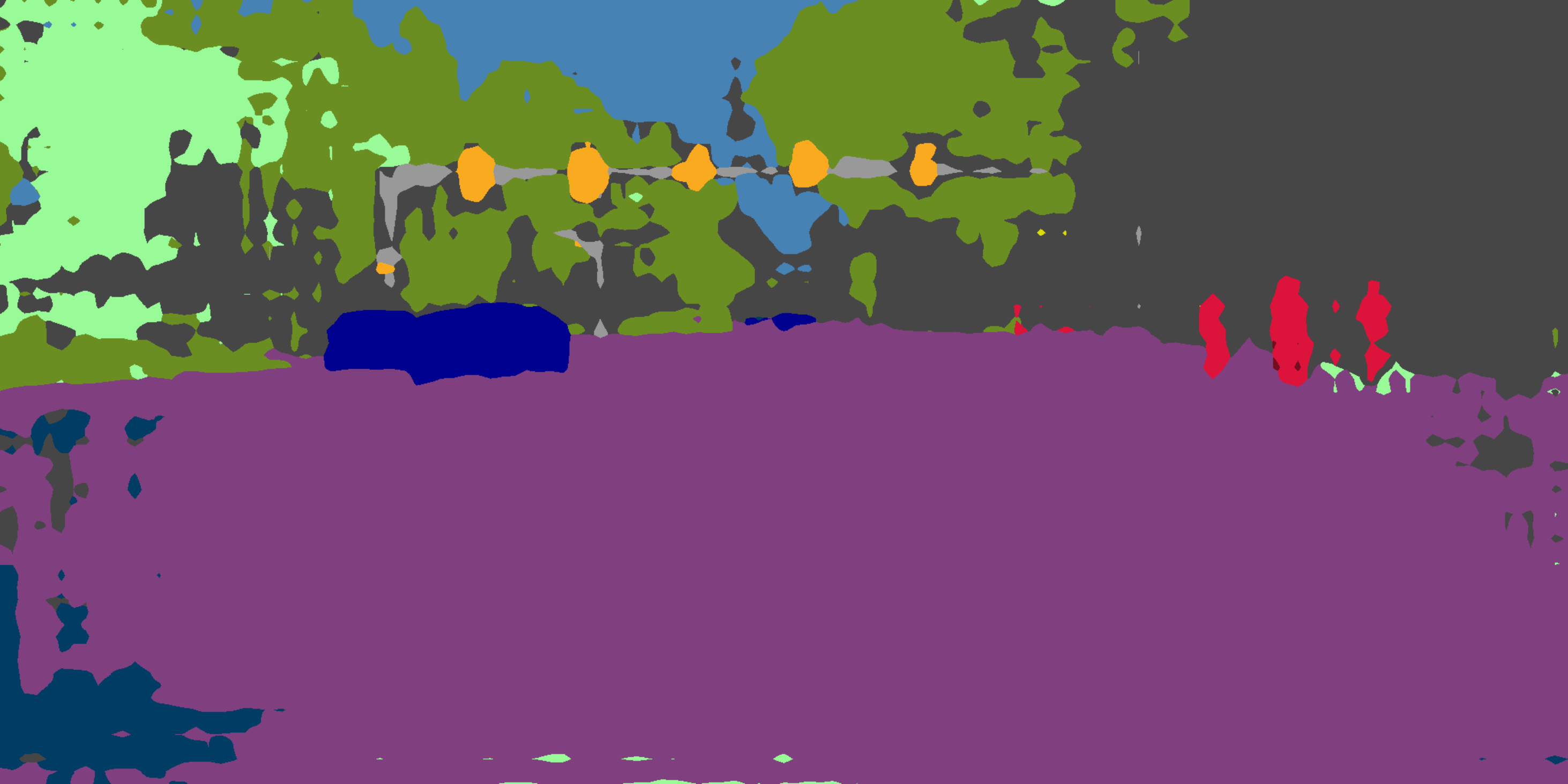}}
  & \raisebox{-0.5\height}{\includegraphics[width=1.02\linewidth,height=0.51\linewidth]{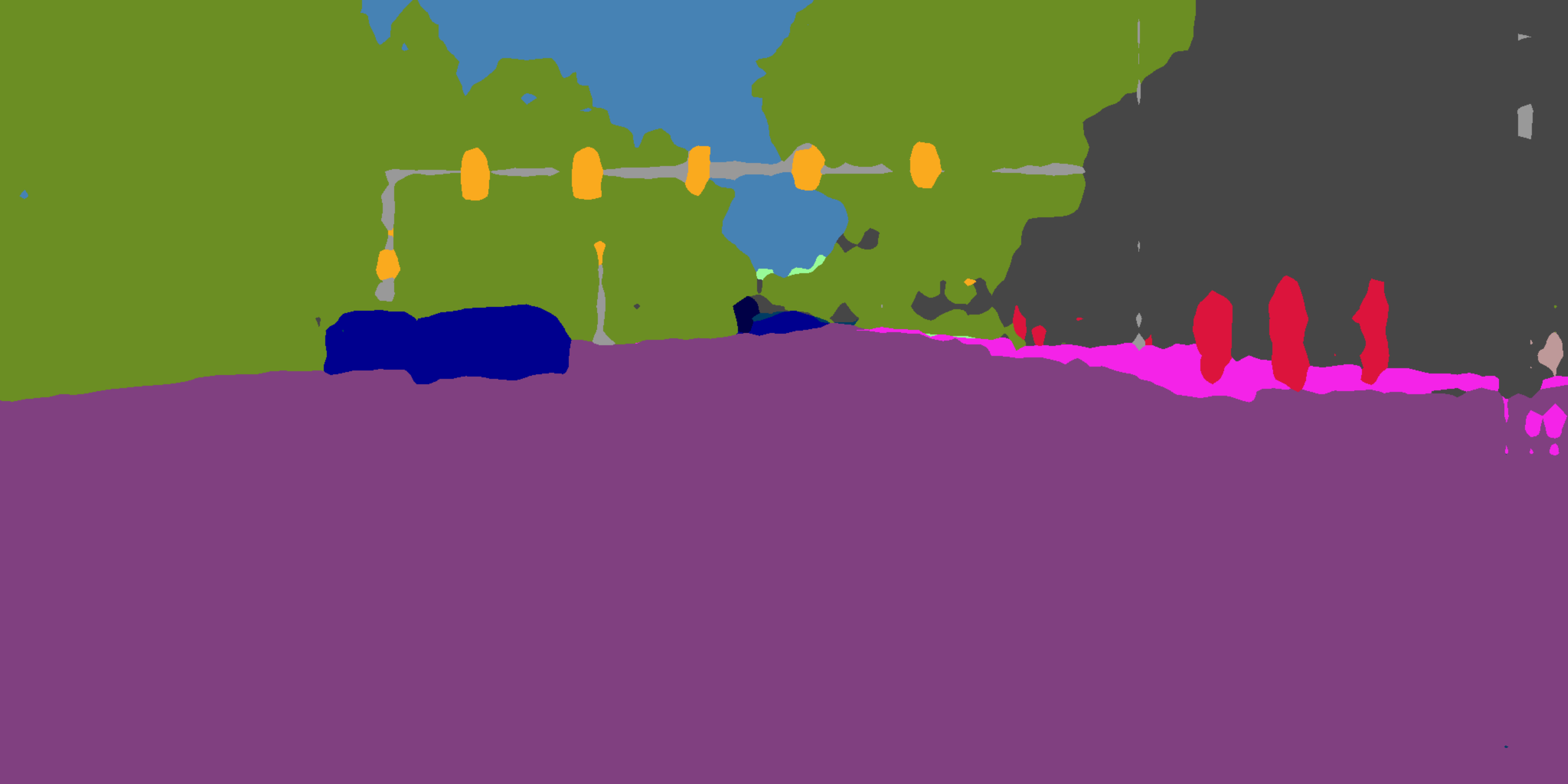}}
& \raisebox{-0.5\height}{\includegraphics[width=1.02\linewidth,height=0.51\linewidth]{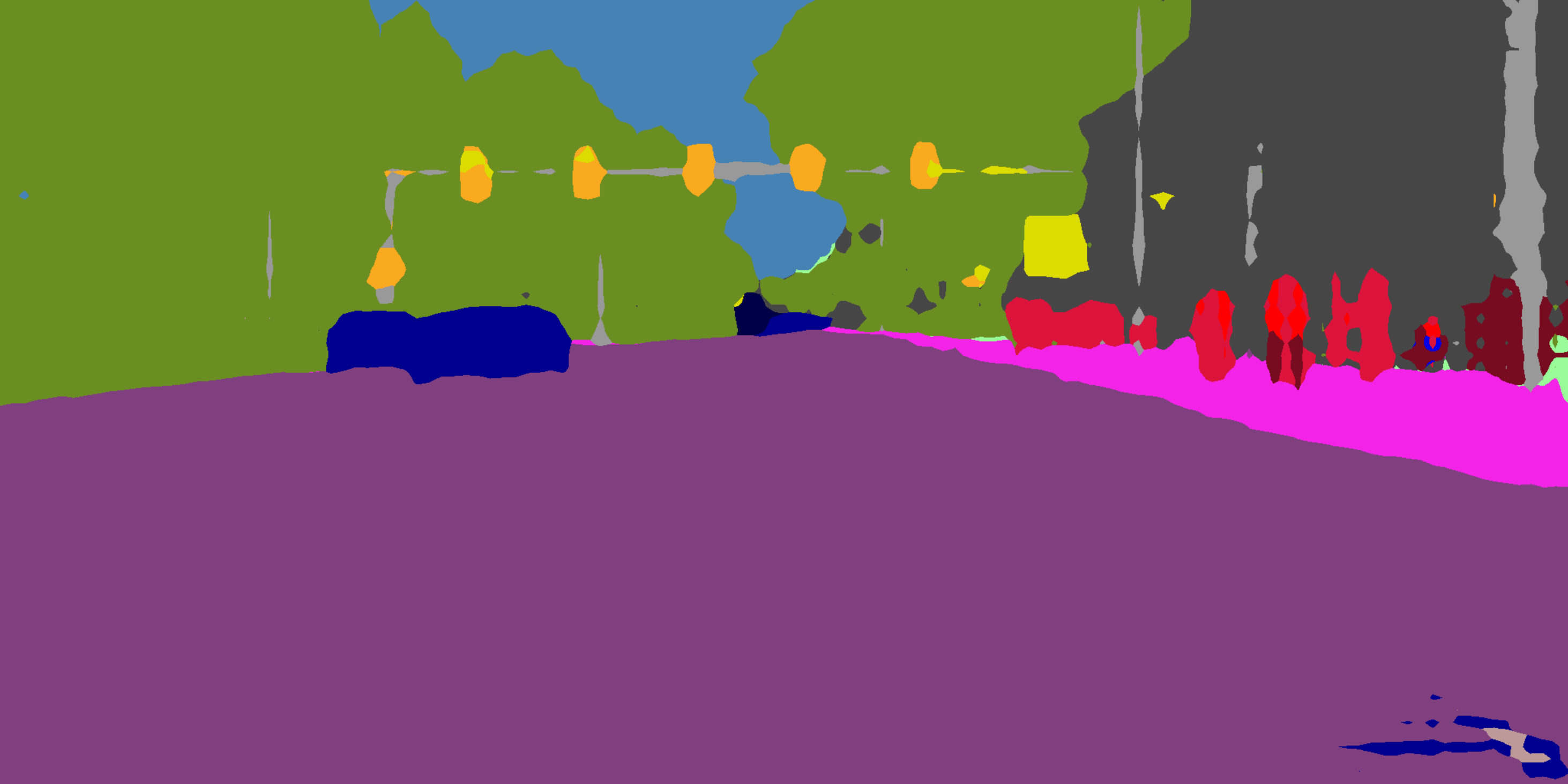}}
& \raisebox{-0.5\height}{\includegraphics[width=1.02\linewidth,height=0.51\linewidth]{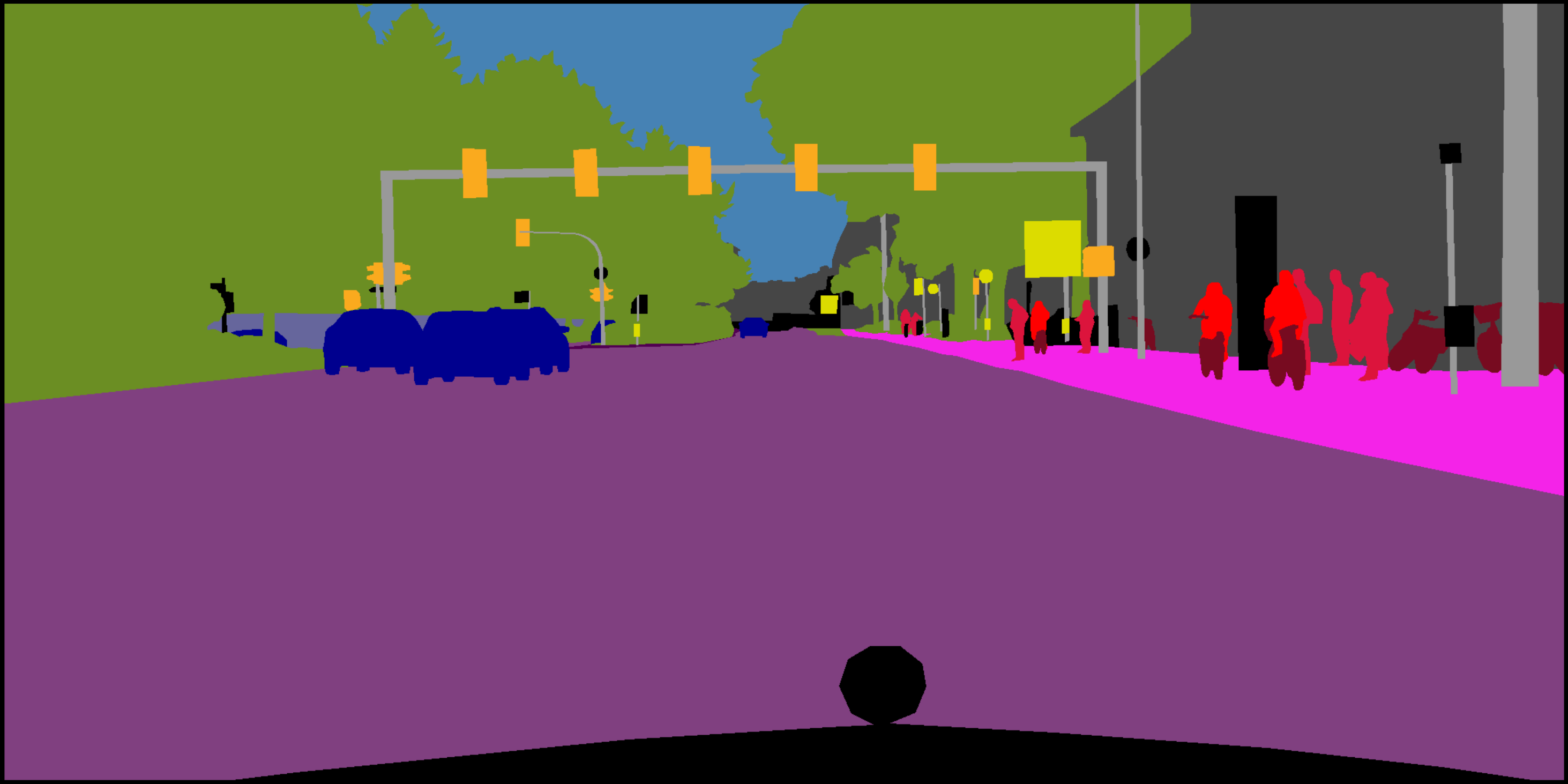}}
\\

\raisebox{-0.5\height}{\includegraphics[width=1.02\linewidth,height=0.51\linewidth]{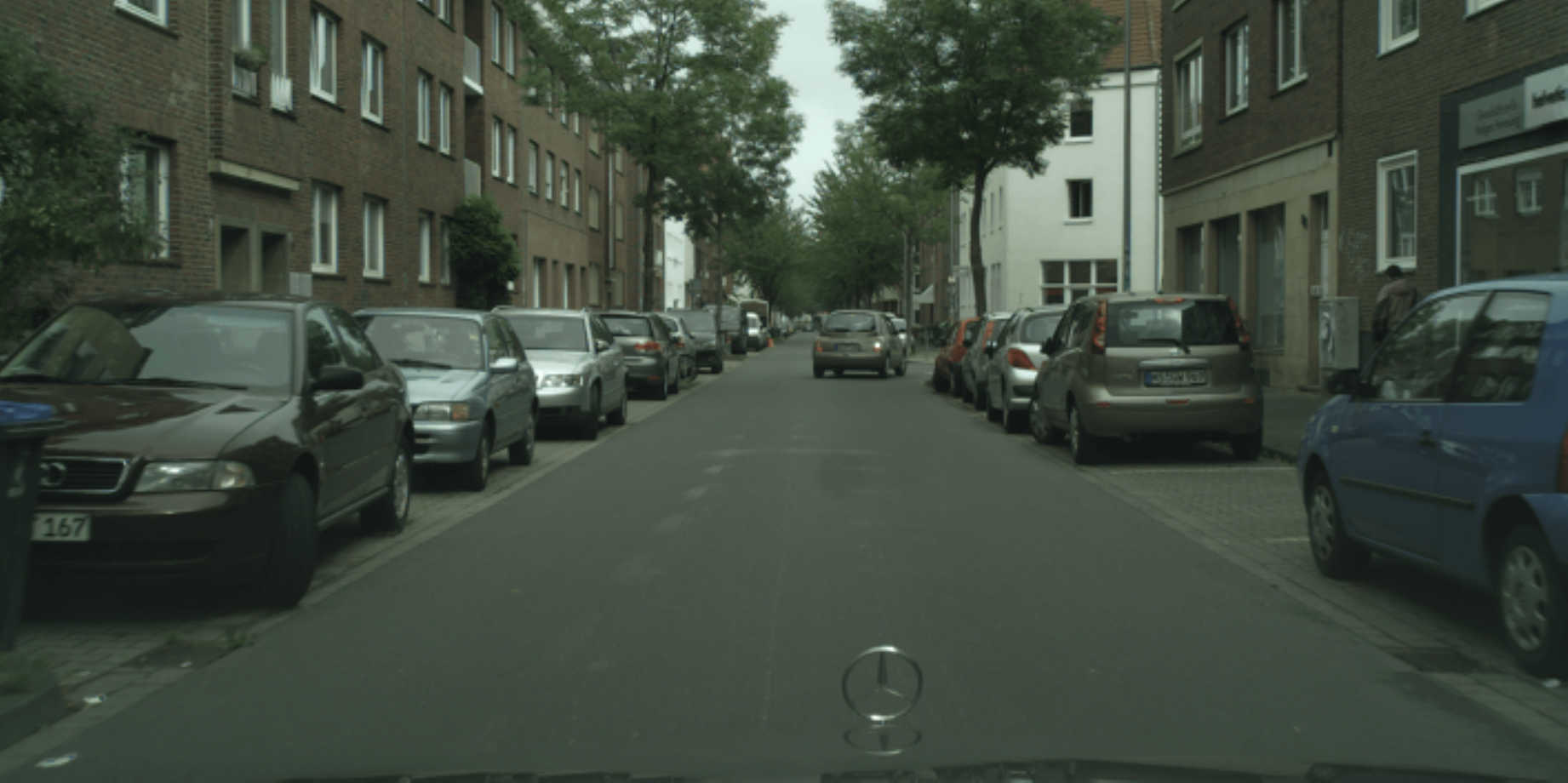}} 
 & \raisebox{-0.5\height}{\includegraphics[width=1.02\linewidth,height=0.51\linewidth]{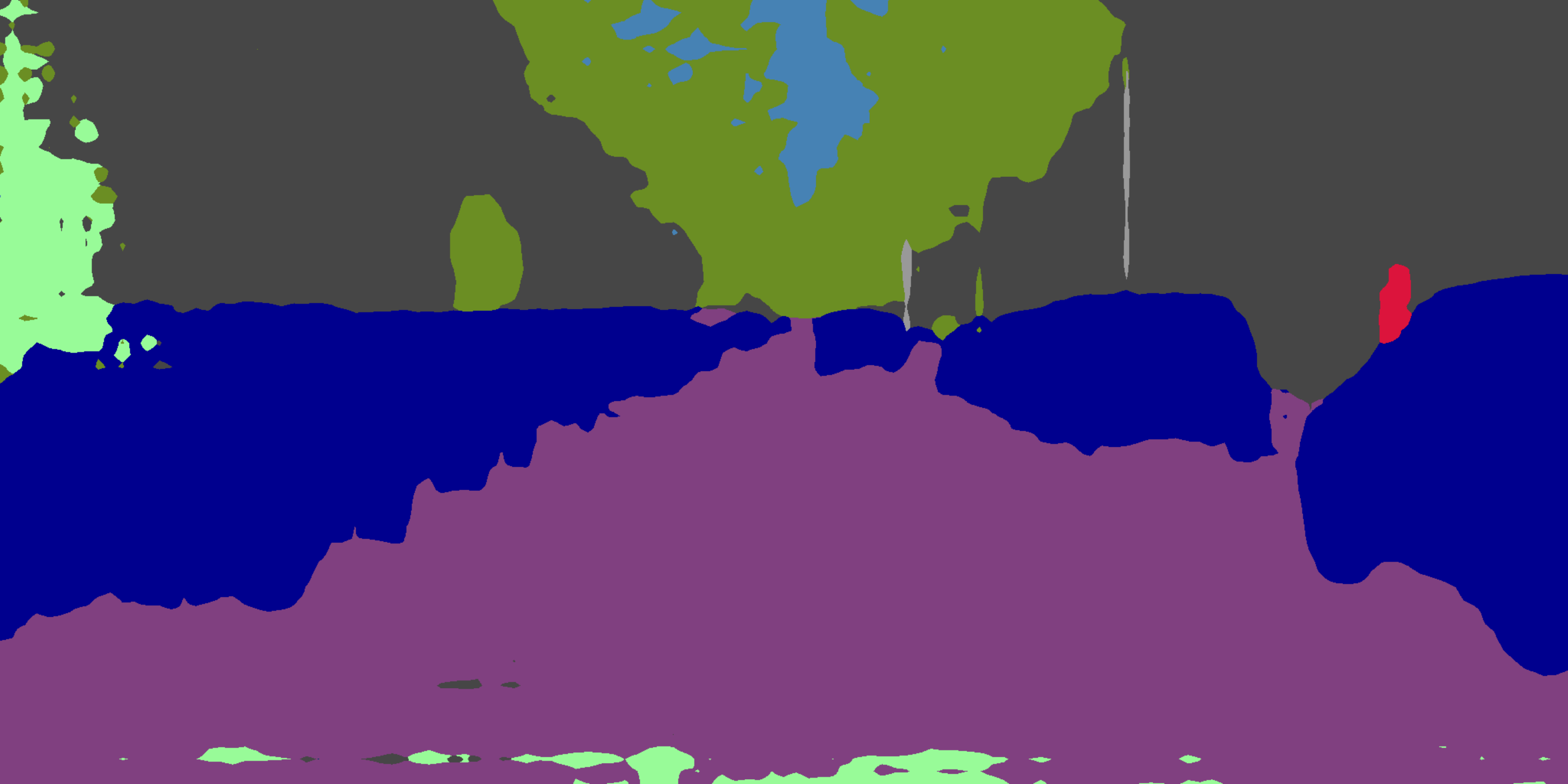}}
  & \raisebox{-0.5\height}{\includegraphics[width=1.02\linewidth,height=0.51\linewidth]{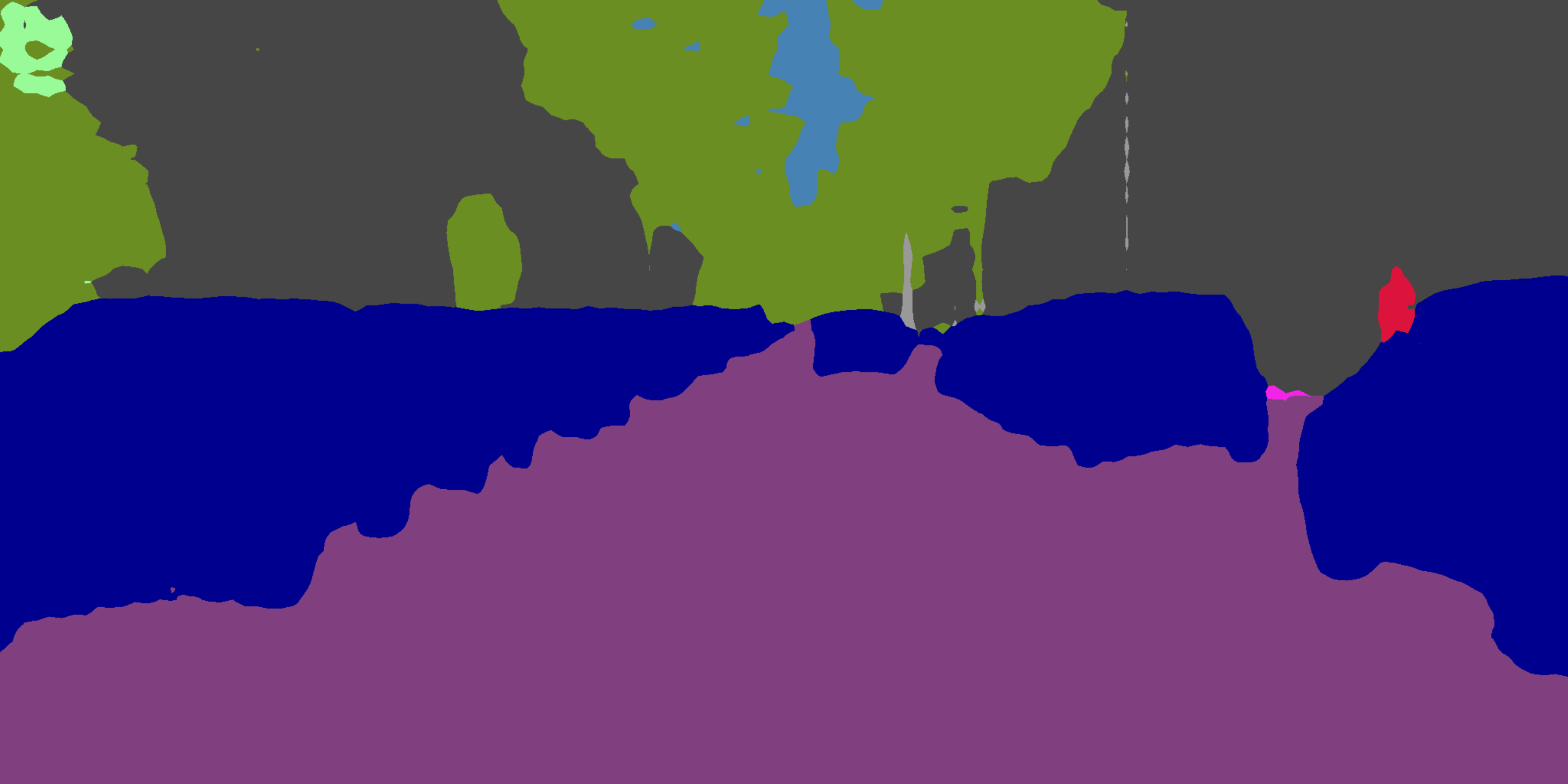}}
& \raisebox{-0.5\height}{\includegraphics[width=1.02\linewidth,height=0.51\linewidth]{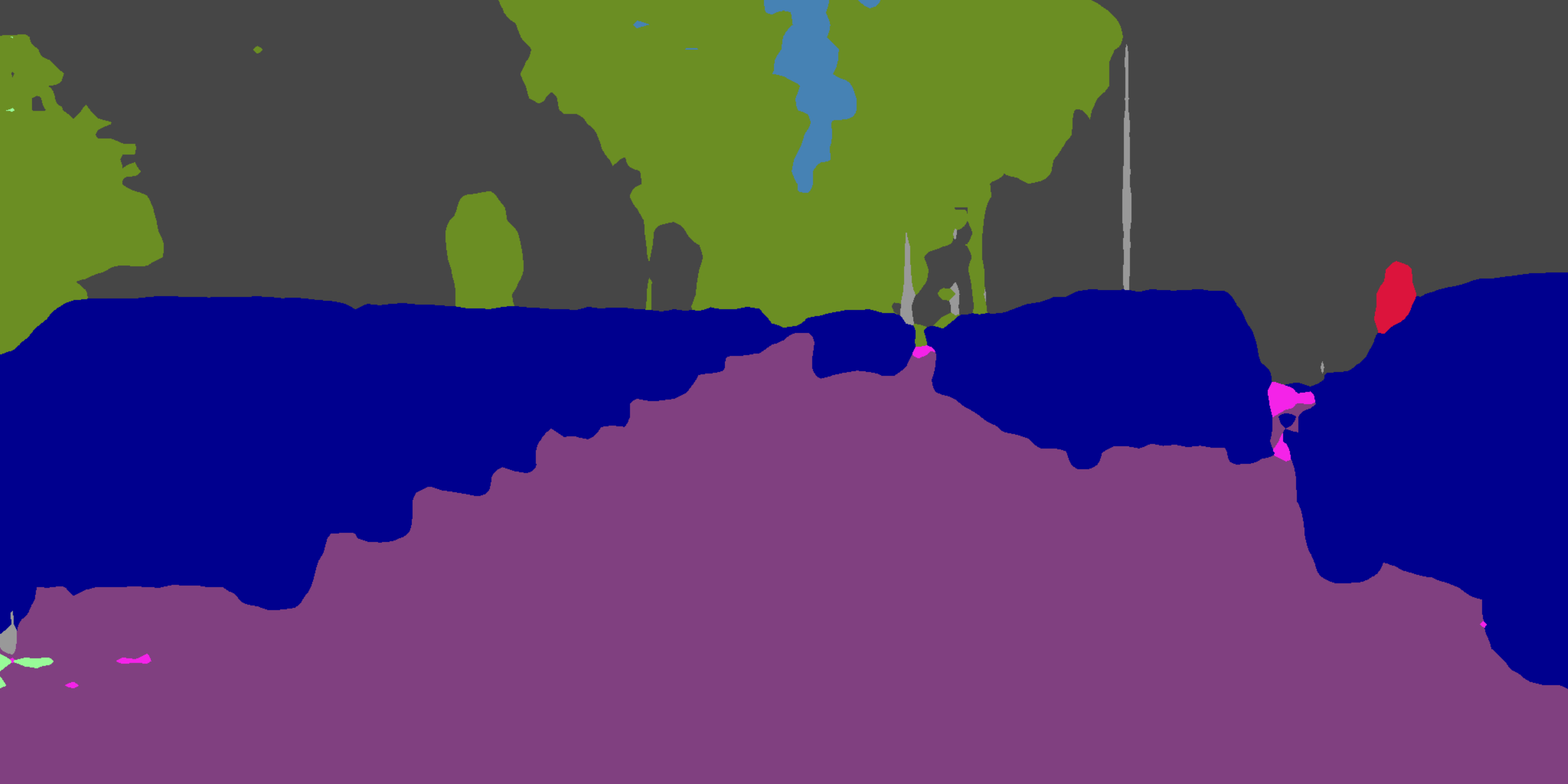}}
& \raisebox{-0.5\height}{\includegraphics[width=1.02\linewidth,height=0.51\linewidth]{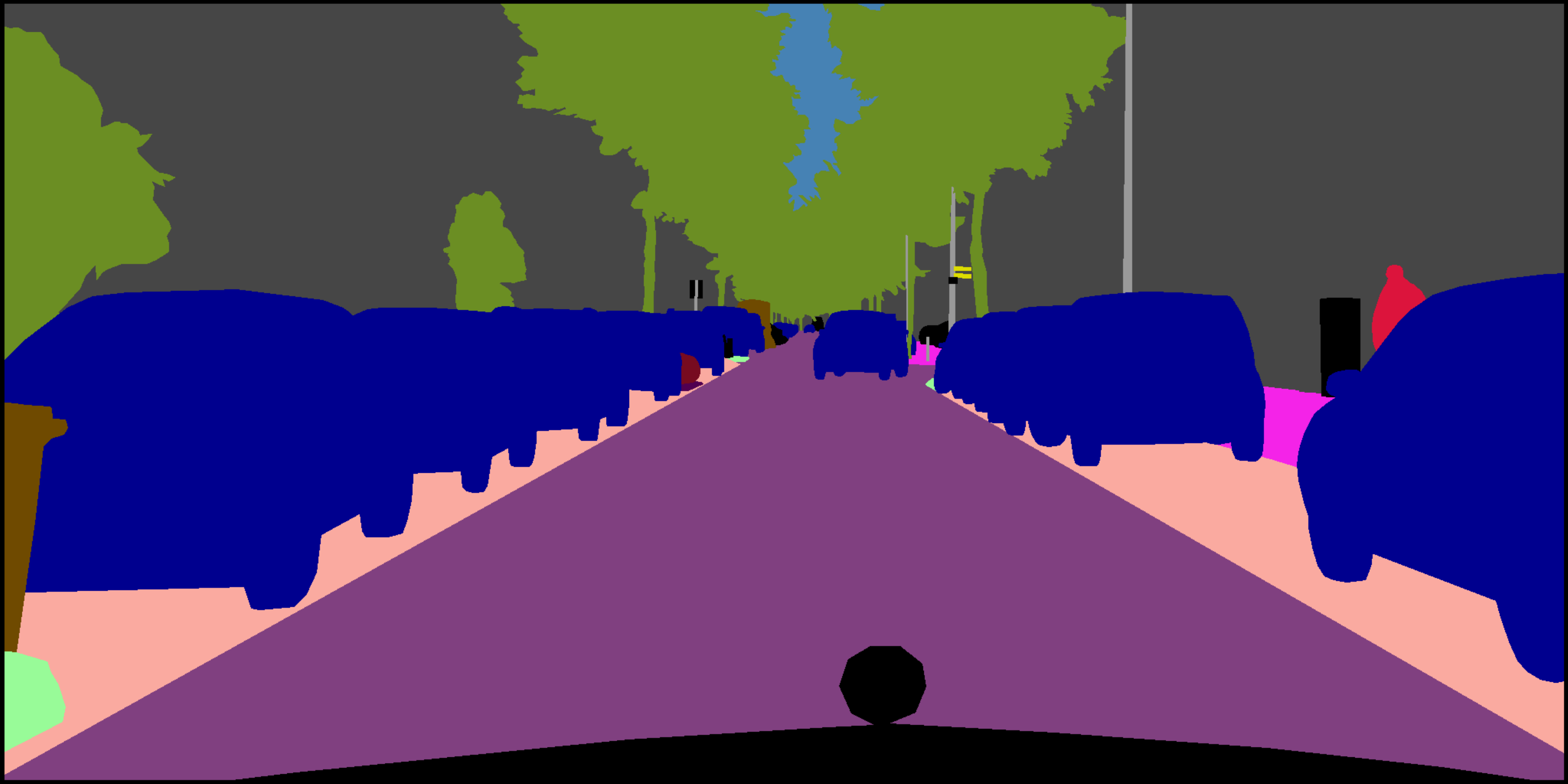}}
\\

\raisebox{-0.5\height}{\includegraphics[width=1.02\linewidth,height=0.51\linewidth]{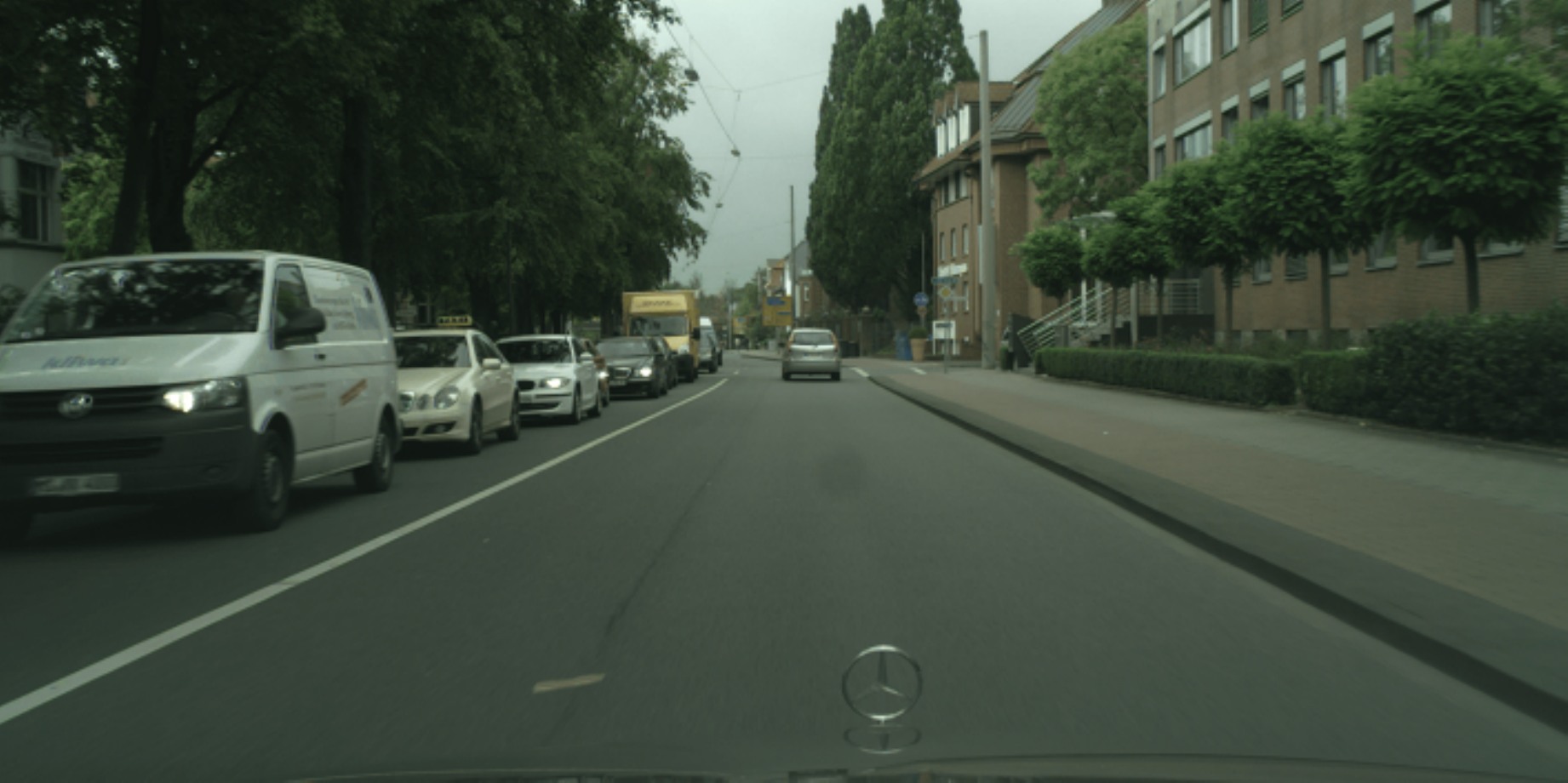}} 
 & \raisebox{-0.5\height}{\includegraphics[width=1.02\linewidth,height=0.51\linewidth]{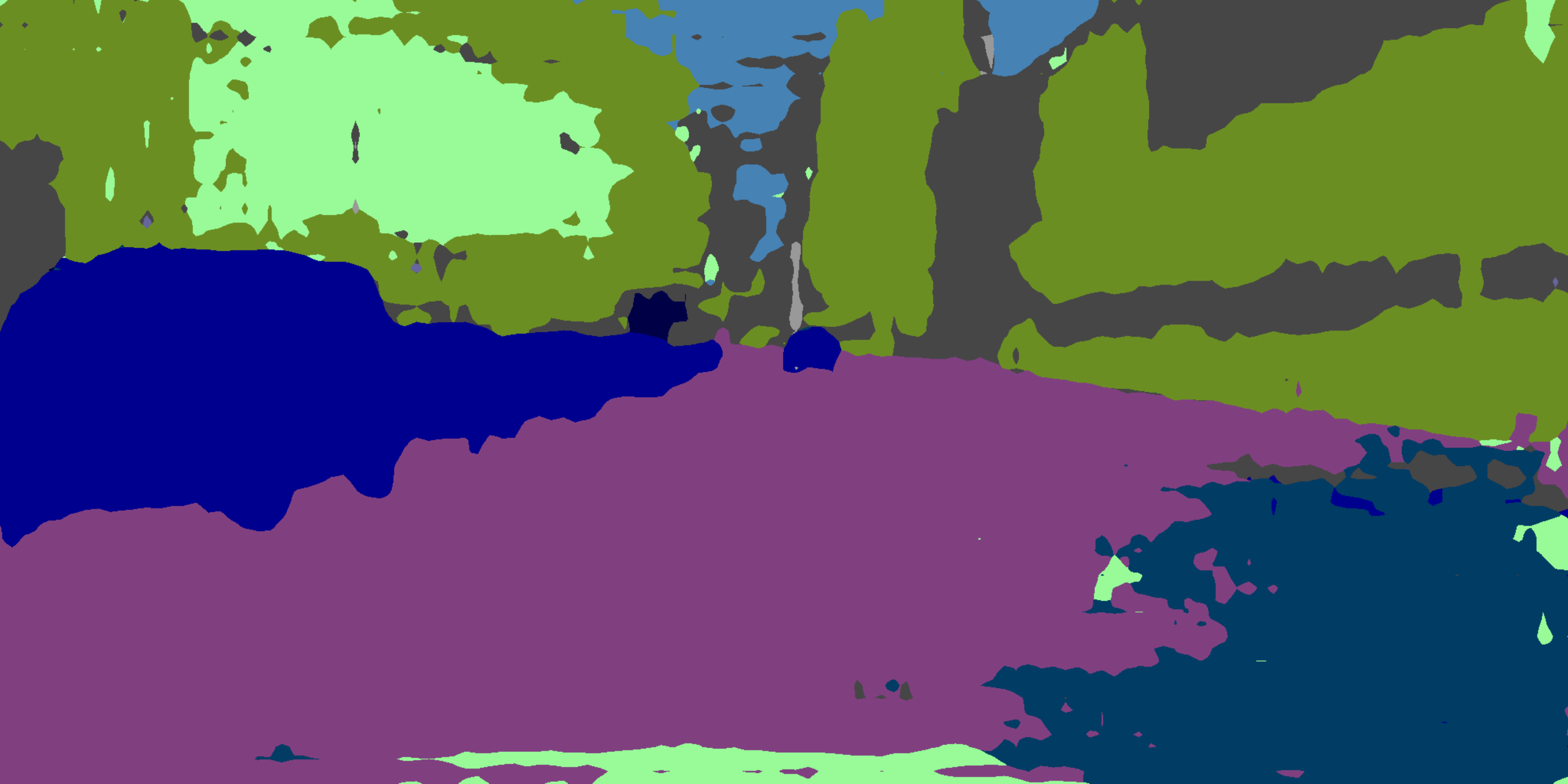}}
  & \raisebox{-0.5\height}{\includegraphics[width=1.02\linewidth,height=0.51\linewidth]{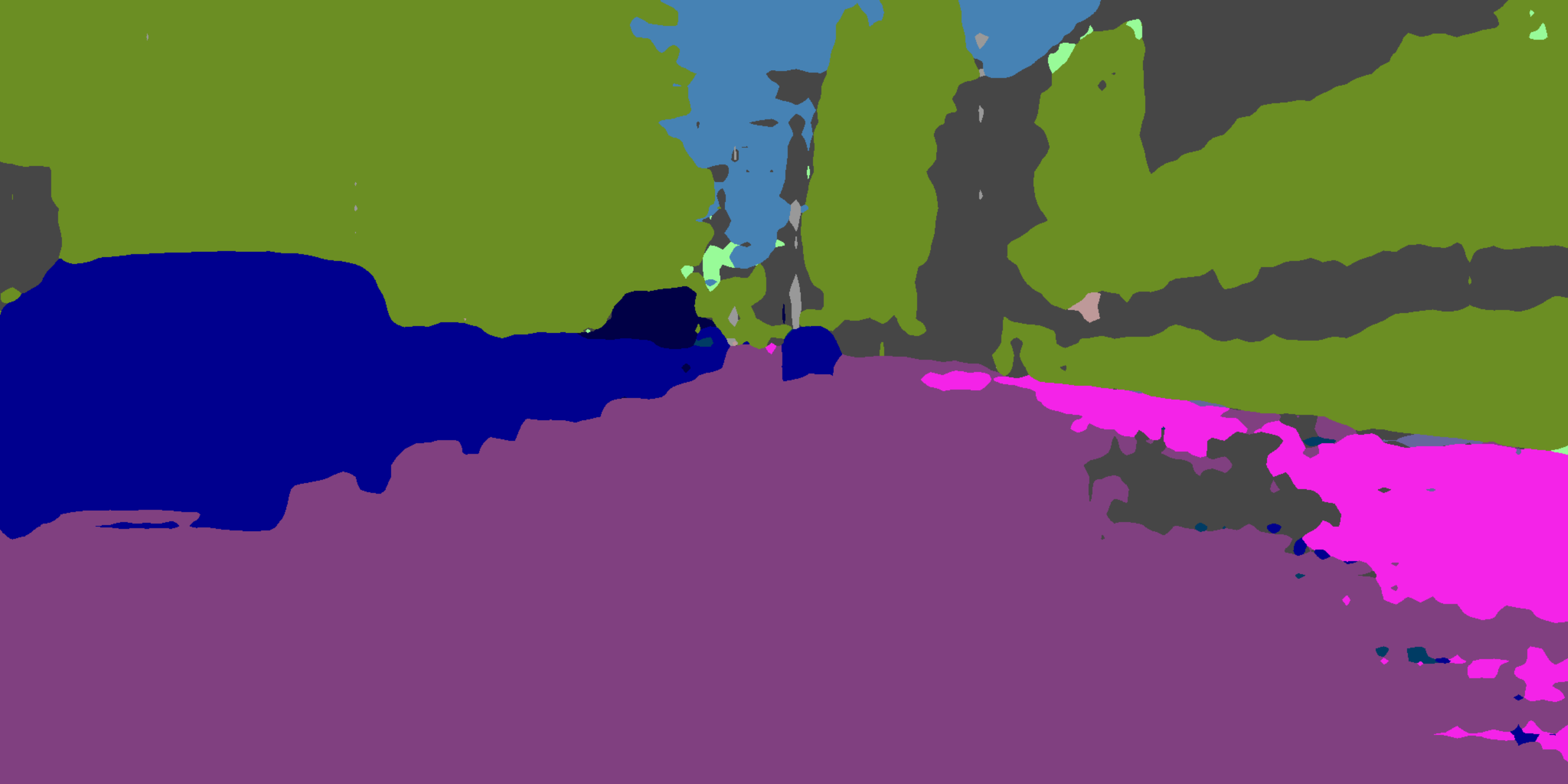}}
& \raisebox{-0.5\height}{\includegraphics[width=1.02\linewidth,height=0.51\linewidth]{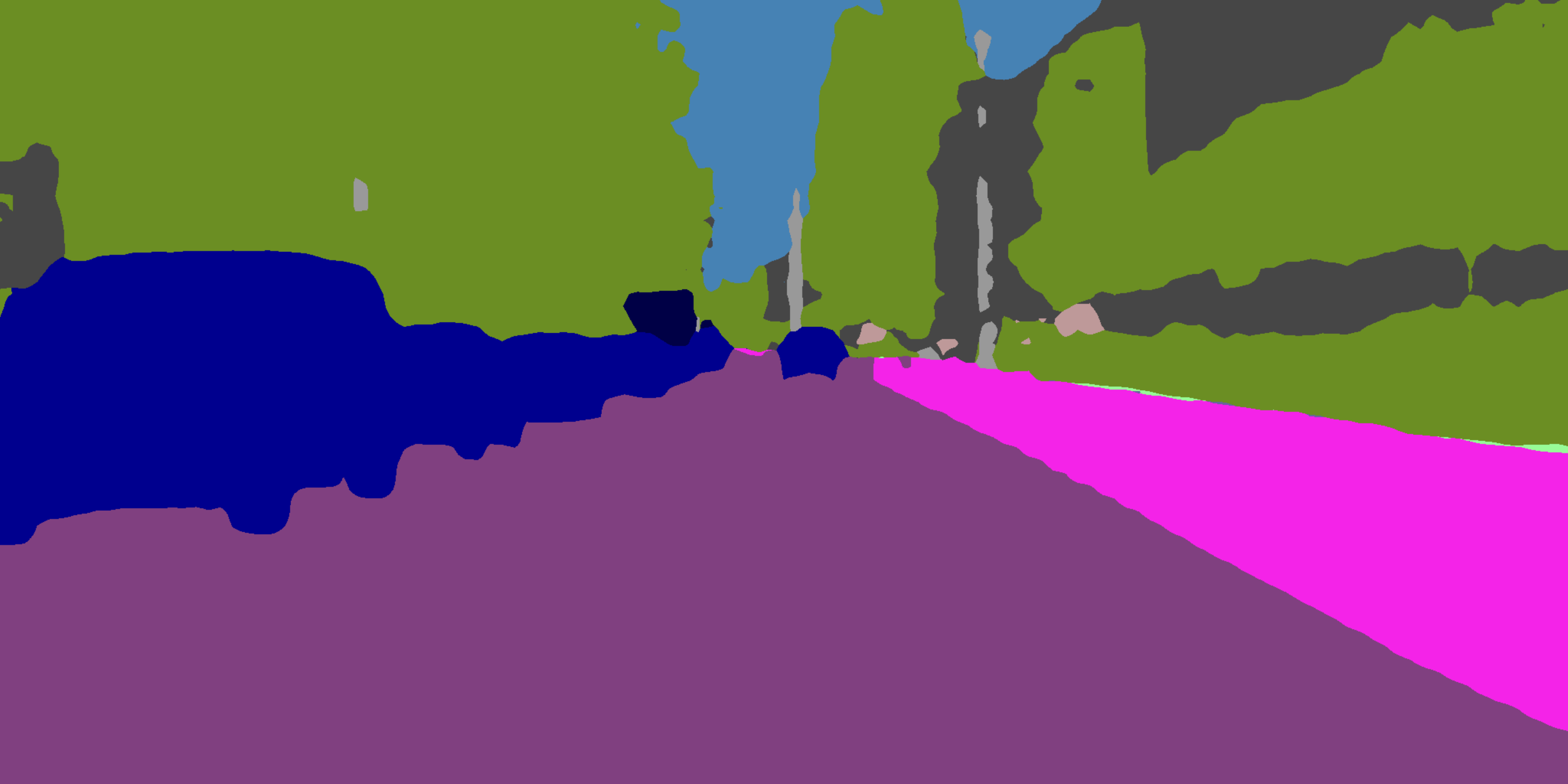}}
& \raisebox{-0.5\height}{\includegraphics[width=1.02\linewidth,height=0.51\linewidth]{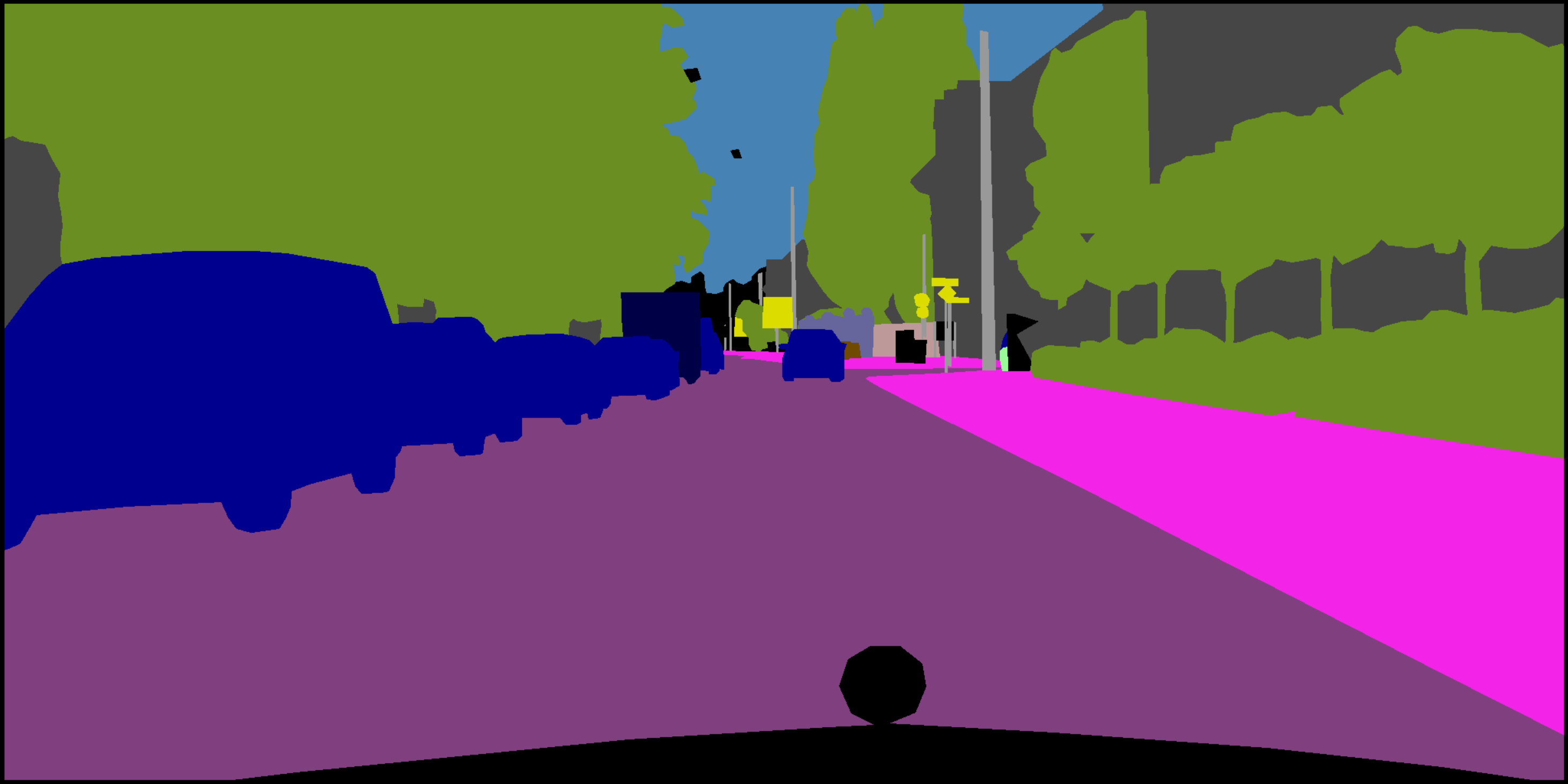}}
\\

\raisebox{-0.5\height}{\includegraphics[width=1.02\linewidth,height=0.51\linewidth]{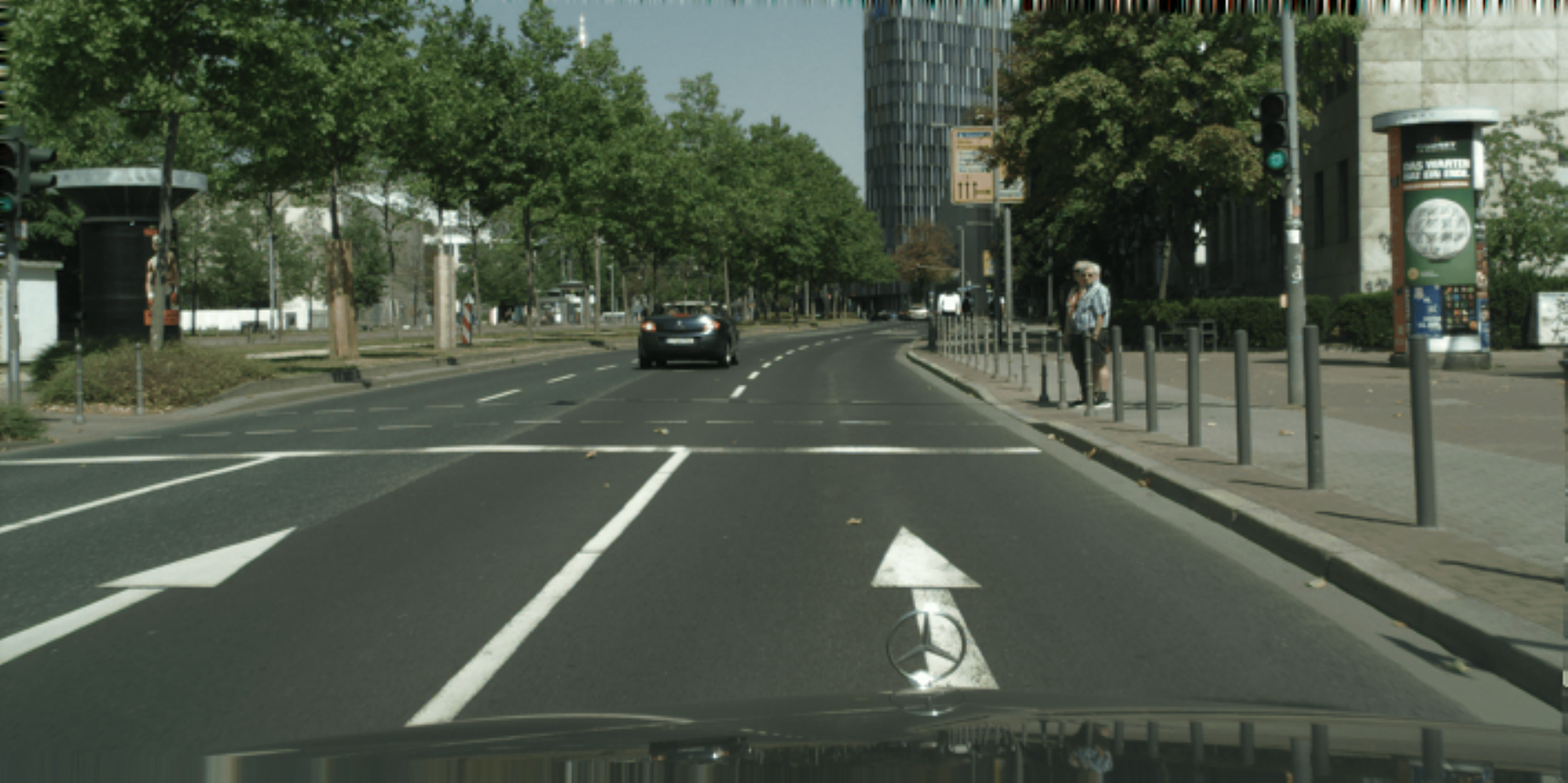}} 
 & \raisebox{-0.5\height}{\includegraphics[width=1.02\linewidth,height=0.51\linewidth]{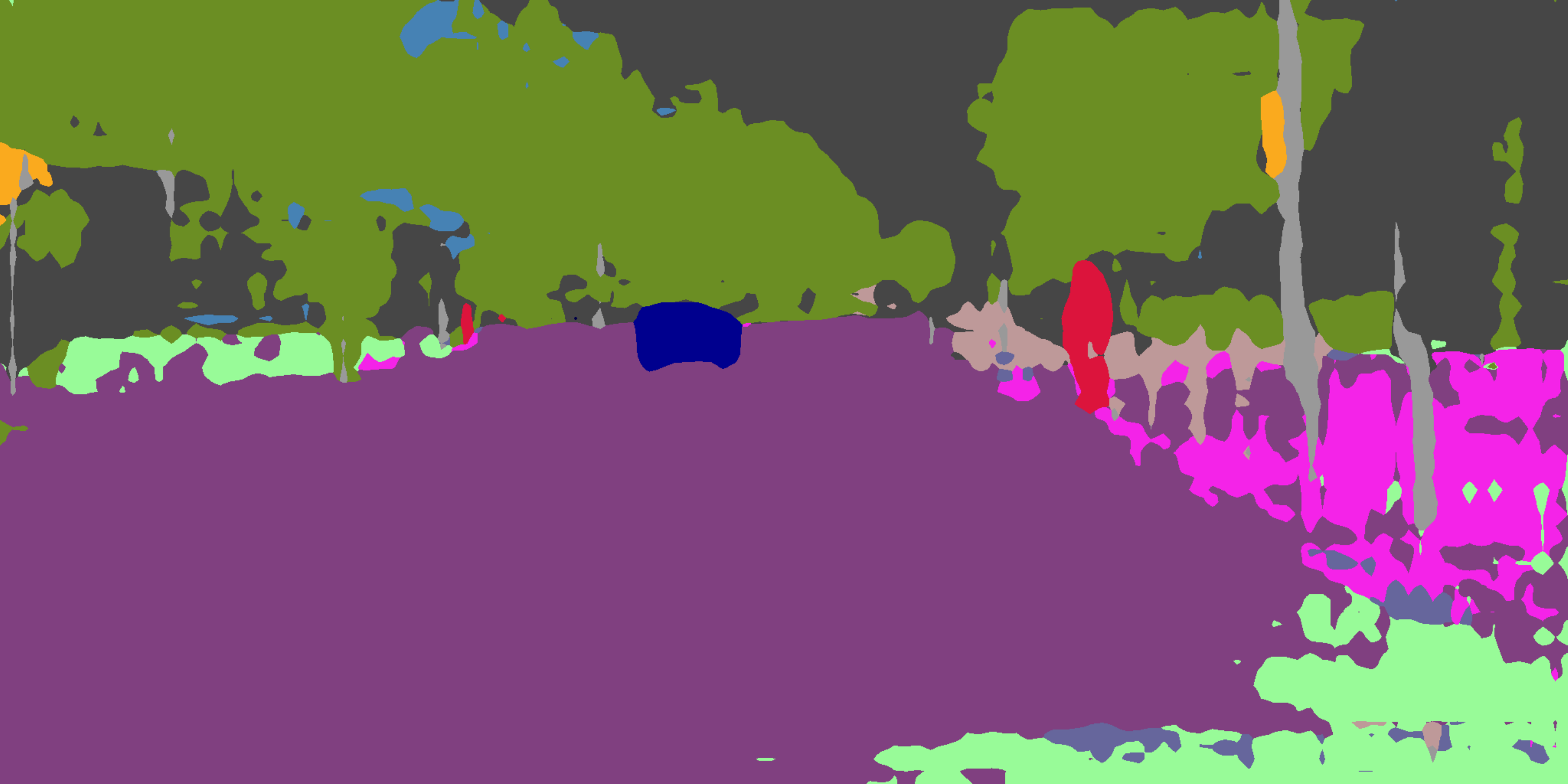}}
  & \raisebox{-0.5\height}{\includegraphics[width=1.02\linewidth,height=0.51\linewidth]{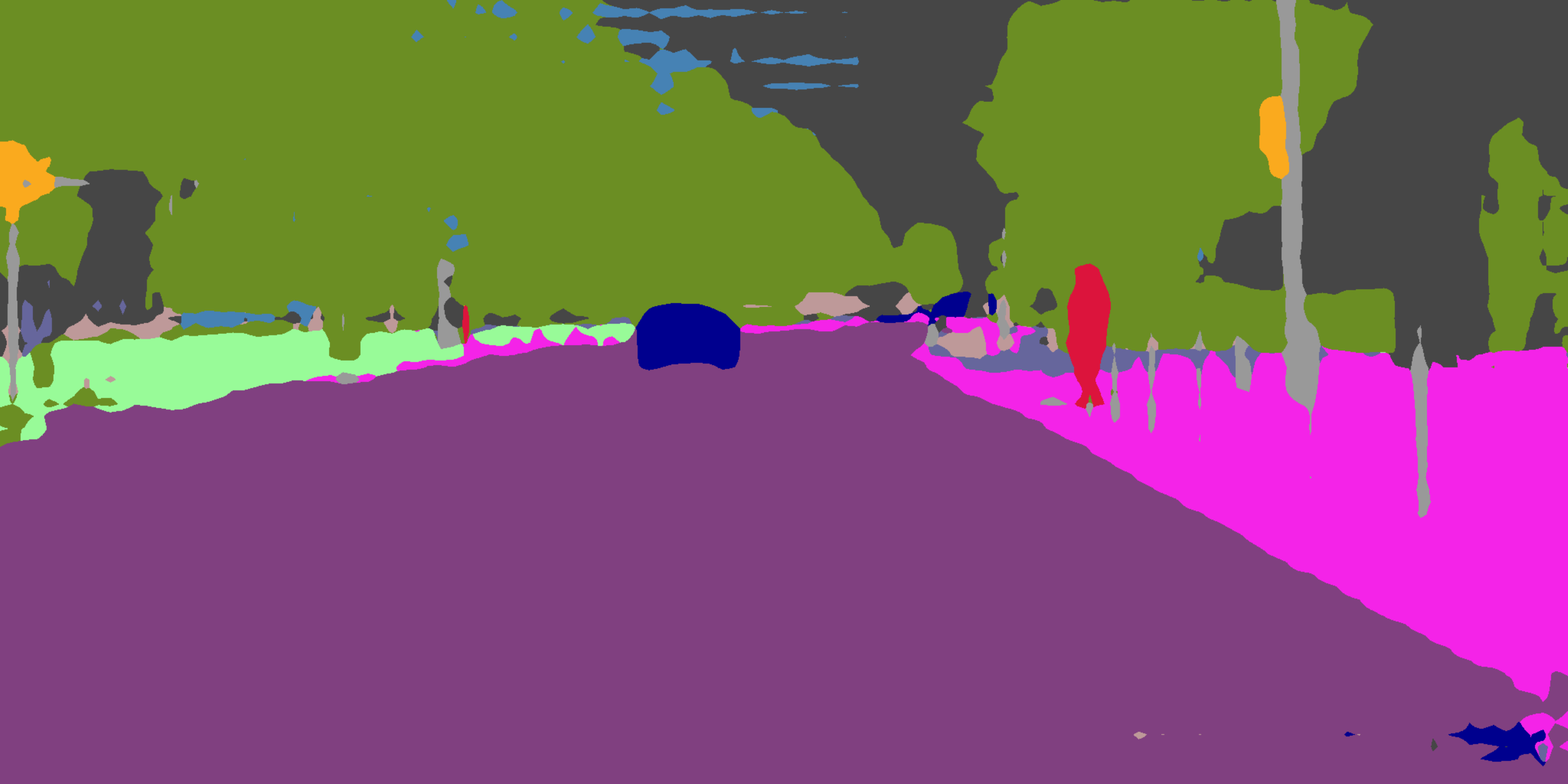}}
& \raisebox{-0.5\height}{\includegraphics[width=1.02\linewidth,height=0.51\linewidth]{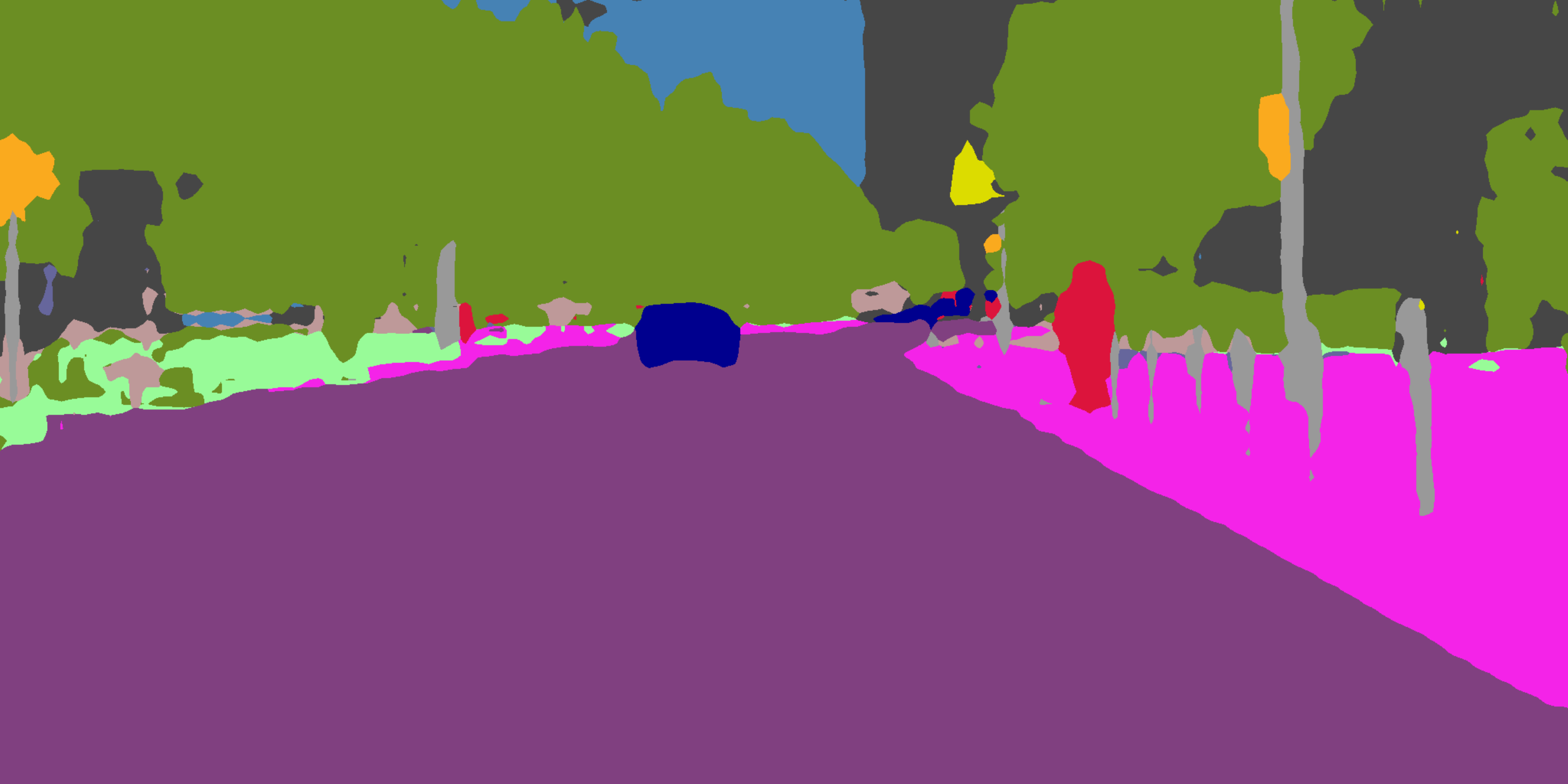}}
& \raisebox{-0.5\height}{\includegraphics[width=1.02\linewidth,height=0.51\linewidth]{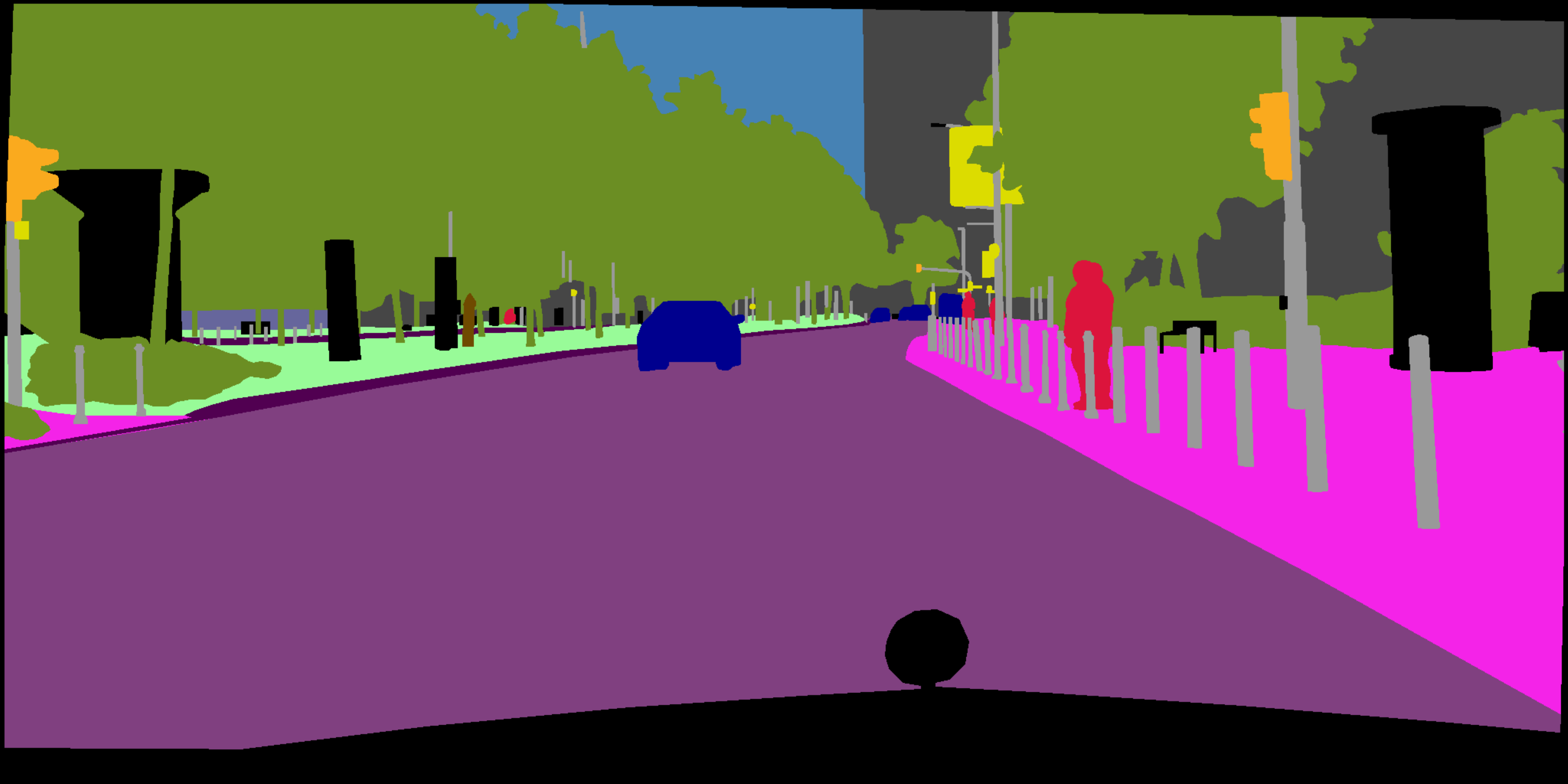}}
\\

\raisebox{-0.5\height}{\includegraphics[width=1.02\linewidth,height=0.51\linewidth]{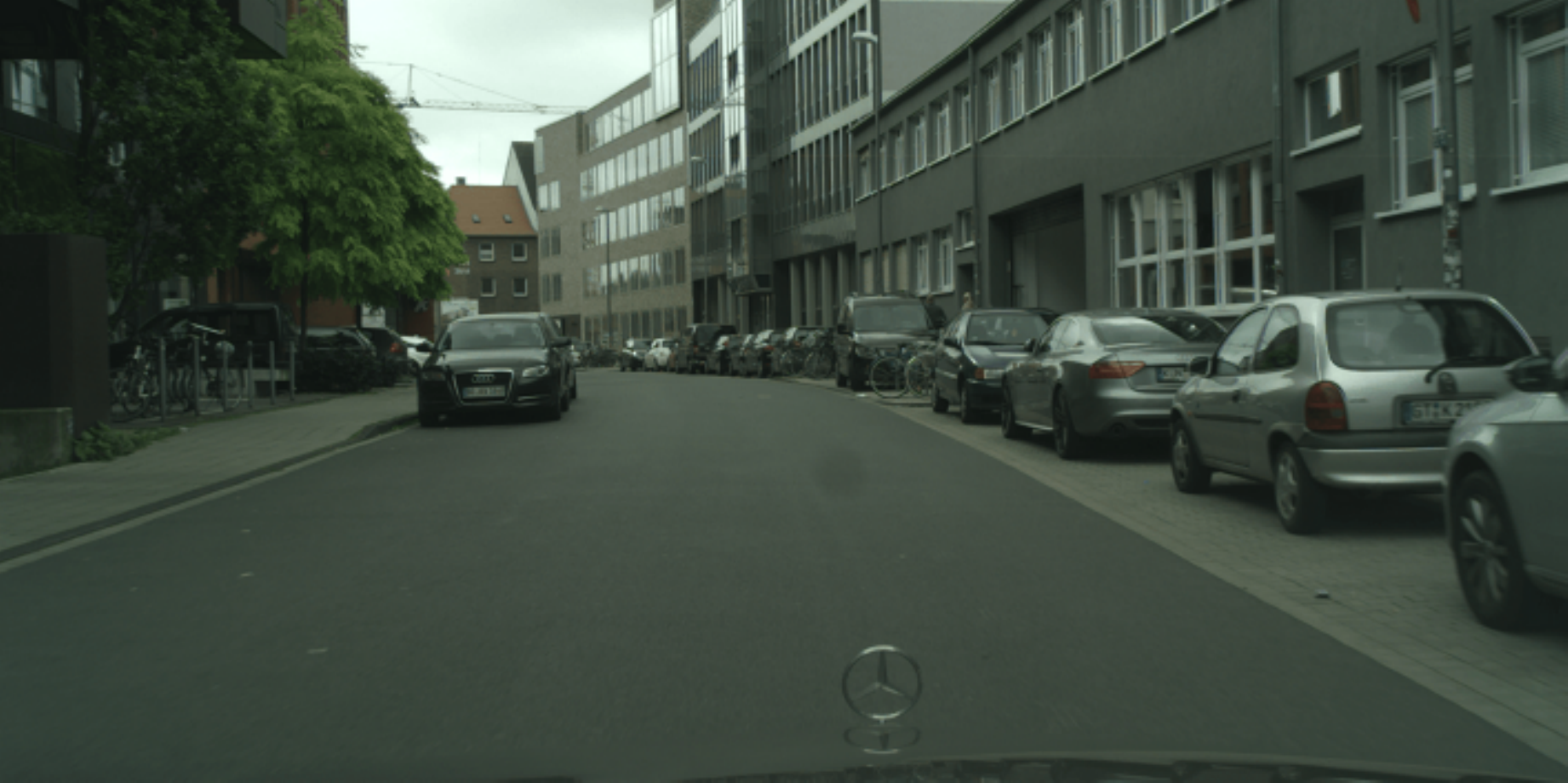}} 
 & \raisebox{-0.5\height}{\includegraphics[width=1.02\linewidth,height=0.51\linewidth]{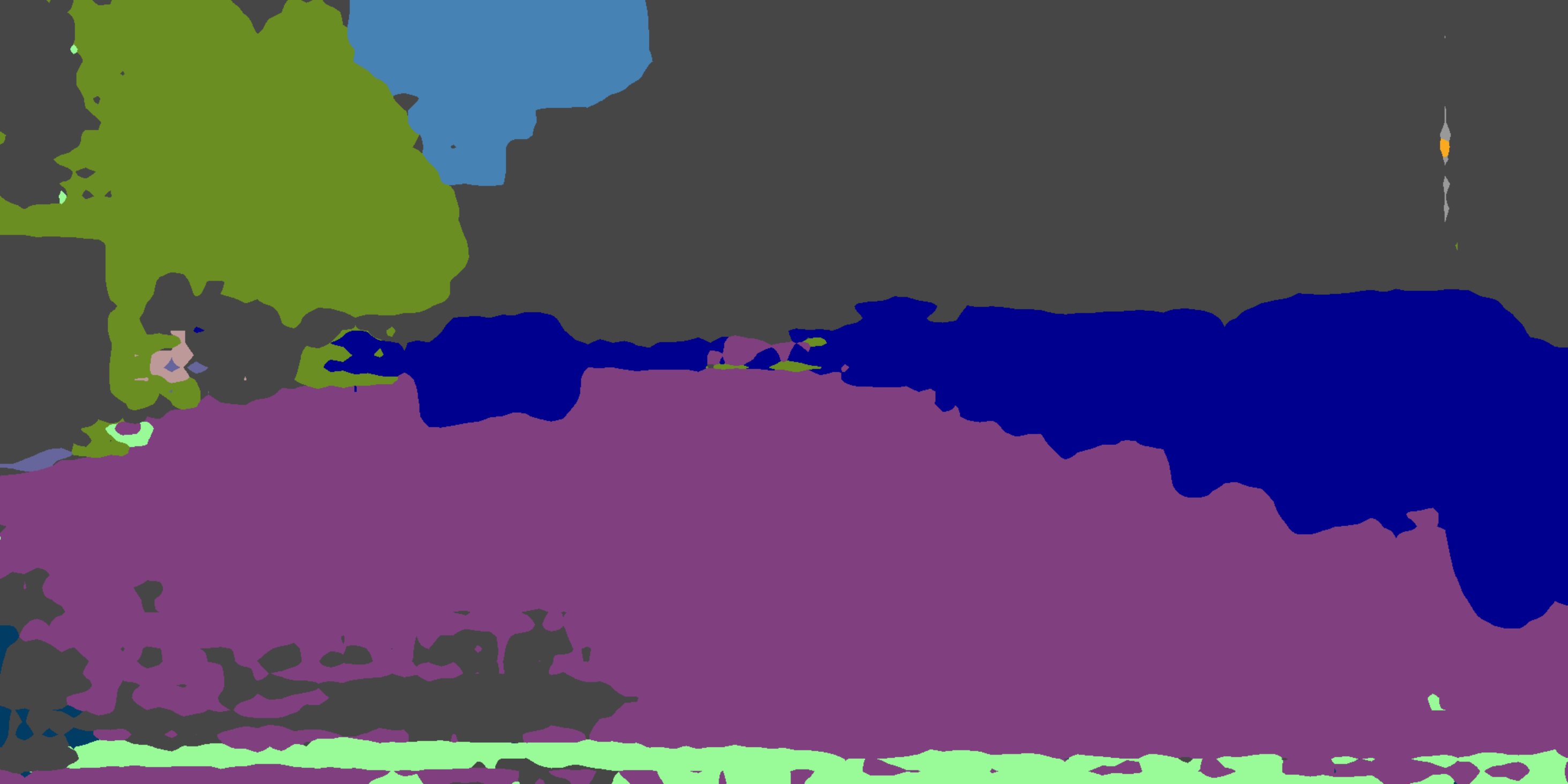}}
  & \raisebox{-0.5\height}{\includegraphics[width=1.02\linewidth,height=0.51\linewidth]{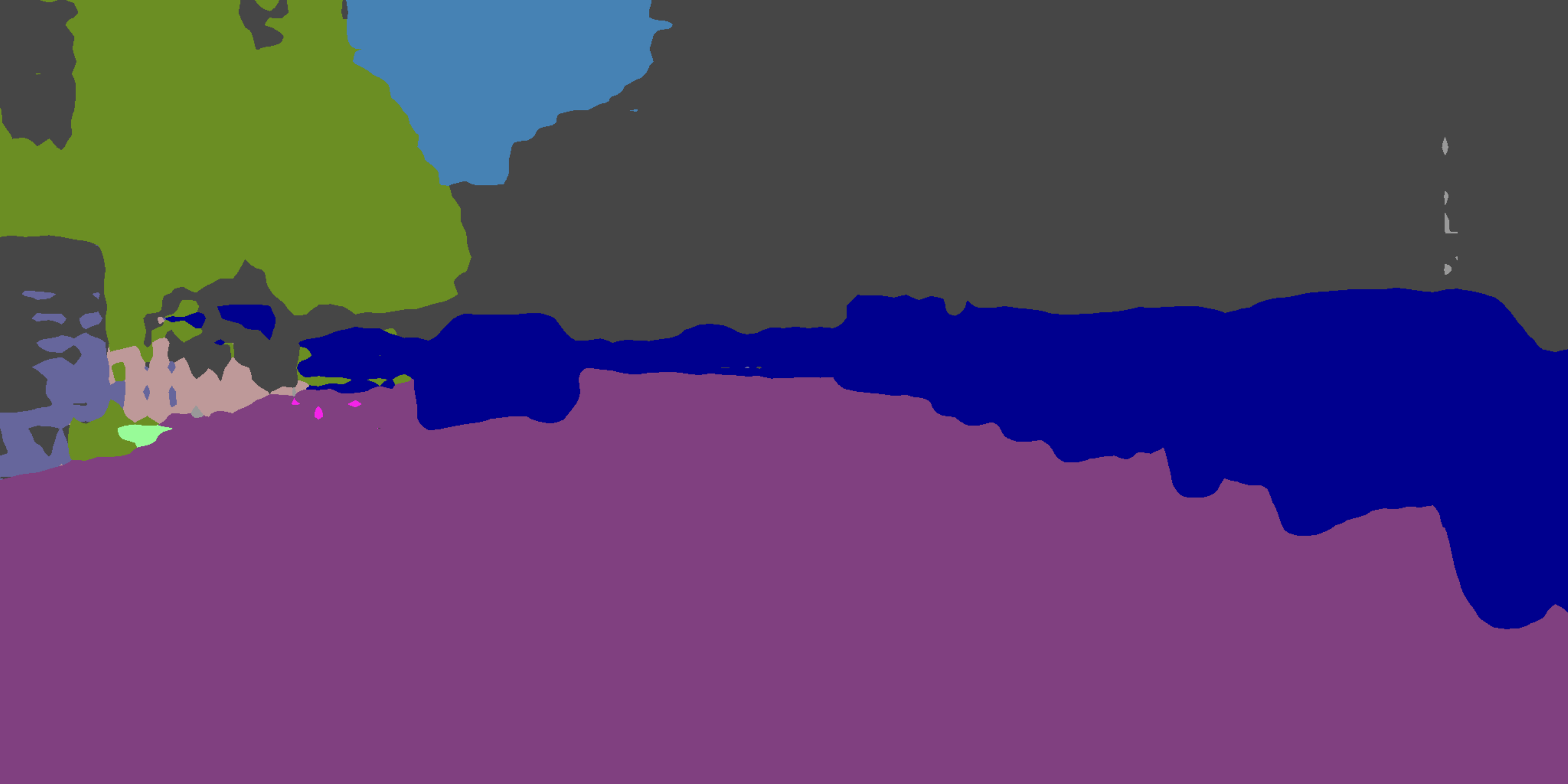}}
& \raisebox{-0.5\height}{\includegraphics[width=1.02\linewidth,height=0.51\linewidth]{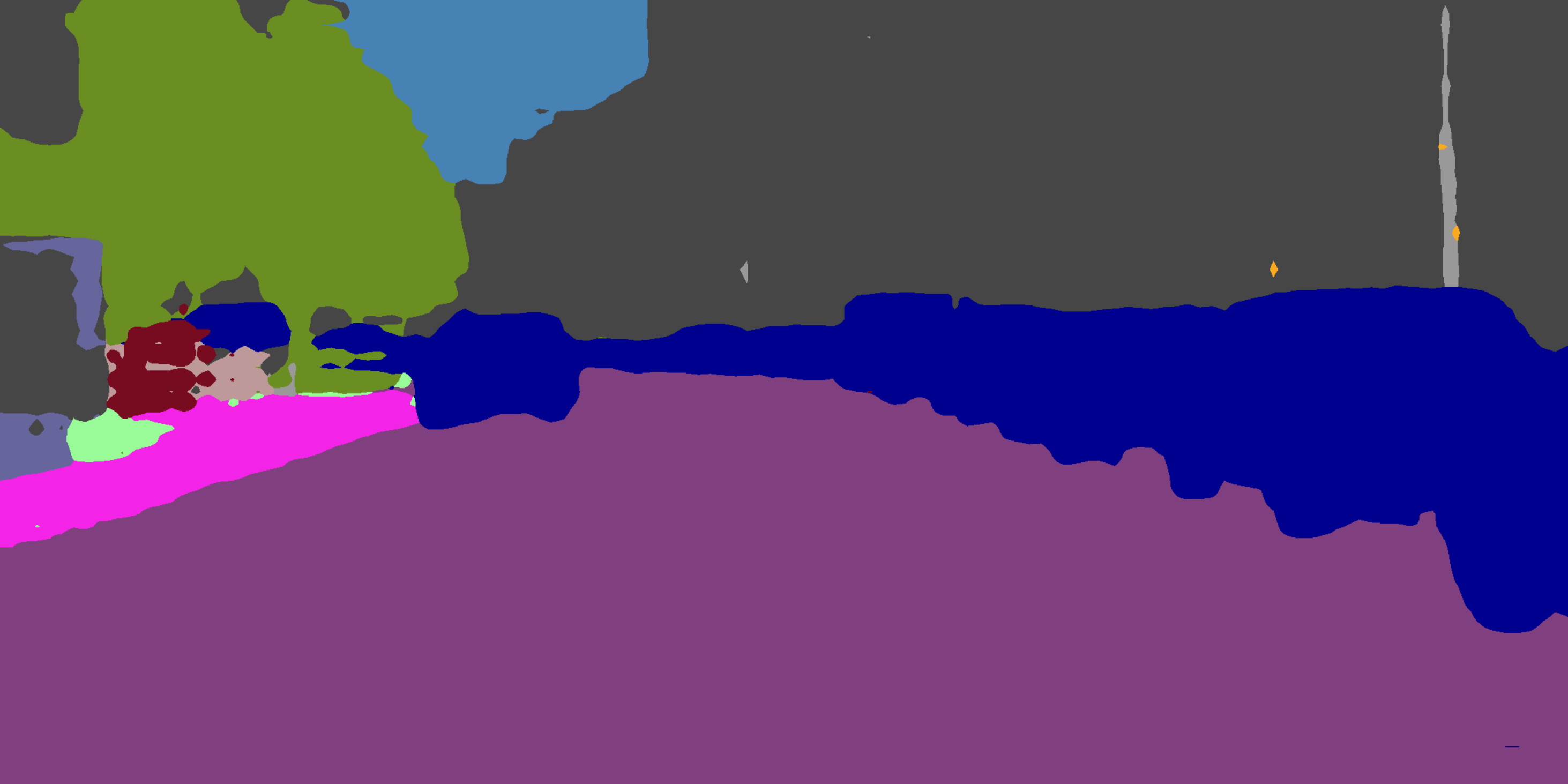}}
& \raisebox{-0.5\height}{\includegraphics[width=1.02\linewidth,height=0.51\linewidth]{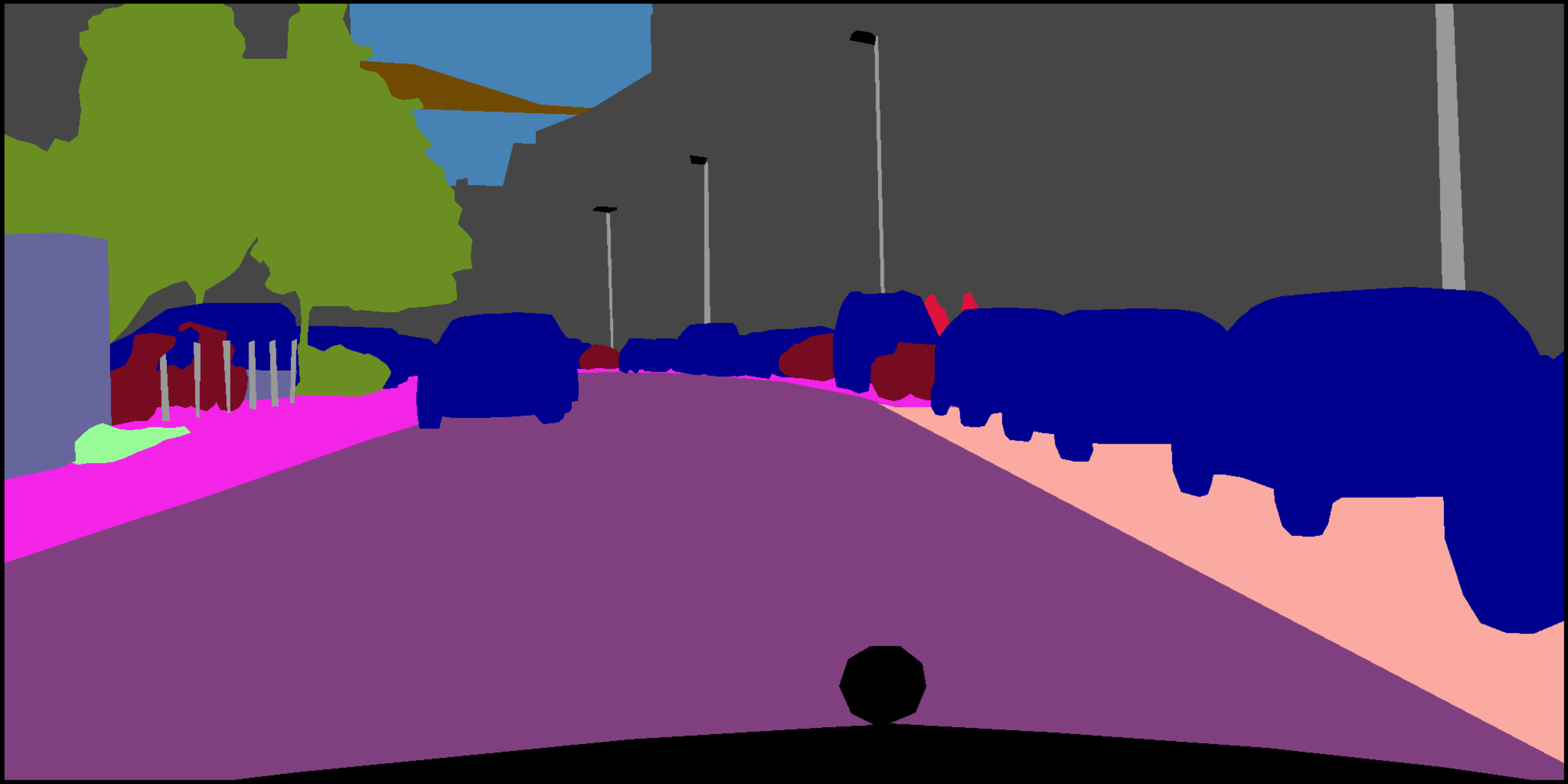}}
\\

\raisebox{-0.5\height}{\includegraphics[width=1.02\linewidth,height=0.51\linewidth]{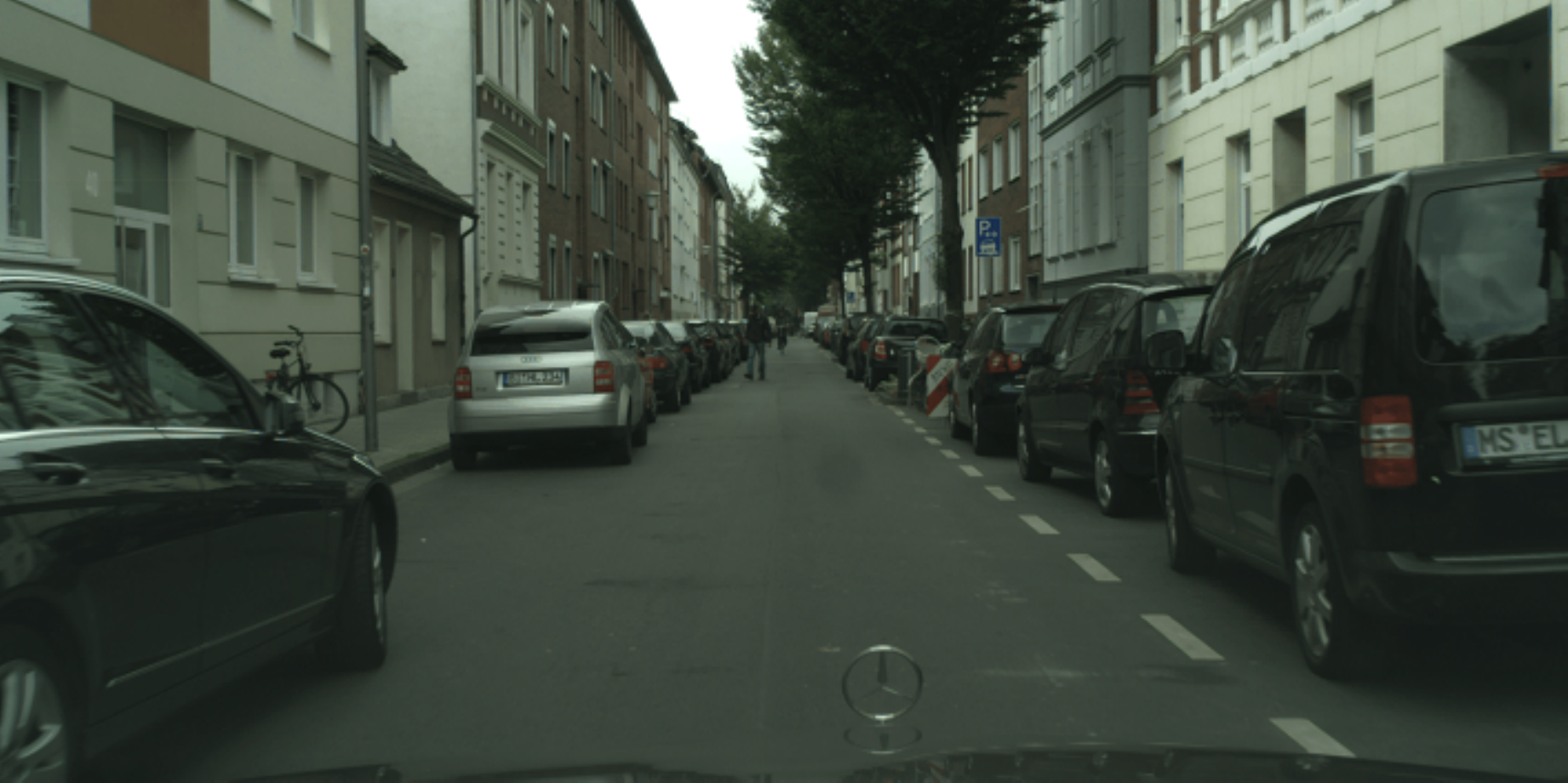}} 
 & \raisebox{-0.5\height}{\includegraphics[width=1.02\linewidth,height=0.51\linewidth]{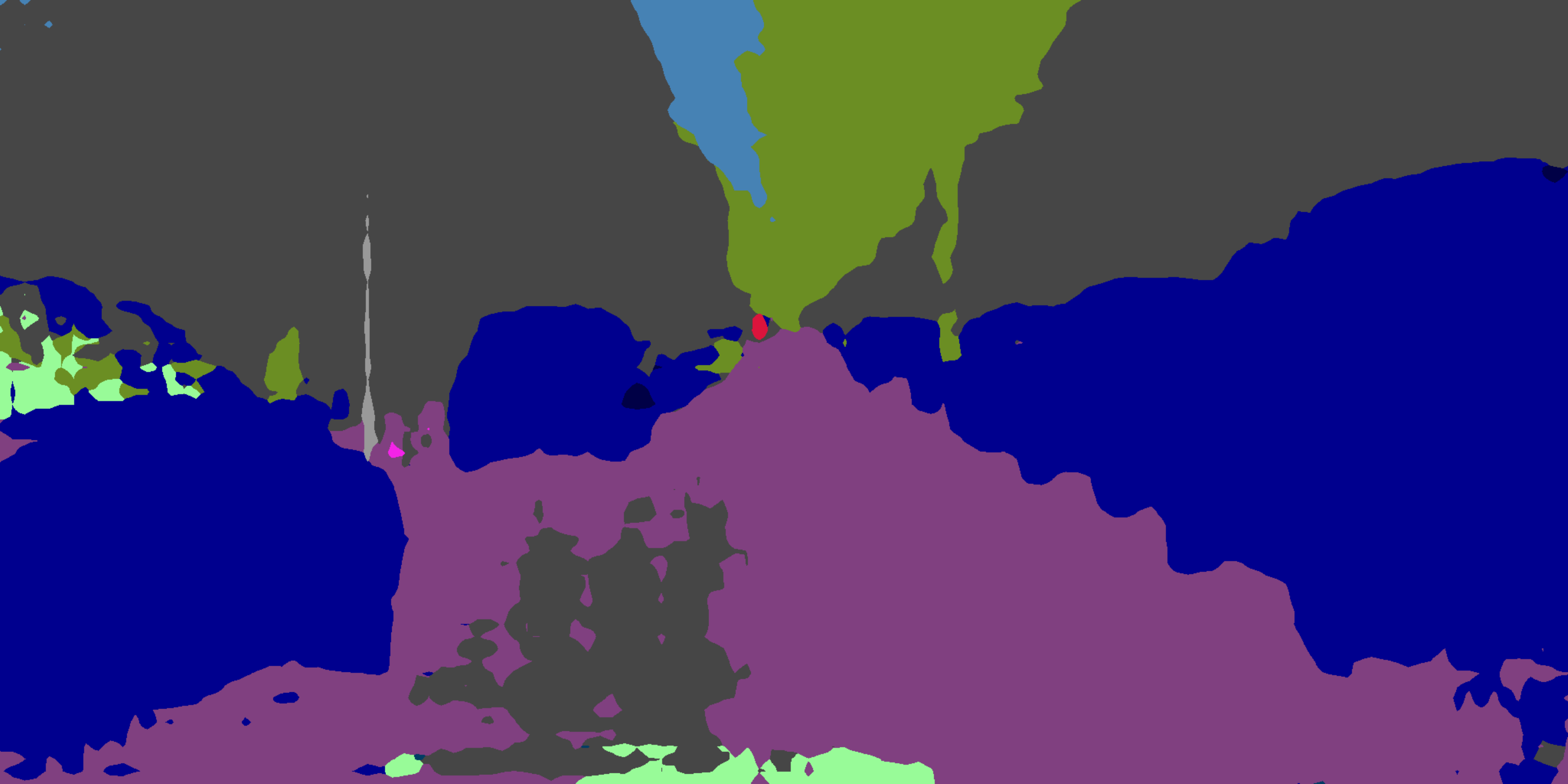}}
  & \raisebox{-0.5\height}{\includegraphics[width=1.02\linewidth,height=0.51\linewidth]{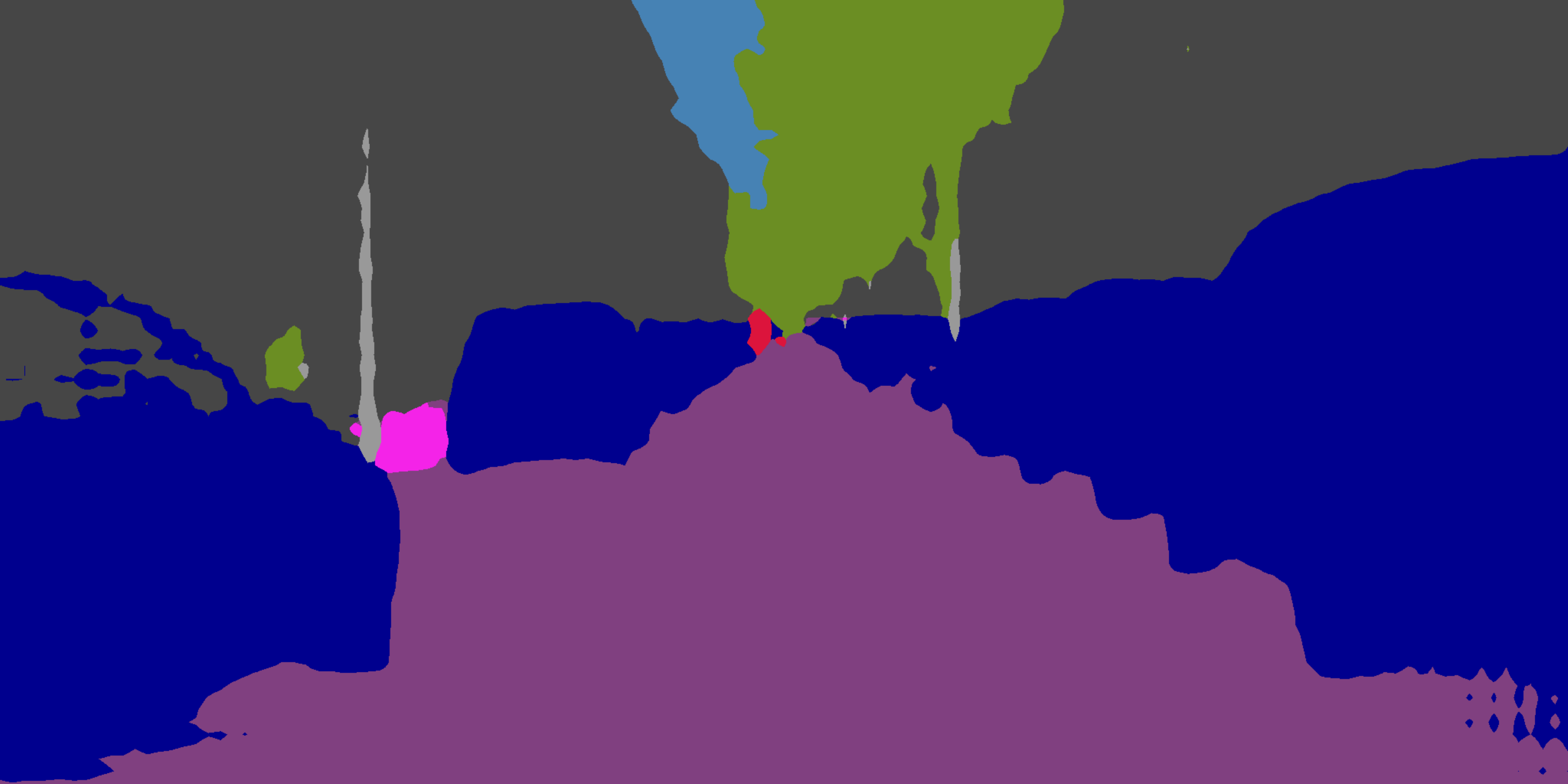}}
& \raisebox{-0.5\height}{\includegraphics[width=1.02\linewidth,height=0.51\linewidth]{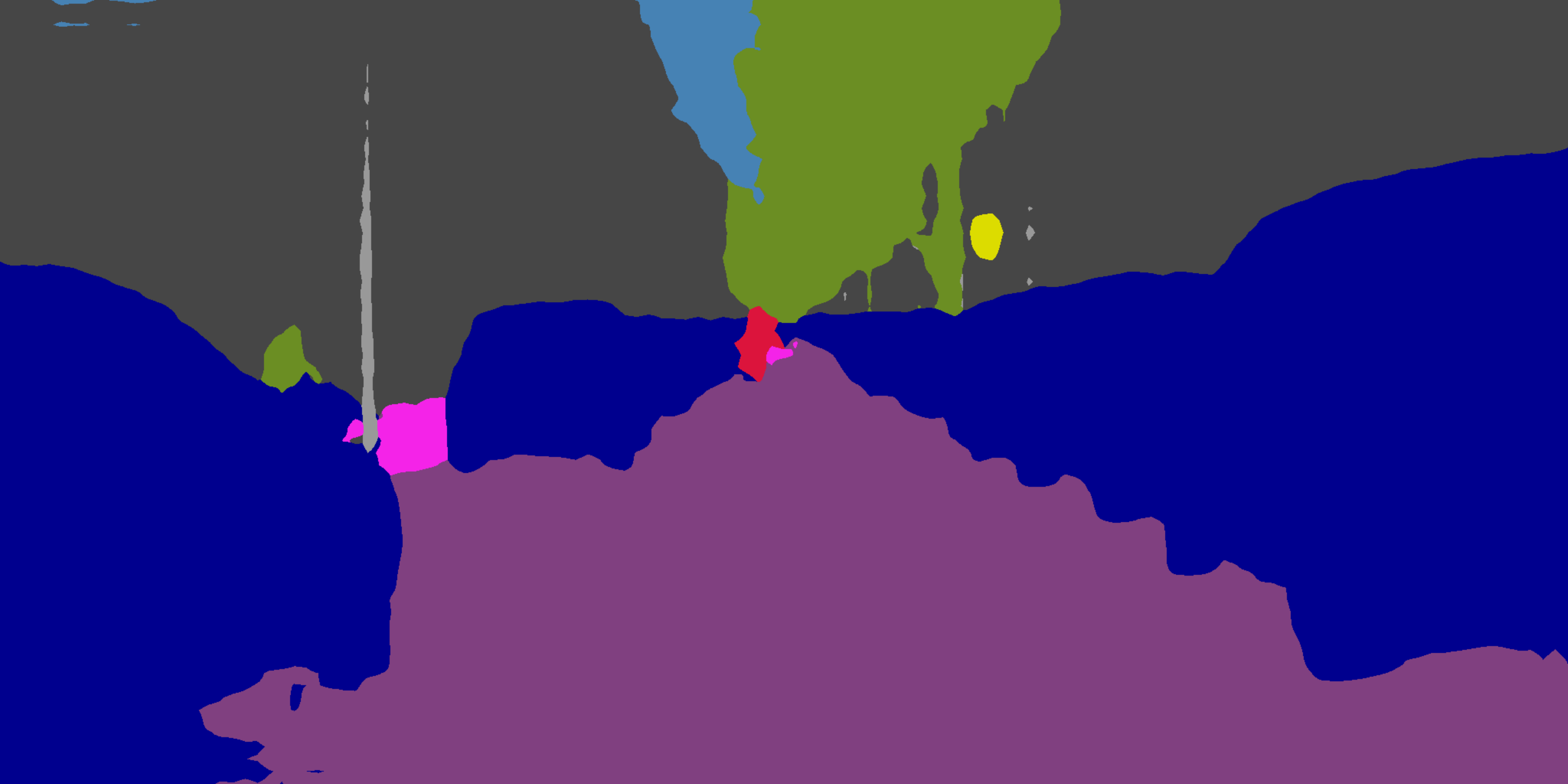}}
& \raisebox{-0.5\height}{\includegraphics[width=1.02\linewidth,height=0.51\linewidth]{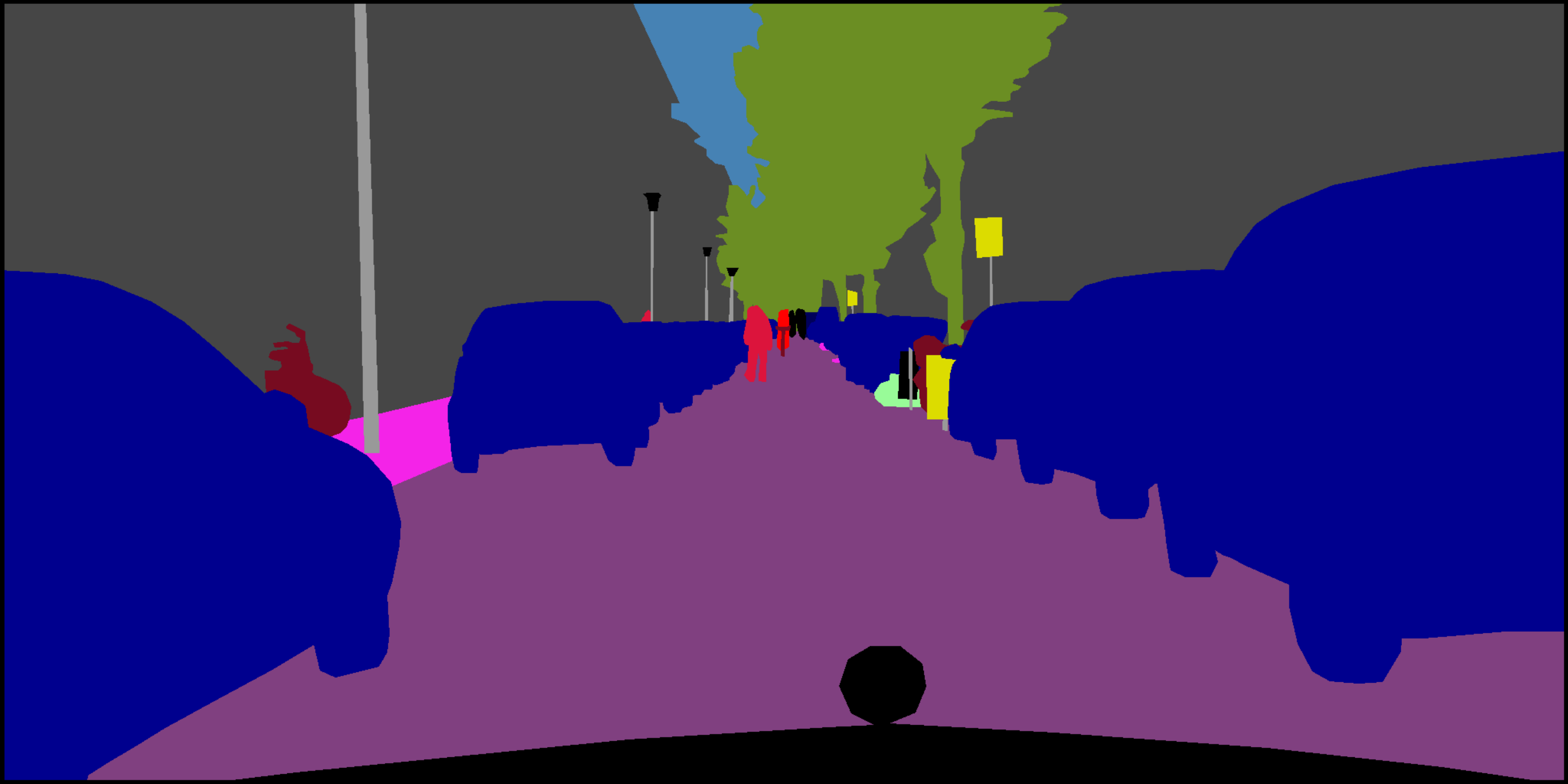}}
\\

\raisebox{-0.5\height}{\footnotesize{Target Image}}
 & \raisebox{-0.5\height}{\footnotesize{Without Ada.}}
& \raisebox{-0.5\height}{\footnotesize{Ada.(AdvEnt)}}
& \raisebox{-0.5\height}{\footnotesize{Ada.(CrCDA)}}
& \raisebox{-0.5\height}{\footnotesize{Ground Truth}}
\\


\end{tabular}
\caption{Qualitative results for GTA5 $\rightarrow$ Cityscapes. Our approach (CrCDA) aligns low-level features ($e.g.$, boundaries of sidewalk, car and person $etc.$) as well as high-level features by multi-scale adversarial learning. In contrast, AdvEnt ignores low-level information because global alignment focuses more on high-level information. Thus, as shown above, CrCDA achieves both local and global consistencies while AdvEnt only achieves global consistency.}
\label{results_supple}
\end{figure*}

We share more qualitative experimental results for GTA5 $\rightarrow$ Cityscapes as shown in Fig. \ref{results_supple}. As Fig. \ref{results_supple} shows, our CrCDA aligns both low-level features ($e.g.$, boundaries of sidewalk, car and person $etc.$) and high-level features by multi-scale adversarial learning. As a comparison, AdvEnt neglects low-level information which focuses more on high-level features. As a result, CrCDA achieves both local and global consistencies in segmentation while AdvEnt achieves global consistency only.

\end{document}